\documentclass[12pt]{article}
\usepackage{amsmath,amssymb,amsthm}
\usepackage{latexsym}
\usepackage{chngcntr}
\usepackage{cite}
\usepackage{hyperref}
\usepackage{appendix}
\usepackage{epsfig}
\usepackage{epstopdf}
\usepackage{amsfonts,amscd}
\usepackage{xspace}
\usepackage[ruled,vlined]{algorithm2e}
\usepackage{color,comment}
\usepackage{amsbsy,bm}
\usepackage{mathrsfs}
\usepackage{makecell}
\usepackage{authblk}
\usepackage{booktabs}
\usepackage{tabularx}
\usepackage{xurl}
\usepackage{array}
\usepackage{rotating}

\usepackage{float}
\usepackage{caption}
\usepackage{placeins}
\counterwithin{figure}{section}
\counterwithin{table}{section}
\counterwithin{equation}{section}

\newcommand{\be}{\begin{eqnarray}}
\newcommand{\ee}{\end{eqnarray}}
\newcommand{\ben}{\begin{eqnarray*}}
\newcommand{\een}{\end{eqnarray*}}

\allowdisplaybreaks
\allowdisplaybreaks[4]

\title{
\textbf{Evidence-Guided Selection of Retinal Vessel Segmentation Pipelines:}\\
\textbf{A Cross-Dataset Benchmark of Performance, Structure, Failure Modes, and Reproducibility}
}

\author[1]{Ruoyu Yan}
\author[1]{Yuejia Liu}
\author[1]{Shixin Xu}
\author[2]{Xiaoyan Li}
\author[3]{Xi Luo}
\author[3]{Yuhui Deng}
\author[1,2,3,4]{Huaxiong Huang\thanks{Corresponding author. Email: huaxiong.huang@dukekunshan.edu.cn}}

\affil[1]{Duke Kunshan University, Kunshan, China}
\affil[2]{University of Toronto, Toronto, Canada}
\affil[3]{Beijing Normal--Hong Kong Baptist University, Zhuhai, China}
\affil[4]{York University, Toronto, Canada}
 
\date{}

\begin{document}
\maketitle

\begin{abstract}
Retinal vessel segmentation is widely used in retinal image analysis, yet model selection based only on aggregate overlap scores can overlook structural fidelity, cross-dataset robustness, and application-specific error costs. We evaluated five executable binary vessel-segmentation pipelines---FSG-Net, SA-UNetv2, DCP-1024, GAVE, and FR-UNet---on 244 images from six public datasets using pixel-level agreement, cross-dataset stability, thin-vessel retention, structural metrics, error composition, and available computational evidence. Results were aggregated with equal weighting across datasets. No pipeline was optimal in every setting. FSG-Net achieved the highest equal-dataset mean Macro-F1 (0.77), IoU (0.63), clDice (0.83), and Skeleton F1 (0.78), whereas SA-UNetv2 achieved the highest mean sensitivity (0.85) and Thin-Vessel Sensitivity (0.63). Differences across datasets were often larger than those among the leading pipelines, and errors shared by all five methods were concentrated near annotated vessel boundaries and faint peripheral vessels. In an ancillary analysis on 45 HRF images, segmentation masks were converted into 11 ring-wise vascular features for leave-one-out classification. Controlled perturbations showed that local mask errors can be amplified in downstream features, while targeted stabilization reduced sensitivity for five features. The results support task-conditioned, benchmark-relative pipeline selection rather than a single universal ranking and highlight the need to validate the selected pipeline on the intended target domain.
\end{abstract}

\section{Introduction}\label{introduction}

\paragraph{Background and Model-Selection Context}

\begin{figure}[htbp]
    \centering
    \includegraphics[width=0.85\linewidth]{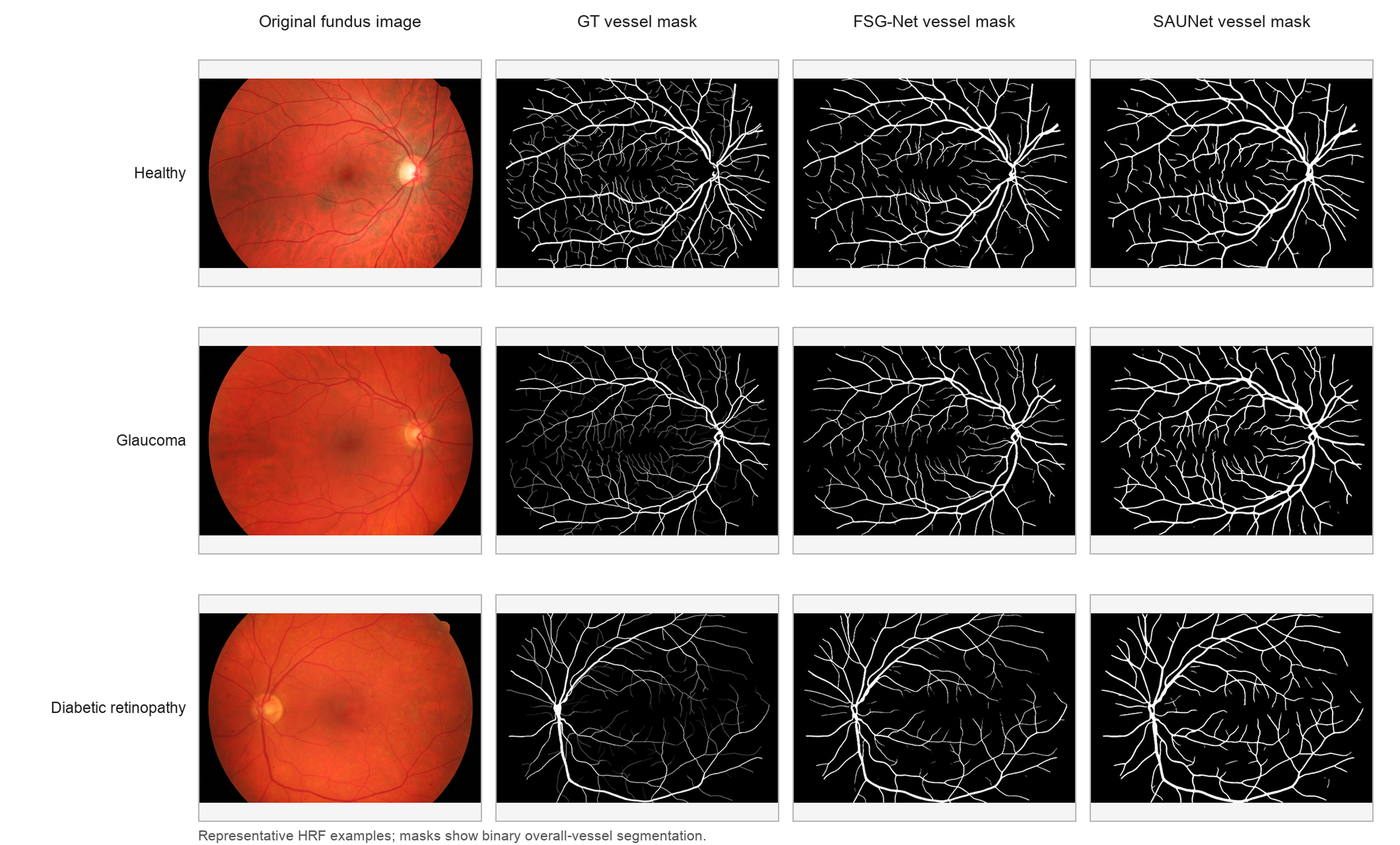}
    \caption{Representative HRF fundus images and pixel-level vessel segmentation results.}
    \label{fig:image1}
\end{figure}

Retinal vessel segmentation is a foundational component of computational fundus-image analysis, supporting vascular classification and other computer-aided retinal analyses. Reliable segmentation remains difficult because vessels occupy only a small fraction of the field of view. Variations in resolution, contrast, illumination, acquisition conditions, and pathology can also further alter vascular visibility and apparent morphology. These challenges can produce different residual errors, including missed vessels, false-positive responses, boundary shifts, and vessel-width errors. Because pipelines may differ in their ability to recover small branches, suppress false positives, or maintain connectivity, no single pipeline should be expected to perform best across all criteria \cite{fraz2012survey,galdran2022minimalistic,khandouzi2022review,qin2024review,lin2021high}. Figure~\ref{fig:image1} presents several representative HRF images and predictions from FSG-Net and SA-UNetv2. These results often do not reveal all the differences among different methods and their impact on downstream applications. For example, some tasks prioritize overall agreement, whereas others may require better thin-vessel recovery or structural preservation. A pipeline preferred under one objective may be unsuitable under another. Retinal vessel segmentation is therefore framed as a task-conditioned pipeline-selection problem.

\paragraph{Existing Evaluation and Review Gaps and Motivation}
 
Conventional model comparisons often emphasize pixel-level performance. However, these metrics may ignore structurally important errors, including vessel disconnections and centerline inconsistencies \cite{shit2021cldice}. 

Previous reviews have summarized retinal vessel segmentation methods, datasets, preprocessing, evaluation metrics, generalization, and clinical applications \cite{fraz2012survey,mookiah2021review,khandouzi2022review,qin2024review,bansal2026review}. However, direct comparison across studies remains difficult because of differences in datasets, preprocessing, thresholds, implementations, and evaluation protocols. For example, recent CMPB fundus-vessel studies evaluated overlapping benchmark datasets under method-specific preprocessing and operating thresholds \cite{gegundez2021new}. Reported metric sets also differ across studies; for example, DilUnet emphasizes Accuracy, Sensitivity, and IoU on DRIVE, STARE, and CHASE \cite{hussain2022dilunet}. Cross-dataset studies have also shown substantial performance degradation under domain shift  \cite{galdran2022minimalistic}. This evidence has not yet been translated into practical pipeline-selection guidance tailored to different segmentation requirements.

\paragraph{ Study Scope and Contributions}

To address this gap, this study evaluates executable pipelines rather than only comparing literature-reported scores. A shared cross-dataset benchmark combines overall segmentation quality, asymmetric error costs, thin-vessel retention, structural preservation, cross-dataset stability, and computational evidence. These evidence dimensions are aggregated with equal dataset weighting and later used in the decision framework described in Section~\ref{evidence-guided-model-selection}. This benchmark compares executable pipelines under source-compatible operating conditions; its results should not be interpreted as architecture-only effects, because training exposure and implementation provenance differ across pipelines.

As an ancillary analysis, we examine how segmentation differences propagate into ring-wise vascular features and downstream HRF classification. The experiment traces the transformation from vessel masks to interpretable centerline-, caliber-, and graph-based features and evaluates whether targeted feature stabilization reduces sensitivity to local mask errors.

The benchmark covers six evaluation datasets and five executable retinal vessel segmentation pipelines, comprising 244 images and 1,220 per-image predictions across 30 model-dataset combinations. The study is restricted to binary overall-vessel segmentation; artery--vein segmentation is not included because artery--vein reference labels and compatible outputs are not consistently available across the evaluated datasets and pipelines, whereas downstream classification is examined only as an ancillary HRF analysis.

Dataset-level results were aggregated with equal weighting across the six datasets. The same decision policy is implemented in a public browser-based tool for inspecting evidence dimensions, scenario-specific rankings, and application-specific weights. Despite some limitations, our study provides a reproducible starting point for researchers who need to use a segmentation method for their retina image-based applications. 

The rest of the paper is organized as follows. In Section~\ref{evaluation-datasets-and-included-pipelines}, we present the selection criteria for the datasets and segmentation pipelines, with more technical details given in Sections~\ref{pipeline-realization-and-evidence-qualification} and~\ref{evaluation-metrics-and-interpretation}. Evidence-guided model selection is presented in Section~\ref{evidence-guided-model-selection}, while the effect of datasets is discussed in Section~\ref{dataset-effects-and-domain-dependence}. Section~\ref{feature-level-evaluation-for-downstream-classification} presents a case study of application-oriented evaluation and selection of segmentation models. And Section~\ref{discussion-limitations-and-conclusions}
discusses the study limitations, future directions, and conclusions. Technical details are given in various appendices. \section{Evaluation Datasets and Included Pipelines}\label{evaluation-datasets-and-included-pipelines}

This section introduces the datasets and segmentation pipelines included in the benchmark and explains the main criteria used for their selection.

\subsection{Evaluation Datasets and Source Traceability}\label{evaluation-datasets-and-source-traceability}

The benchmark includes 244 image-mask pairs from six public evaluation datasets: AFIO, CHASEDB1, DualModal2019, HRF, LES, and STARE. Dataset selection required publicly available fundus images with corresponding vessel ground-truth masks, which could be used for binary overall-vessel evaluation. The datasets also differ in image resolution, acquisition conditions, pathology composition, and annotation procedures, providing heterogeneous domains for cross-dataset evaluation \cite{galdran2022minimalistic}.

Datasets with training exposure were excluded from evaluation. FIVES was used to train FSG-Net, FR-UNet, and SA-UNetv2 locally and was also included in the development of DCP-1024 \cite{jin2022fives,wei2025dcp}. DRIVE was also excluded because the GAVE output in this benchmark was generated with a checkpoint trained on DRIVE (Appendix~\ref{appendix-a.-source-local-and-released-configuration-boundaries}, Table~\ref{tab:training_exposure_ledger}).

Together, the six datasets and their original sources are summarized in Table~\ref{tab:datasets}. Their supplied ground-truth masks are used as reference annotations throughout the benchmark, while possible annotation and boundary uncertainty is considered separately in Section~\ref{discussion-limitations-and-conclusions}
\cite{lyu2022reta}.

\begin{table}
\centering
\caption{Evaluation datasets used in the benchmark. 
$n$ denotes the number of retained image--mask pairs. 
DualModal2019 was evaluated using the RGB image--mask pair.}
\label{tab:datasets}
\begin{tabular}{llll}
\toprule
Dataset & $n$ & Resolution & Source \\
\midrule
AFIO          & 99 & $1504\times1000$ & \cite{akram2020data} \\
CHASEDB1      & 28 & $999\times960$   & \cite{fraz2012ensemble} \\
DualModal2019 & 30 & $1024\times1024$ & \cite{zhang2019dualmodal} \\
HRF           & 45 & $3504\times2336$ & \cite{budai2013robust} \\
LES           & 22 & $1620\times1444^{*}$ & \cite{orlando2018glaucoma} \\
STARE         & 20 & $700\times605$   & \cite{hoover2000locating} \\
\bottomrule
\end{tabular}

\begin{flushleft}
$^{*}$Twenty-one LES images have resolution $1620\times1444$; one image has resolution $2196\times1958$.
\end{flushleft}
\end{table}

\subsection{Included Pipelines and Source Traceability}\label{included-pipelines-and-source-traceability}

The benchmark was chosen intentionally. Eligible pipelines needed to generate binary vessel masks through an executable evaluation pathway. They also had to support batch prediction on evaluation datasets and provide a clearly defined execution flow to enable consistent evaluation. This analysis excluded methods that require image-specific ground truth or manually entered prompts. For instance, MedSAM does not meet these criteria in its standard workflow because it relies on user-provided prompts instead of directly producing fully automatic vessel masks. Its original study also notes difficulties with branching vessel-like structures under bounding-box prompting \cite{ma2024medsam}.  The selected pipelines represent methods that could be consistently evaluated under the present benchmark and should not be interpreted as a comprehensive survey of the field.

Five pipelines met these requirements. FSG-Net, FR-UNet, and SA-UNetv2 provided reproducible training and inference workflows and thus were retrained locally on FIVES \cite{jin2022fives,liu2022frunet,seo2025fsgnet,guo2026saunetv2}. DCP-1024 and GAVE were evaluated using released checkpoints because reproducible training workflows were unavailable \cite{dai2024ripav,wei2025dcp}. Table~\ref{tab:pipeline_traceability} summarizes the evaluated outputs, source resources, and benchmark route for each pipeline.

\begin{table}
\caption{Included pipelines, source provenance, and reproducible benchmark routes.}
\label{tab:pipeline_traceability}
\scriptsize
\begin{tabularx}{\textwidth}{
@{}
>{\raggedright\arraybackslash}p{0.12\textwidth}
>{\raggedright\arraybackslash}p{0.12\textwidth}
>{\raggedright\arraybackslash}p{0.44\textwidth}
>{\raggedright\arraybackslash}X
@{}
}
\toprule
\textbf{Benchmark pipeline} &
\textbf{Source method} &
\textbf{Source publication} &
\textbf{Public benchmark route} \\
\midrule

FSG-Net &
FSG-Net &
\href{https://doi.org/10.1186/s12880-025-02021-4}
{\textit{Full-scale Representation Guided Network for Retinal Vessel Segmentation}} &
\href{https://github.com/ZombaSY/FSG-Net-pytorch}
{Official training and inference code} \\

\addlinespace

FR-UNet &
FR-UNet + DTI\textsuperscript{1} &
\href{https://doi.org/10.1109/JBHI.2022.3188710}
{\textit{Full-Resolution Network and Dual-Threshold Iteration for Retinal Vessel and Coronary Angiograph Segmentation}} &
\href{https://github.com/lseventeen/FR-UNet}
{Official training and inference code} \\

\addlinespace

SA-UNetv2 &
SA-UNetv2 &
\href{https://doi.org/10.1109/ISBI61048.2026.11515447}
{\textit{SA-UNetv2: Rethinking Spatial Attention U-Net for Retinal Vessel Segmentation}} &
\href{https://github.com/clguo/SA-UNetv2}
{Official training and inference code} \\

\addlinespace

DCP-1024 &
DCP &
\href{https://doi.org/10.1109/ICASSP49660.2025.10888001}
{\textit{Convolutional Prompting for Broad-Domain Retinal Vessel Segmentation}} &
\href{https://huggingface.co/AIMClab-RUC/UNet_DCP_1024}
{Released 1024-resolution checkpoint and inference route} \\

\addlinespace

GAVE &
RIP-AV &
\href{https://papers.miccai.org/miccai-2024/660-Paper1711.html}
{\textit{RIP-AV: Joint Representative Instance Pre-training with Context-Aware Network for Retinal Artery/Vein Segmentation}} &
\href{https://github.com/weidai00/RIP-AV}
{Released checkpoint and inference code} \\

\bottomrule
\end{tabularx}

\vspace{0.5em}
\begin{minipage}{\textwidth}
\textsuperscript{1}DTI denotes the dual-threshold iteration procedure proposed with FR-UNet.
Complete paper, code, checkpoint, dataset, license, and access-date records are provided in the public provenance index.
\end{minipage}
\end{table}

\section{Pipeline Realization and Evidence Qualification}\label{pipeline-realization-and-evidence-qualification}

This section describes how the pipelines included were implemented in the benchmark. It summarizes their operating points, configuration sensitivity, and available computational evidence.

\subsection{Original Design and Source-Paper Operating Procedures}\label{original-design-and-source-paper-operating-procedures}

The pipelines included were developed under different architectural, training, preprocessing, and inference settings. This section summarizes the main design characteristics for each pipeline. Detailed settings reported in the source papers and released implementation settings are provided in Appendix~\ref{appendix-a.-source-local-and-released-configuration-boundaries}.

\subsection{Benchmark Threshold Policy and Operating Points}\label{benchmark-threshold-policy-and-operating-points}

Consistent with this source-preserving approach, the probability-to-mask conversion threshold used in the benchmark followed the source implementation if it was specified. FSG-Net used its reported threshold of 0.50, while FR-UNet retained dual-threshold iteration. GAVE followed the latest released implementation: the combined-vessel mask used a primary threshold of 0.35, supplemented by skeleton predictions at thresholds of 0.10, 0.20, and 0.25. For SA-UNetv2 and DCP-1024, whose binary thresholds were not specified in the original sources, a default threshold of 0.50 was applied.

The sensitivity of the available probability outputs to alternative thresholds was examined in Appendix~\ref{appendix-b.-threshold-sensitivity-and-benchmark-operating-points}. This analysis was used only as a robustness check and did not replace the retained benchmark operating points. FSG-Net, SA-UNetv2, and DCP-1024 showed limited variation across the tested thresholds, while GAVE was extremely sensitive to threshold changes. The main model-selection conclusions remained unaffected.

\subsection{Configuration Sensitivity and Selection Stability}\label{configuration-sensitivity-and-selection-stability}

Configuration sensitivity was examined for the three locally trained pipelines by adjusting several model-specific hyperparameters as a bounded robustness check. Across the tested settings, the principal recommendation pattern remained stable. The two leading pipelines, FSG-Net and SA-UNetv2, showed only limited changes in their cross-dataset performance within the examined ranges, and their overall recommendation roles haven't been affected. FR-UNet showed greater configuration sensitivity and could achieve noticeable improvement under some settings, but these changes were not sufficient to surpass the leading pipeline. The analysis therefore supports the robustness of the principal model-selection conclusions to the tested configuration changes. Detailed configurations and dataset-level results are reported in Appendix~\ref{appendix-c.-configuration-sensitivity-and-selection-stability}.

\subsection{Local Training and Computational Evidence}\label{local-training-and-computational-evidence}

The available computational and training evidence is summarized in
Table~\ref{tab:computational_evidence}.
FSG-Net, FR-UNet, and SA-UNetv2 were locally trained on FIVES \cite{jin2022fives}. Model-specific input sizes, preprocessing, precision modes, schedules, and splits are retained in the resource table below; the full source-paper operating details are provided in Appendix~\ref{appendix-a.-source-local-and-released-configuration-boundaries}.

\begin{table}
\caption{Available computational-cost and local-training evidence for the evaluated pipelines.}
\label{tab:computational_evidence}
\scriptsize
\begin{tabularx}{\textwidth}{
@{}
>{\raggedright\arraybackslash}p{0.11\textwidth}
>{\raggedright\arraybackslash}p{0.12\textwidth}
>{\raggedright\arraybackslash}p{0.15\textwidth}
>{\raggedright\arraybackslash}p{0.18\textwidth}
>{\raggedright\arraybackslash}p{0.10\textwidth}
>{\raggedright\arraybackslash}X
@{}
}
\toprule
\textbf{Pipeline} &
\textbf{Evidence source} &
\textbf{Local setup} &
\textbf{Model / compute} &
\textbf{Best val. ckpt} &
\textbf{Memory / training time} \\
\midrule

FR-UNet &
FIVES local training + profiling &
$512\times512$; BS=4; AMP &
7.38 M params; 430.77 GFLOPs/img; 28.32 MiB ckpt &
Epoch 47; val. F1=0.93 &
Peak VRAM: 7.08 GB; median 44.57 s/epoch; best ckpt: 35.10 min; total: 52.16 min \\

\addlinespace

SA-UNetv2 &
FIVES local log + profiling &
$1028\times1028$; BS=2 &
0.26 M params; 78.18 GFLOPs/img; 3.24 MiB ckpt &
Epoch 55; val. F1=0.93 &
Peak VRAM: 13.18 GB; mean 13.45 s/epoch (75 epochs)\textsuperscript{2}; best-ckpt time: N/A; total: 16.85 min \\

\addlinespace

FSG-Net &
FIVES local training + profiling &
$288\times288$; BS=4; FP32 &
16.85 M params; 199.61 GFLOPs/img; 64.45 MiB ckpt &
Epoch 181; val. F1=0.90 &
Peak VRAM: 7.94 GB; median 34.91 s/epoch; best ckpt: 105.37 min; total: 116.42 min \\

\addlinespace

DCP-1024 &
Released ckpt + published complexity &
$1024\times1024$ local inference; no local training &
12.50 M params\textsuperscript{1}; FLOPs/ckpt not locally profiled &
N/A &
Training VRAM/time: N/A \\

\addlinespace

GAVE &
Released ckpt &
Overlapping $256\times256$ patch inference; no local training &
Params/FLOPs/ckpt not locally profiled &
N/A &
Training VRAM/time: N/A \\

\bottomrule
\end{tabularx}

\vspace{0.5em}
\begin{minipage}{\textwidth}
\textsuperscript{1}Published parameter count for DCP-1024; not independently re-counted from the released implementation.
\textsuperscript{2}SA-UNetv2 epoch time is the mean across 75 recorded epochs.
AMP, automatic mixed precision; BS, batch size; FP32, 32-bit floating-point precision; GFLOPs, billion floating-point operations.
\end{minipage}
\end{table}

Among the three locally trained pipelines, SA-UNetv2 had the smallest model scale and lowest profiled forward-pass cost, whereas FR-UNet had the lowest recorded peak training VRAM. These results indicate that model size, arithmetic cost, and training-memory demand can be used as separate resource dimensions for computing cost ranking.

However, resource comparisons remain incomplete because DCP-1024 and GAVE were only evaluated from released checkpoints without corresponding local training runs, and several resource fields were unavailable. Accordingly, resource-oriented comparisons are considered provisional.

\section{Evaluation Metrics and Interpretation}\label{evaluation-metrics-and-interpretation}

The benchmark uses complementary pixel-level, thin-vessel, and structural metrics to evaluate segmentation quality from different perspectives. Table~\ref{tab:metric_overview} gives a concise interpretation of the principal metrics used below. Formal mathematical definitions, implementation settings, and aggregation rules are provided in Appendix~\ref{appendix-d.-metric-definitions-and-cross-dataset-aggregation-rules}.

\begin{table}
\caption{Overview of the principal evaluation metrics used in the benchmark.}
\label{tab:metric_overview}
\footnotesize
\begin{tabularx}{\textwidth}{
@{}
>{\raggedright\arraybackslash}p{0.20\textwidth}
>{\raggedright\arraybackslash}p{0.38\textwidth}
>{\raggedright\arraybackslash}X
@{}
}
\toprule
\textbf{Metric} & \textbf{What it measures} & \textbf{Interpretation} \\
\midrule
Accuracy & Fraction of correctly classified pixels. & Can be misleading because retinal vessel segmentation is strongly class-imbalanced. \\
F1 / \(F_\beta\) & Balance between vessel precision and sensitivity. & F1 gives equal weight to precision and sensitivity; \(F_\beta\) allows different FN--FP priorities. Image-level F1 is the standard pixel-level F1 computed separately for each image, and Macro-F1 summarizes these image-level values under the benchmark aggregation procedure. \\
Sensitivity & Fraction of reference vessel pixels recovered. & Higher values mean fewer false negatives. \\
Precision & Fraction of predicted vessel pixels that are correct. & Higher values mean fewer false positives. \\
IoU & Overlap between predicted and reference vessel regions. & Higher values indicate stronger region-level agreement. \\
MCC & Agreement considering all four confusion-matrix components. & Useful under class imbalance because true positives, true negatives, false positives, and false negatives all contribute. \\
Thin-Vessel Sensitivity (TVS) & Recovery of ground-truth-defined thin vessels. & Higher values indicate better retention of fine vascular branches. \\
clDice & Bidirectional centerline-to-region agreement. & Higher values indicate better preservation of vascular centerline structure. \\
Skeleton F1 (SF1) & Tolerance-based agreement between predicted and reference skeletons. & Allows small localization offsets when assessing skeleton alignment. \\
\bottomrule
\end{tabularx}
\end{table}

\subsection{Limitations of Pixel-Based Evaluation}\label{limitations-of-pixel-based-evaluation}

\begin{figure}[htbp]
    \centering
    \includegraphics[width=0.75\linewidth]{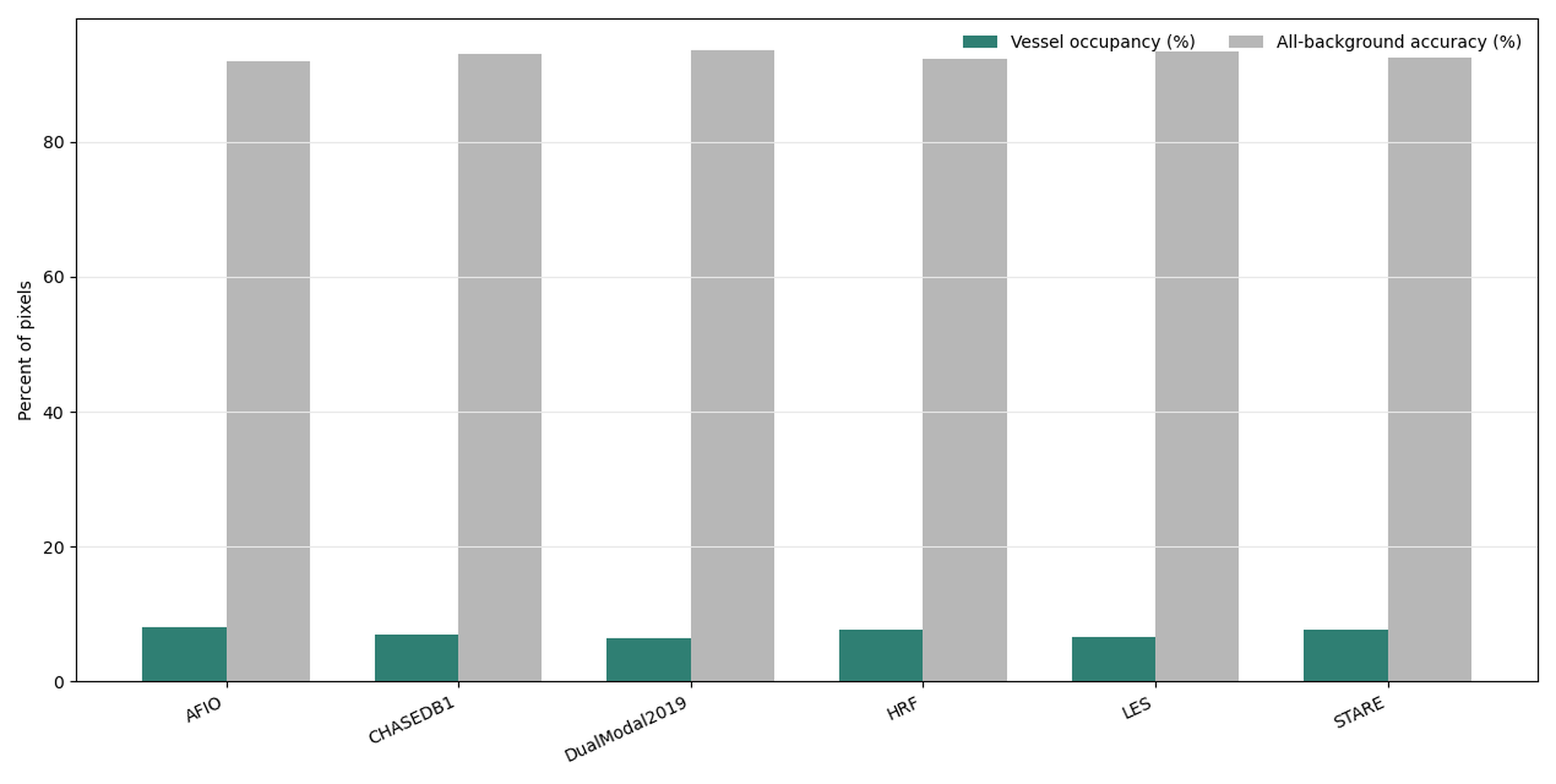}
    \caption{Dataset-level vessel occupancy and the apparent Accuracy of an all-background prediction.}
    \label{fig:image2}
\end{figure}

Vessel pixels occupy only a small fraction of the field of view, making Accuracy potentially misleading under severe class imbalance \cite{bansal2026review,kovacs2022baseline}. Let \(G_{d}\) and \(\text{FOV}_{d}\) denote the pooled ground-truth vessel pixels and valid field-of-view pixels, respectively, across all images in dataset \(d\). Vessel occupancy was calculated as
\begin{equation}
O_{d} = \frac{\mid G_{d} \mid}{\mid \text{FOV}_{d} \mid},
\end{equation}
\noindent
and an all-background prediction therefore achieves an Accuracy of \(1 - O_{d}\). Figure~\ref{fig:image2} compares vessel occupancy with this apparent all-background Accuracy across the six datasets.

A prediction can therefore achieve high Accuracy by classifying background correctly even when it detects no vessels. Figure~\ref{fig:image2} illustrates this class-imbalance failure at the dataset level. Accuracy is therefore reported only as contextual information and is not used as a primary model-selection criterion.

\begin{figure}
    \centering
    \includegraphics[width=0.70\linewidth]{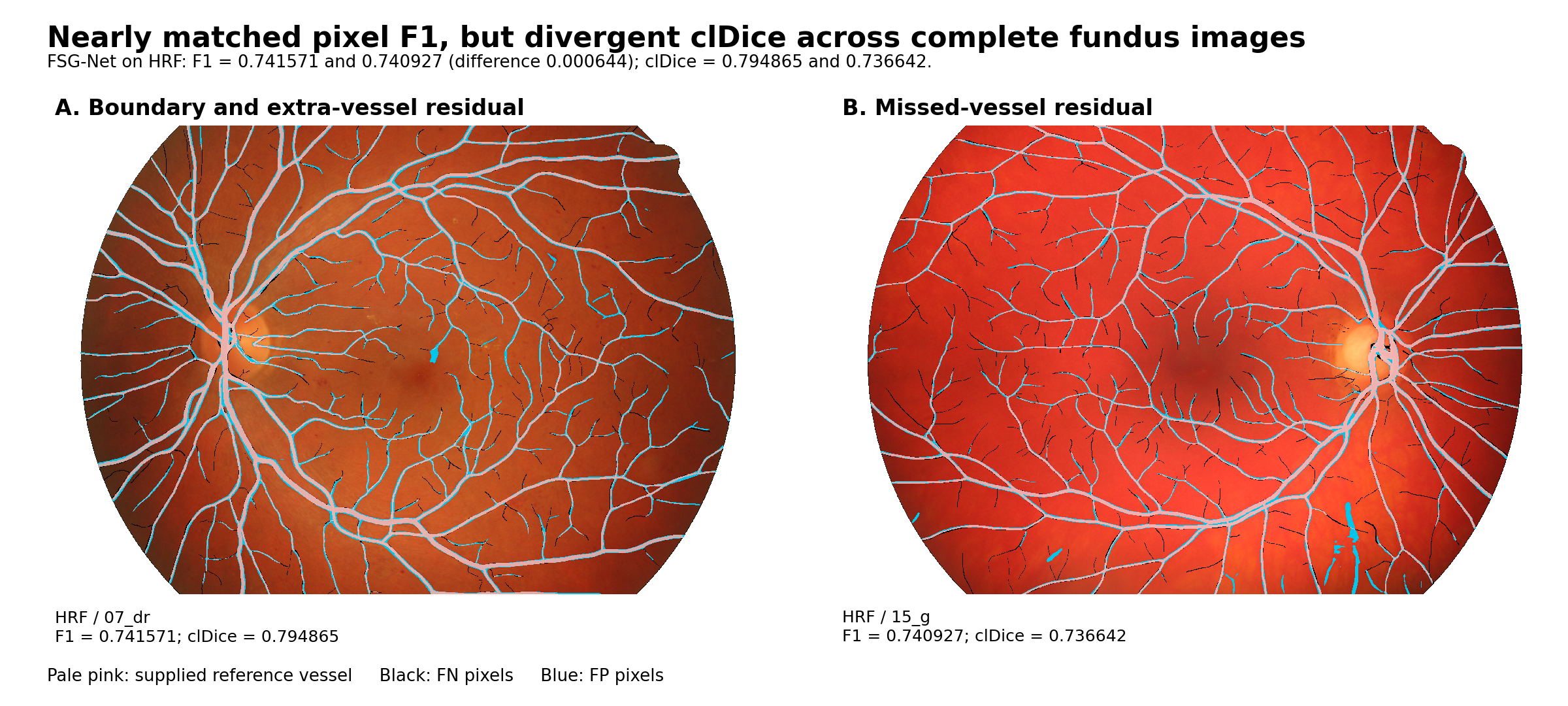}
    \caption{Nearly matched pixel-level F1 values with divergent structural fidelity and spatial error profiles. White denotes the supplied reference vessel, black denotes false-negative (FN) pixels, and blue denotes false-positive (FP) pixels.}
    \label{fig:image3}
\end{figure}

The limitation of pixel-based evaluation is not restricted to Accuracy. Pixel-level metrics such as F1 and IoU measure agreement with the reference mask but do not identify the structural consequences of the errors. Figure~\ref{fig:image3} compares two HRF images, identified in the dataset as 07\_dr and 15\_g, using FSG-Net predictions. The two cases have nearly matched F1 values but substantially different clDice values and spatial error patterns.

Although the two HRF images \cite{budai2013robust}, 07\_dr and 15\_g, have nearly 
matched F1 values (0.741571 for 07\_dr and 0.740927 for 15\_g), their spatial error 
patterns differ visibly in Figure~\ref{fig:image3}. Image 07\_dr is from the diabetic 
retinopathy subset, whereas 15\_g is from the glaucoma subset. In 07\_dr, many residual 
errors are concentrated along vessel boundaries, while the main vascular centerlines 
remain largely preserved. In 15\_g, false-negative (FN) errors are more widely distributed 
across thin and peripheral vessel segments, resulting in greater loss of reference-vessel 
structure. Consistent with this difference, structural fidelity is lower for 15\_g than 
for 07\_dr (0.736642 versus 0.794865), despite their nearly identical pixel-level metric 
values.

Pixel-level metrics are therefore necessary but not sufficient for model evaluation. These examples motivate the multidimensional evaluation framework defined in Section~\ref{required-multi-dimensional-task-conditioned-evidence}.

\subsection{Required Multi-Dimensional, Task-Conditioned Evidence}\label{required-multi-dimensional-task-conditioned-evidence}

Accordingly, this study organizes the evaluation evidence into four complementary dimensions: pixel-level performance, cross-dataset robustness, thin-vessel retention, and structural fidelity. Table~\ref{tab:evidence_dimensions} summarizes how the individual metrics contribute to these dimensions and their roles in model selection.

\begin{table}
\caption{Four evidence dimensions used for task-conditioned model selection.}
\label{tab:evidence_dimensions}
\scriptsize
\begin{tabularx}{\textwidth}{
@{}
>{\raggedright\arraybackslash}p{0.12\textwidth}
>{\raggedright\arraybackslash}p{0.27\textwidth}
>{\raggedright\arraybackslash}X
@{}
}
\toprule
\textbf{Evidence dimension} &
\textbf{Measures} &
\textbf{Role in model selection} \\
\midrule

Pixel-level performance &
F1/$F_{\beta}$, sensitivity, precision, IoU, and MCC &
Characterizes overall pixel-wise agreement; supports selection for overall performance, low-FN, low-FP, or balanced objectives. \\

\addlinespace

Cross-dataset robustness &
Equal-dataset mean $F_{\beta}$, cross-dataset standard deviation, and worst-dataset performance &
Quantifies expected performance, cross-dataset variability, and downside risk under domain uncertainty. \\

\addlinespace

Thin-vessel retention &
Thin-Vessel Sensitivity (TVS) &
Measures recovery of vessels and supports selection when preserving thin vessels or reducing missed branches is a priority. \\

\addlinespace

Structural fidelity &
clDice and tolerant Skeleton F1 (SF1) &
Assesses centerline containment and tolerant skeleton alignment; supports selection when focusing on vascular continuity and structural fidelity. \\

\bottomrule
\end{tabularx}
\end{table}

\subsection{Evaluation Implementation and Aggregation Principles}\label{evaluation-implementation-and-aggregation-principles}

All metrics were calculated within the valid evaluation region from binarized predictions and reference masks. Pixel-level evidence included F1, sensitivity, precision, IoU, and MCC. TVS assessed recovery of ground-truth-defined thin branches. For structural fidelity, clDice and tolerant Skeleton F1 assessed centerline preservation and alignment. Image-level values were averaged within each model-dataset pair. The six datasets were then given equal weight. Cross-dataset SD and worst-dataset performance described cross-dataset robustness. Full definitions, equations, and benchmark settings are provided in Appendix~\ref{appendix-d.-metric-definitions-and-cross-dataset-aggregation-rules}.

\section{Evidence-Guided Model Selection}\label{evidence-guided-model-selection}

This section converts the evidence from Section~\ref{evaluation-metrics-and-interpretation} into model recommendations for different application priorities. The framework returns either a primary pipeline or a two-pipeline shortlist.

\subsection{Decision-Scoring Framework and Scenario Presets}\label{decision-scoring-framework-and-scenario-presets}

The decision framework combines pixel-level performance, thin-vessel retention, structural fidelity, and available computational evidence into task-conditioned scores. Residual-error composition is used separately for post-ranking diagnostic interpretation rather than as a scoring component.

\subsubsection{Evidence Utilities and Score Construction}\label{evidence-utilities-and-score-construction}

The decision score uses selected metrics from Section~\ref{evaluation-implementation-and-aggregation-principles} rather than all reported evaluation measures. In the notation below, \(m\), \(d\), and \(i\) index model, dataset, and image, respectively. The superscripts \(P\), \(T\), \(S\), and \(R\) denote pixel-level, thin-vessel, structural, and resource utilities. Pixel utility was represented by \(F_{\beta}\), thin-vessel utility by TVS, and structural utility by a weighted combination of clDice and tolerant Skeleton F1:
\begin{equation}
\begin{aligned}
U_{\mathrm{mdi}}^{P} &= F_{\beta,\mathrm{mdi}}, \\
U_{\mathrm{mdi}}^{T} &= \mathrm{TVS}_{\mathrm{mdi}}, \\
U_{\mathrm{mdi}}^{S} &=
\eta\,\mathrm{clDice}_{\mathrm{mdi}}
+(1-\eta)\,\mathrm{SF1}_{\mathrm{mdi}} .
\end{aligned}
\end{equation}
where \(\beta\) controls the FN--FP preference and \(\eta=0.80\) in the predefined scenarios.

The value $\eta=0.80$  was a prespecified analytic preference, not a value estimated from clinical outcomes or adopted as a universal literature standard. It gives clDice the primary role while retaining the $r = 2$ tolerant Skeleton F1 as a secondary check on small localization offsets. Its influence on the ranking was tested in Appendix~\ref{appendix-e.-decision-parameter-sensitivity-analysis}, Figure~\ref{fig:image15}.

Following the aggregation procedure defined in Section~\ref{evaluation-implementation-and-aggregation-principles}, the dataset-level utilities were combined using equal dataset weighting. An optional worst-dataset term was added to reflect downside risk.

\begin{equation}
{\overline{U}}_{m}^{q} = \frac{1}{D}\sum_{d = 1}^{D}U_{\text{md}}^{q},
\qquad q\in\{P,T,S\},
\end{equation}
where \(D\) is the number of evaluation datasets. The risk-adjusted utility is then
\begin{equation}
{\widetilde{U}}_{m}^{q}
=
\left(1-\rho\right){\overline{U}}_{m}^{q}
+\rho\min_{d}U_{\text{md}}^{q},
\qquad q\in\{P,T,S\}.
\end{equation}

The cross-dataset risk parameter \(\rho\) was set to 0.00, 0.30, or 0.70. The first provides a mean-only baseline, while 0.30 and 0.70 were chosen symmetrically around 0.50 to create two clearly distinct 70/30 aggregation regimes: one dominated by mean performance and the other by worst-dataset performance. This provides three interpretable risk levels without implying that the selected values are empirically optimal. Sensitivity to this discretization was examined across the full \(\rho\) range in Appendix~\ref{e.2-sensitivity-to-the-cross-dataset-risk-weight-ux3c1}.

For resource-related scenarios, resource utility was calculated from the pipeline/model-level computational evidence introduced in Section~\ref{local-training-and-computational-evidence}: the available parameter count, GFLOPs, checkpoint size, peak allocated VRAM, total training time, and epoch time:

\begin{equation}
U_{m}^{R} = \frac{1}{K}\sum_{k = 1}^{K}u_{\text{mk}},
\qquad
u_{\text{mk}} = \frac{\min_{j}C_{\text{jk}}}{C_{\text{mk}}},
\end{equation}
where \(K\) is the number of recorded resource fields, \(C_{\text{mk}}\) is the recorded cost of model \(m\) for resource field \(k\), and \(j\) ranges over the models with available values for that field. Missing resource fields received a neutral utility of 0.50 for score display, but pipelines with incomplete resource coverage were not assigned a formal second-place resource recommendation.
In this term, \(u_{\text{mk}}\) denotes the normalized resource utility of model \(m\) for resource field \(k\). The benchmark used up to six resource fields (\(K=6\)): parameter count, GFLOPs, checkpoint size, peak training VRAM, total training time, and epoch time. The superscript \(R\) denotes the aggregate resource utility.

The final decision score was

\begin{equation}
S_{m} = 100\left( w_{P}{\widetilde{U}}_{m}^{P} + w_{T}{\widetilde{U}}_{m}^{T} + w_{S}{\widetilde{U}}_{m}^{S} + w_{R}U_{m}^{R} \right),
\end{equation}
with \(w_{P} + w_{T} + w_{S} + w_{R} = 1.\)

\subsubsection{Scenario Presets and Benchmark-Relative Recommendations}\label{scenario-presets-and-benchmark-relative-recommendations}

Table~\ref{tab:preset_decision_rules} illustrates the decision framework using one mixed-objective default and five task-focused presets. The preset weights and parameters were manually specified before scenario scoring, rather than estimated from the evaluation datasets. They were chosen to represent interpretable use cases. The mixed-objective default assigns the largest weight to pixel-level performance while retaining thin-vessel, structural, and resource evidence. The pixel-level mean preset sets all weight on pixel performance and uses \(\rho=0.00\) to show the mean-performance baseline. The low-FN/thin-vessel preset uses \(\beta=2\) and equal pixel/thin-vessel weights to emphasize missed-vessel reduction. The structural-fidelity preset assigns all weight to clDice/SF1-based structure preservation. The unknown-domain preset keeps the pixel-performance objective but increases \(\rho\) to 0.70 to emphasize worst-dataset robustness. The resource-oriented preset gives the largest weight to resource utility while retaining a smaller pixel-performance term so that computational efficiency is not evaluated entirely apart from segmentation quality. These settings provide convenient starting points rather than fixed decision rules; users can modify the four evidence weights and choose the cross-dataset risk parameter \(\rho\) in the public tool.

\begin{table}[H]
\caption{Preset decision rules implemented in the public model-selection tool.
Weights are reported as pixel/thin-vessel/structure/resource percentages.}
\label{tab:preset_decision_rules}
\scriptsize
\begin{tabularx}{\textwidth}{
@{}
>{\raggedright\arraybackslash}p{0.17\textwidth}
c
c
c
>{\raggedright\arraybackslash}X
>{\raggedright\arraybackslash}p{0.16\textwidth}
>{\raggedright\arraybackslash}p{0.14\textwidth}
@{}
}
\toprule
\textbf{Scenario} &
$\boldsymbol{\beta}$ &
\textbf{Weights (P/T/S/R)} &
$\boldsymbol{\rho}$ &
\textbf{First / second choice} &
\textbf{Evidence gap} &
\textbf{Decision} \\
\midrule

Mixed-objective default &
1 &
50/15/20/15 &
0.30 &
SA-UNetv2 / FSG-Net &
75.82 vs.\ 66.76 ($+9.06$) &
Provisional \\

\addlinespace
Pixel-level mean &
1 &
100/0/0/0 &
0.00 &
FSG-Net / DCP-1024 &
77.09 vs.\ 76.40 ($+0.69$) &
Two-model shortlist \\

\addlinespace
Low FN / thin vessels &
2 &
50/50/0/0 &
0.30 &
SA-UNetv2 / FSG-Net &
68.01 vs.\ 66.99 ($+1.02$) &
SA-UNetv2 preferred \\

\addlinespace
Structural fidelity &
1 &
0/0/100/0 &
0.30 &
FSG-Net / SA-UNetv2 &
79.46 vs.\ 78.85 ($+0.61$) &
Two-model shortlist \\

\addlinespace
Unknown-domain robustness &
1 &
100/0/0/0 &
0.70 &
SA-UNetv2 / FSG-Net &
71.95 vs.\ 71.81 ($+0.15$) &
Two-model shortlist \\

\addlinespace
Resource-oriented &
1 &
30/0/0/70 &
0.30 &
SA-UNetv2 / insufficient evidence &
86.94; no valid second-place comparison &
Provisional \\

\bottomrule
\end{tabularx}
\end{table}

Under these representative settings, no pipeline ranked first across all objectives. But FSG-Net and SA-UNetv2 accounted for all first-ranked results. FSG-Net led the pixel-level and structural-fidelity settings, while SA-UNetv2 led the mixed-objective example, low-FN/thin-vessel, unknown-domain, and resource-oriented settings. DCP-1024 appeared as the second-ranked pipeline for pixel-level mean performance, while FR-UNet and GAVE did not enter the top two under these settings. These rankings illustrate how the benchmark evidence changes with the selected objective. The public tool allows the task weights and \(\rho\) to be adjusted directly.

\subsubsection{Public Decision Tool and Interpretation Boundary}
\label{public-decision-tool-and-interpretation-boundary}

\begin{figure}[htbp]
    \centering
    \setlength{\fboxsep}{3pt}
    \setlength{\fboxrule}{0.5pt}
    {%
        \includegraphics[
            width=0.95\linewidth,
            height=0.78\textheight,
            keepaspectratio
        ]{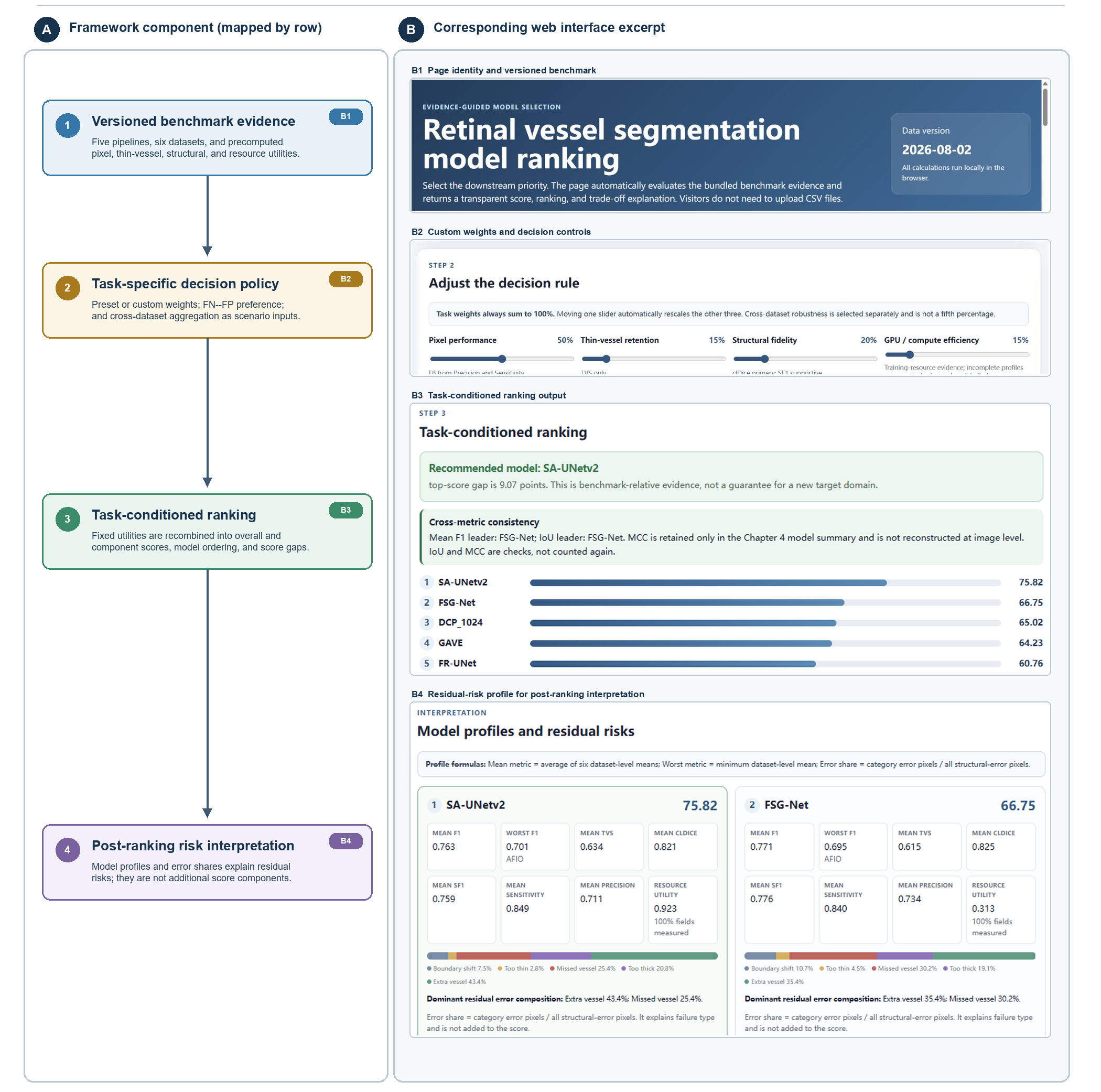}%
    }
    \caption{Public browser-based decision tool for task-conditioned retinal vessel segmentation pipeline selection. Each framework component in panel A corresponds by row to the interface excerpt in panel B: versioned benchmark evidence (B1), custom decision controls (B2), task-conditioned ranking output (B3), and post-ranking residual-risk profiles (B4). The tool recomputes scenario-specific rankings over fixed benchmark evidence; it does not imply a universal pipeline ranking.}
    \label{fig:public-decision-tool}
\end{figure}

The scoring policy is implemented in a browser-based decision tool that loads the versioned benchmark evidence and supports custom weights. It reports overall rankings, dimension-level utilities, and residual-risk profiles. Figure~\ref{fig:public-decision-tool} summarizes the correspondence between the decision-tool framework and its interface excerpts, including benchmark evidence, custom decision weights, task-conditioned rankings, and post-ranking residual-risk interpretation.

\begin{enumerate}
\item Interactive decision tool:
\href{https://ritaaaa-yry.github.io/retinal-segmentation-guidance/model-ranking-page/}
{\underline{retinal-segmentation-guidance/model-ranking-page}}

\item Evidence and source repository:
\href{https://github.com/ritaaaa-yry/retinal-segmentation-guidance}
{\underline{github.com/ritaaaa-yry/retinal-segmentation-guidance}}
\end{enumerate}

\subsection{Pixel-Level Performance and Cross-Dataset Robustness}\label{pixel-level-performance-and-cross-dataset-robustness}

This section examines the pixel-level evidence and gives targeted recommendations introduced in Section~\ref{decision-scoring-framework-and-scenario-presets}. F1 measures overall pixel agreement; Sensitivity and Precision expose the FN--FP operating profile; and IoU and MCC provide cross-metric consistency checks. Equal-dataset means, cross-dataset SDs, and worst-dataset results are jointly used to separate average performance from dataset-dependent risk.

\subsubsection{Overall Pixel Agreement and FN--FP Trade-offs}\label{overall-pixel-agreement-and-fnfp-trade-offs}

To establish the pixel-level baseline for the scenario recommendations, Table~\ref{tab:mean_pixel_performance} summarizes the equal-dataset mean performance of the five evaluated pipelines. The table focuses on overall pixel agreement and the Sensitivity--Precision trade-off.

These equal-dataset summaries are descriptive benchmark comparisons; no inferential claim is made that small observed gaps between leading pipelines are statistically significant across heterogeneous datasets.

\begin{table}[H]
\caption{Mean pixel-level segmentation performance across the evaluation datasets. 
Bold values indicate the best performance for each metric.}
\label{tab:mean_pixel_performance}
\scriptsize
\begin{tabularx}{\textwidth}{
@{}
l
>{\centering\arraybackslash}X
>{\centering\arraybackslash}X
>{\centering\arraybackslash}X
>{\centering\arraybackslash}X
>{\centering\arraybackslash}X
@{}
}
\toprule
\textbf{Model} &
\textbf{Macro-F1} &
\textbf{Sensitivity} &
\textbf{Precision} &
\textbf{MCC} &
\textbf{IoU} \\
\midrule

DCP-1024 &
0.76 &
0.79 &
\textbf{0.77} &
\textbf{0.76} &
0.62 \\

FR-UNet &
0.73 &
0.77 &
0.72 &
0.72 &
0.58 \\

FSG-Net &
\textbf{0.77} &
0.84 &
0.73 &
\textbf{0.76} &
\textbf{0.63} \\

GAVE &
0.73 &
0.80 &
0.70 &
0.72 &
0.58 \\

SA-UNetv2 &
0.76 &
\textbf{0.85} &
0.71 &
0.75 &
0.62 \\

\bottomrule
\end{tabularx}
\end{table}

FSG-Net achieved the highest equal-dataset Macro-F1 (0.77) and IoU (0.63), and tied DCP-1024 for the highest mean MCC (0.76). SA-UNetv2 and DCP-1024 both achieved equal-dataset Macro-F1 scores only 0.01 lower than FSG-Net. SA-UNetv2 achieved the highest mean Sensitivity (0.85), while DCP-1024 achieved the highest mean Precision (0.77). These results suggest selection of SA-UNetv2 when missed-vessel reduction receives greater priority, and DCP-1024 when false-positive suppression is emphasized.

\begin{figure}
    \centering
    \includegraphics[width=0.75\linewidth]{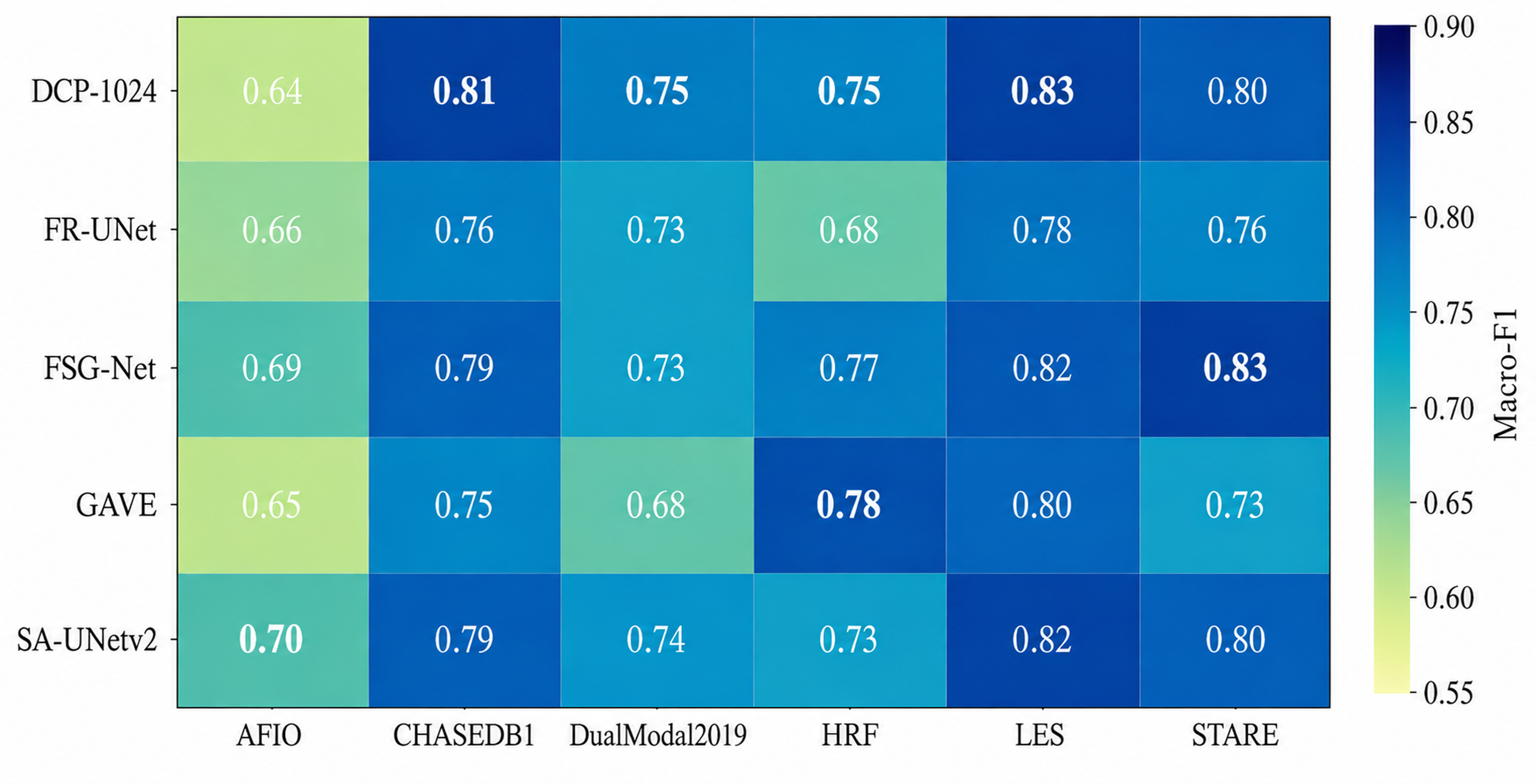}
    \caption{Dataset-level Macro-F1 across five evaluated pipelines and six evaluation datasets. Bold values indicate the highest Macro-F1 within each dataset.}
    \label{fig:image4}
\end{figure}

Equal dataset means provide an overall ranking but do not show whether the same pipeline leads on every dataset. Figure~\ref{fig:image4} presents the detailed dataset-level Macro-F1 values for each model--dataset pair.

The dataset-level leader varied across domains, confirming that the best-performing pipeline was dataset dependent. Although FSG-Net did not rank first on every dataset, it remained consistently competitive and was never the lowest-performing pipeline across the six evaluated domains. This supports its use as a stable cross-dataset candidate for pixel-level performance, while not implying that it must be optimal on an unseen target domain.

\subsubsection{Relative Dataset Response and Robustness}
\label{relative-dataset-response-and-robustness}

Following the aggregation procedure defined in 
Section~\ref{evaluation-implementation-and-aggregation-principles}, 
image-level F1 refers to the pixel-level F1 calculated separately for each image. 
The mean of these image-level F1 values within each model--dataset pair is reported 
as dataset-level Macro-F1, and the unweighted mean of the six dataset-level values 
is reported as equal-dataset Macro-F1.

\begin{figure}
\centering
\includegraphics[width=0.75\linewidth]{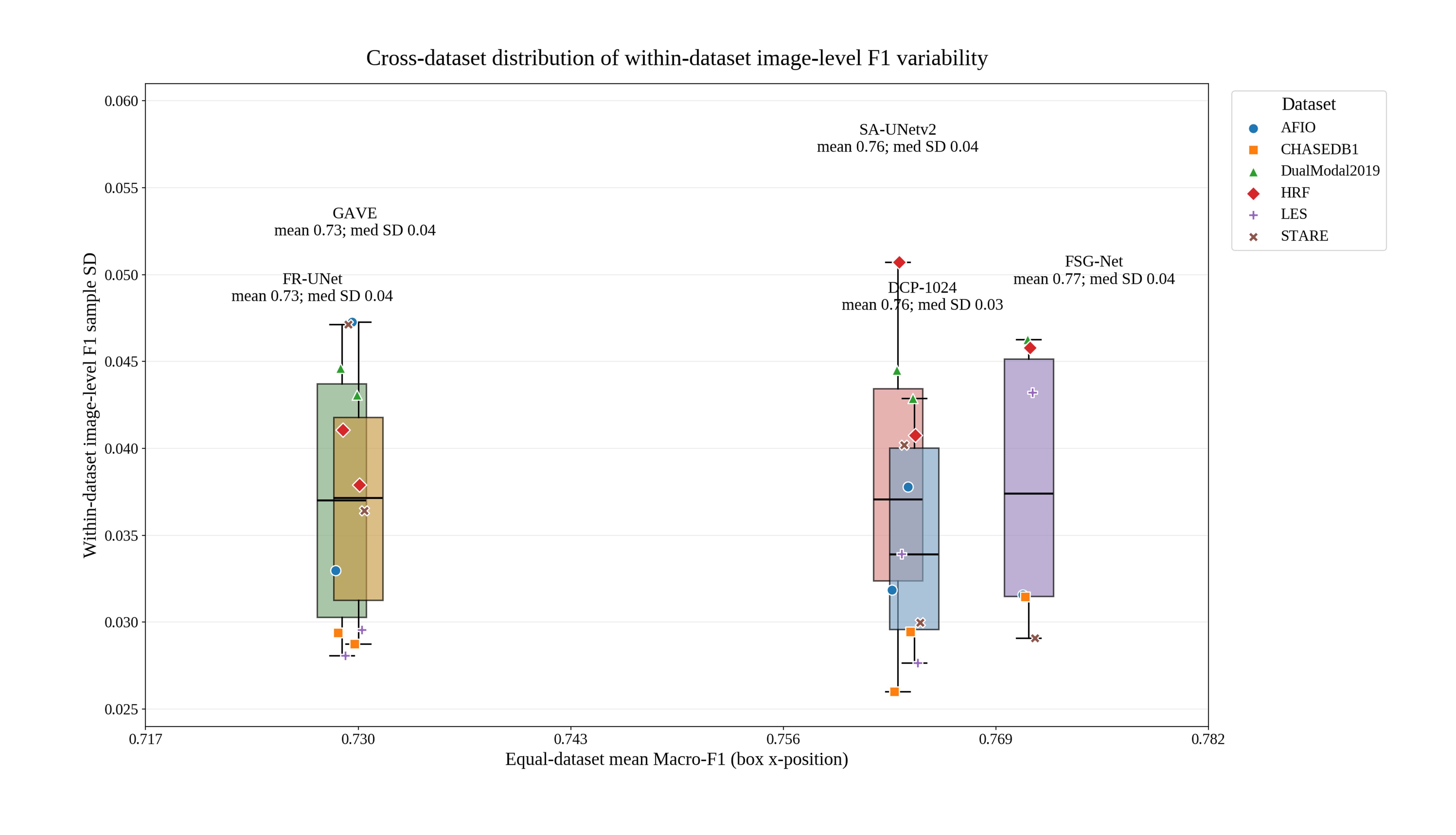}
\caption{Distribution of within-dataset image-level F1 variability across the six evaluation datasets.}
\label{fig:image5}
\end{figure}

Equal-dataset Macro-F1 summarizes average performance across datasets but does not 
capture performance stability. We therefore examined two complementary sources of 
variation: image-level variability within each dataset 
(Figure~\ref{fig:image5}) and variability in dataset-level Macro-F1 across the six 
datasets (Table~\ref{tab:cross_dataset_robustness}).

For each model--dataset pair, we calculated the sample standard deviation (SD) of 
the image-level F1 values to quantify within-dataset variability. The resulting six 
dataset-specific SD values for each model are summarized in 
Figure~\ref{fig:image5}, together with the corresponding equal-dataset Macro-F1.

FSG-Net, DCP-1024, and SA-UNetv2 achieved similar equal-dataset Macro-F1 values (0.76--0.77), and their within-dataset F1 variability largely overlapped. Thus, no clear separation among the three leading pipelines was observed in terms of image-level consistency within individual datasets.

Cross-dataset robustness was assessed using three complementary summaries: equal-dataset Macro-F1, cross-dataset SD, and the worst observed dataset-level Macro-F1.

\begin{table}[H]
\centering
\caption{Cross-dataset robustness of the evaluated pipelines.
Bold values indicate the best result for each robustness measure.}
\label{tab:cross_dataset_robustness}
\scriptsize

\begin{tabular}{lllll}
\toprule
Model & \makecell[l]{Equal-dataset\\Macro-F1} & Cross-dataset SD & Worst Macro-F1 & Worst dataset \\
\midrule
FSG-Net     & \textbf{0.77} & \textbf{0.05} & \textbf{0.70} & AFIO \\
DCP-1024    & 0.76          & 0.07          & 0.64          & AFIO \\
SA-UNetv2   & 0.76          & \textbf{0.05} & \textbf{0.70} & AFIO \\
GAVE        & 0.73          & 0.06          & 0.65          & AFIO \\
FR-UNet     & 0.73          & \textbf{0.05} & 0.66          & AFIO \\
\bottomrule
\end{tabular}

\end{table}

FSG-Net showed the strongest overall cross-dataset profile, combining the highest equal-dataset Macro-F1 (0.77) with the joint-lowest cross-dataset SD (0.05) and joint-best worst-case Macro-F1 (0.70). SA-UNetv2 was closely comparable, with the same cross-dataset SD and worst-case performance but a slightly lower equal-dataset Macro-F1 (0.76). In contrast, FR-UNet also had a low cross-dataset SD (0.05), but this stability occurred at a substantially lower mean performance level (0.73). This illustrates that variability alone is not a sufficient measure of robustness; mean performance and downside risk must be considered jointly.

Overall, FSG-Net and SA-UNetv2 form the strongest shortlist when robustness to an unknown target domain is important, with FSG-Net providing the higher average performance and SA-UNetv2 showing nearly identical cross-dataset stability and worst-case behavior.

\subsection{Structure-level Performance and Model Selection}\label{structure-level-performance-and-model-selection} 

Pixel-level agreement does not fully describe whether vascular structure is 
preserved. This section therefore evaluates structure-level performance using 
two complementary evidence groups: Thin-Vessel Sensitivity (TVS) for fine-branch 
recovery, and clDice together with tolerant Skeleton F1 (SF1) for centerline and 
skeleton fidelity.

\subsubsection{Thin-Vessel Retention}\label{thin-vessel-retention}

Thin-vessel recovery was evaluated using TVS, together with its cross-dataset 
variation and worst-dataset result. Overall sensitivity was retained as 
complementary FN-oriented evidence because it reflects missed-vessel errors 
across the full vascular mask. Table~\ref{tab:thin_vessel_summary} summarizes 
the mean, variability, and worst-dataset TVS together with overall sensitivity.

\begin{table}[H]
\centering
\caption{Equal-dataset thin-vessel retention summary. Means and standard
deviations were calculated from the six dataset-level values with equal
dataset weighting. TVS variability should be interpreted jointly with the
mean and worst-case TVS. Bold values indicate the strongest
selection-relevant results.}
\label{tab:thin_vessel_summary}
\scriptsize
\begin{tabular}{lllll}
\toprule
\textbf{Model} &
\textbf{Mean TVS} &
\textbf{TVS SD} &
\textbf{Worst-case TVS (dataset)} &
\textbf{Mean Sensitivity} \\
\midrule

SA-UNetv2 &
\textbf{0.63} &
0.12 &
\textbf{0.52 (HRF)} &
\textbf{0.85} \\

FSG-Net &
0.62 &
0.12 &
0.51 (AFIO) &
0.84 \\

DCP-1024 &
0.52 &
0.13 &
0.40 (HRF) &
0.79 \\

GAVE &
0.51 &
0.13 &
0.35 (CHASEDB1) &
0.80 \\

FR-UNet &
0.47 &
0.18 &
0.29 (LES) &
0.77 \\

\bottomrule
\end{tabular}
\end{table}

SA-UNetv2 achieved the highest mean TVS (0.63), worst-dataset TVS (0.52), and mean Sensitivity (0.85). Compared to FSG-Net, the advantages in mean TVS and mean Sensitivity were small, suggesting both pipelines are strong candidates when missed-vessel reduction is prioritized.

\begin{figure}
    \centering
    \includegraphics[width=0.75\linewidth]{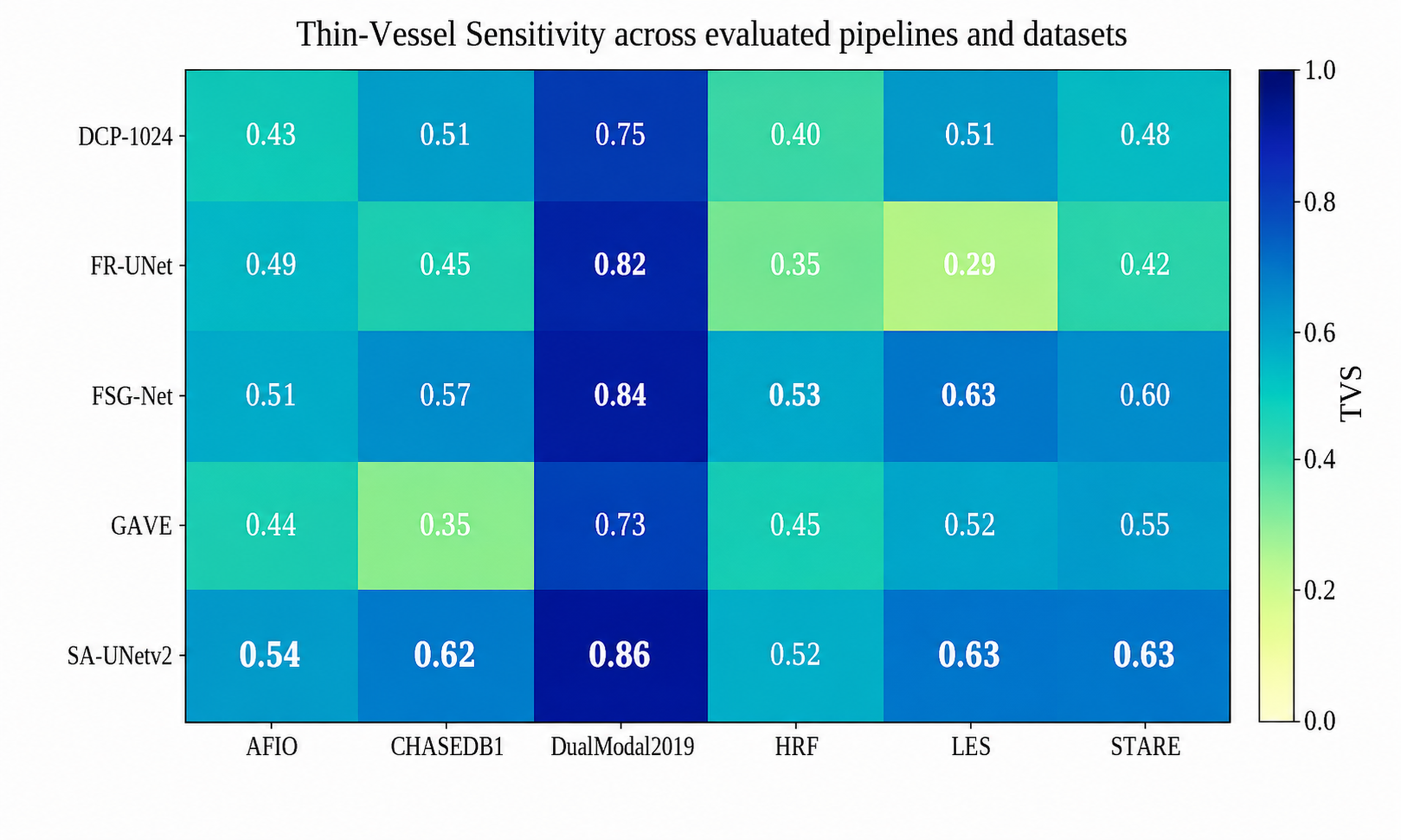}
    \caption{Thin-Vessel Sensitivity across evaluated pipelines and datasets.}
    \label{fig:image6}
\end{figure}

The aggregate results do not show whether the SA-UNetv2 advantage was limited to individual datasets. Figure~\ref{fig:image6} therefore presents the dataset-level TVS values for all model-dataset pairs.

SA-UNetv2 achieved the highest TVS on five of the six evaluated datasets and had only a 0.01 gap from FSG-Net (0.53) on HRF, indicating that its aggregate advantage was not driven by a single dataset. The result supports SA-UNetv2 as the preferred pipeline when thin-vessel omission carries the greatest cost.

\subsubsection{Centerline and Structural Fidelity}
\label{centerline-and-structural-fidelity}

\begin{figure}
    \centering
    \includegraphics[width=0.75\linewidth]{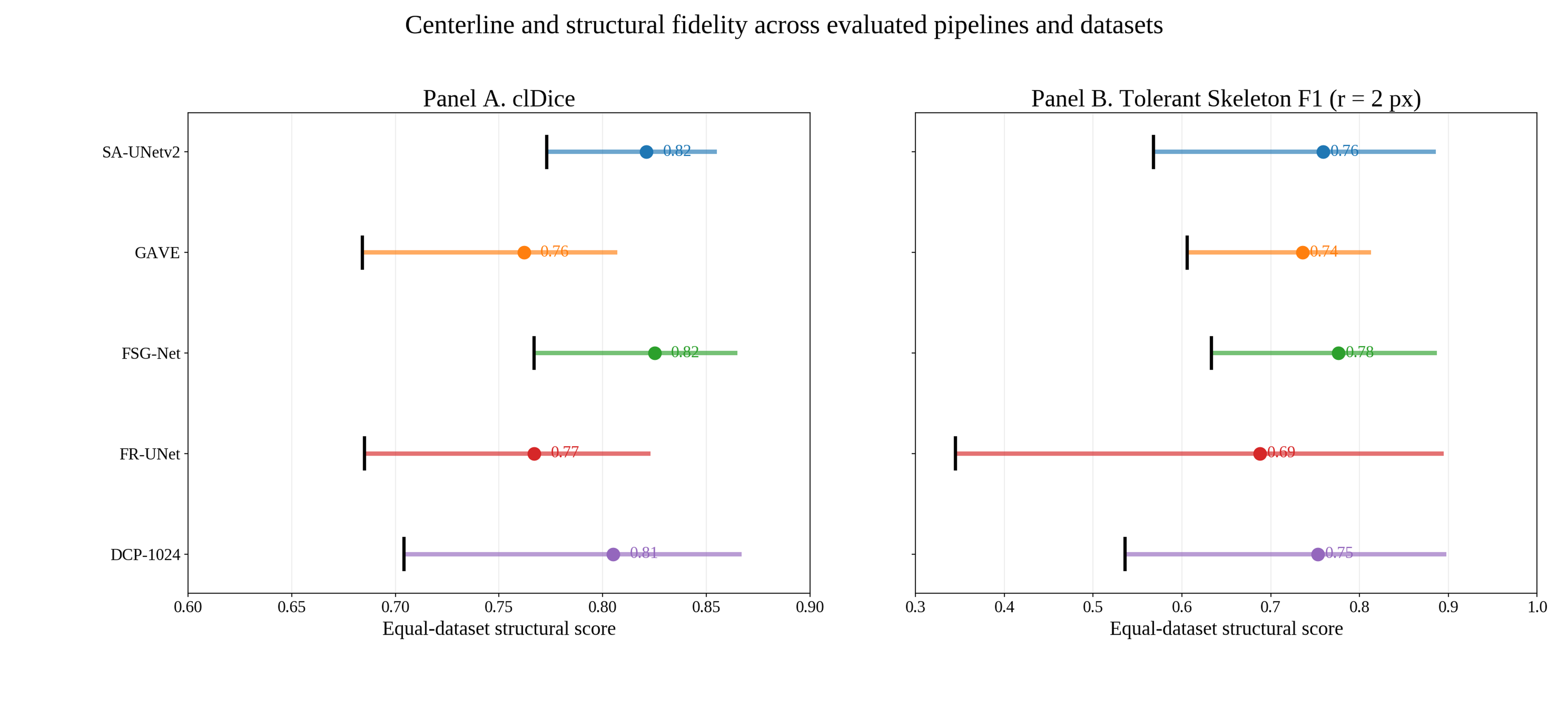}
    \caption{Cross-dataset structural fidelity measured by clDice
(left) and tolerant Skeleton F1 with a 2-pixel tolerance (right).}
    \label{fig:image7}
\end{figure}

Following the metric definitions in 
Section~\ref{evaluation-metrics-and-interpretation}, clDice measures 
bidirectional centerline-to-region agreement \cite{shit2021cldice}, whereas 
tolerant Skeleton F1 (SF1) measures alignment between the predicted and reference 
skeletons within a specified spatial tolerance. The main analysis used a 
2-pixel tolerance for SF1. Figure~\ref{fig:image7} summarizes both structural 
metrics across the evaluated pipelines and datasets. Together, clDice and SF1 
provide complementary evidence of vascular-structure preservation.

FSG-Net achieved the highest equal-dataset means for both structural measures, while SA-UNetv2 remained close in clDice.

Table~\ref{tab:structural_fidelity_summary} quantifies these differences and separates mean structural performance from worst-dataset behavior.

\begin{table}[H]
\caption{Equal-dataset centerline and structural-fidelity summary. Means and standard deviations were calculated from the six dataset-level values with
equal dataset weighting. clDice and SF1 represent complementary structural
criteria and are not interpreted as interchangeable measures. Bold values
indicate the highest mean and worst-case structural scores; SDs are reported
as measures of cross-dataset variability.}
\label{tab:structural_fidelity_summary}
\scriptsize
\begin{tabular}{lllll}
\toprule
\textbf{Model} &
\textbf{Mean clDice $\pm$ SD} &
\textbf{Worst-case clDice (dataset)} &
\textbf{Mean SF1 $\pm$ SD} &
\textbf{Worst-case SF1 (dataset)} \\
\midrule

FSG-Net &
\textbf{0.83 $\pm$ 0.04} &
\textbf{0.77 (AFIO)} &
\textbf{0.78 $\pm$ 0.11} &
\textbf{0.63 (HRF)} \\

SA-UNetv2 &
0.82 $\pm$ 0.03 &
\textbf{0.77 (AFIO)} &
0.76 $\pm$ 0.13 &
0.57 (HRF) \\

DCP-1024 &
0.81 $\pm$ 0.06 &
0.70 (AFIO) &
0.75 $\pm$ 0.15 &
0.54 (HRF) \\

FR-UNet &
0.77 $\pm$ 0.06 &
0.69 (HRF) &
0.69 $\pm$ 0.21 &
0.35 (HRF) \\

GAVE &
0.76 $\pm$ 0.05 &
0.68 (AFIO) &
0.74 $\pm$ 0.08 &
0.61 (AFIO) \\

\bottomrule
\end{tabular}
\end{table}

FSG-Net achieved the highest mean clDice (0.83) and SF1 (0.78) and retained the strongest worst-dataset SF1 (0.63). Taken together, these results support FSG-Net as the benchmark-preferred pipeline when centerline and skeleton fidelity are prioritized, with SA-UNetv2 as the main alternative when thin-vessel retention is also important.

\subsection{Model-Specific Failure Profiles}\label{model-specific-failure-profiles} 

Aggregate pixel and structural metrics indicate which pipelines perform better under a selected objective, but they do not identify the types of errors that remain. This section therefore decomposes the residual segmentation errors into model-specific structural failure profiles.

\subsubsection{Structural Error Taxonomy}\label{structural-error-taxonomy}

FN and FP pixels were separated into five interpretable categories. The taxonomy distinguishes local displacement and width bias from complete structural omission and separate spurious structures.

\begin{table}[H]
\caption{Structural error taxonomy and its interpretation for model selection.}
\label{tab:structural_error_taxonomy}
\scriptsize
\begin{tabularx}{\textwidth}{
@{}
>{\raggedright\arraybackslash}p{0.12\textwidth}
>{\raggedright\arraybackslash}X
>{\raggedright\arraybackslash}X
@{}
}
\toprule
\textbf{Error type} &
\textbf{Operational definition} &
\textbf{Model-selection interpretation} \\
\midrule

Associated metric signal &
FN-oriented errors are reflected by Sensitivity and TVS, FP-oriented errors by Precision, and centerline or continuity errors by clDice and SF1 &
Links each error type to the metric dimensions used in the benchmark; the taxonomy is complementary to, not a replacement for, the aggregate metrics. \\

\addlinespace

Boundary shift &
Paired FN and FP pixels with one-pixel 8-neighbor adjacency; the shared count is defined as
$\min(\text{adjacent FN},\,\text{adjacent FP})$ &
Represents a localized boundary displacement and avoids double-counting the same spatial mismatch as independent FN and FP errors. \\

\addlinespace

Too thin &
Residual adjacent FN pixels remaining after shared boundary-shift pairing &
Indicates localized underestimation of vessel width. \\

\addlinespace

Missed vessel &
FN pixels with no predicted-mask pixel in their 8-neighborhood &
Indicates vessel omission under the documented legacy classification rule. \\

\addlinespace

Too thick &
Residual adjacent FP pixels remaining after shared boundary-shift pairing &
Indicates localized overestimation of vessel width. \\

\addlinespace

Extra vessel &
FP pixels with no ground-truth-mask pixel in their 8-neighborhood &
Indicates an unsupported vessel response under the documented legacy classification rule. \\

\bottomrule
\end{tabularx}
\end{table}

\begin{figure}
    \centering
    \includegraphics[width=0.75\linewidth]{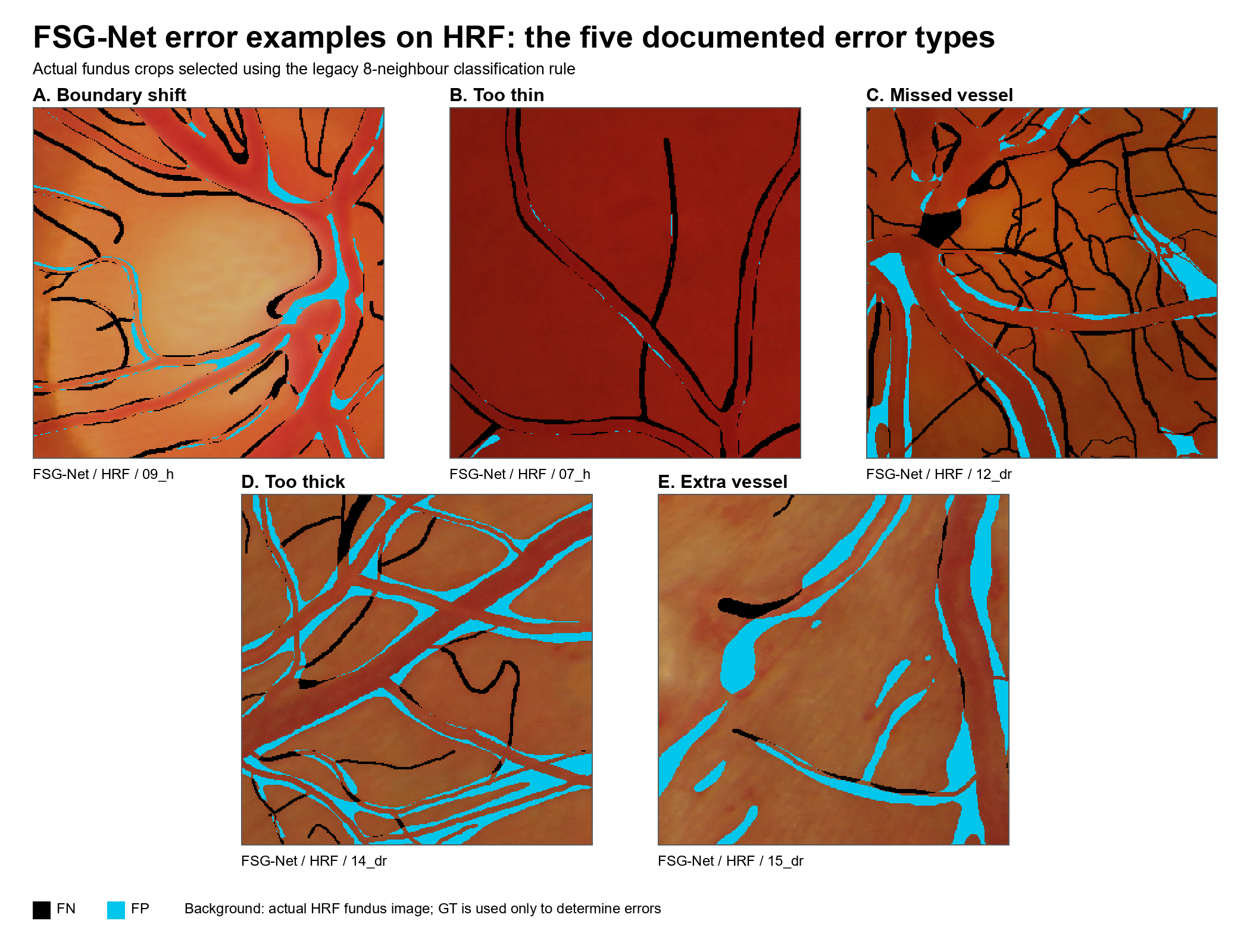}
    \caption{Representative FSG-Net error patterns on HRF under the structural error taxonomy. Black denotes false-negative (FN) pixels, and blue denotes false-positive (FP) pixels. The original fundus image is retained as the background.}
    \label{fig:image8}
\end{figure}

The categories represent different practical risks. Boundary shift indicates spatial misalignment, while too-thin and too-thick predictions indicate width bias. Missed vessels represent structural omission, and extra vessels indicate unsupported responses to texture, pathology, shadow, or illumination. Pipelines with similar total error rates may show different residual-error profiles. Table~\ref{tab:structural_error_taxonomy} summarizes these five categories, and Figure~\ref{fig:image8} provides representative FSG-Net examples on HRF to show how they appear in image space.

Figure~\ref{fig:image8} illustrates the distinct spatial patterns represented by the five error categories. Section~\ref{model-wise-residual-error-composition} next compares their composition across pipelines.

\subsubsection{Model-Wise Residual Error Composition}
\label{model-wise-residual-error-composition}

\begin{figure}
    \centering
    \includegraphics[width=0.75\linewidth]{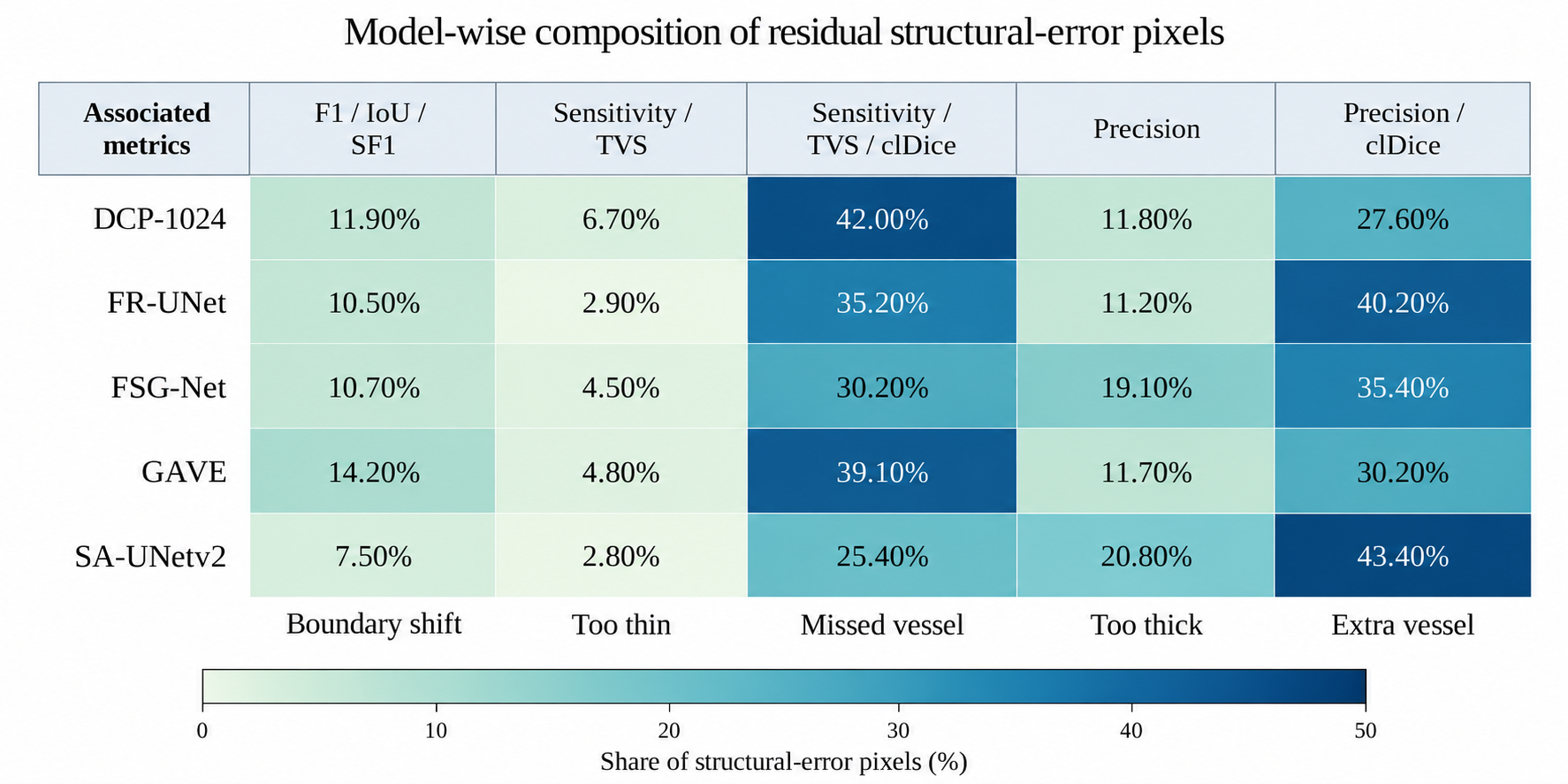}
    \caption{Model-wise composition of residual structural-error pixels. 
    The annotation row indicates the evaluation metrics most directly associated 
    with each error category; these associations are complementary rather than 
    one-to-one.}
    \label{fig:image9}
\end{figure}

The five structural-error categories were aggregated by pipeline to characterize 
model-specific residual-error profiles. Figure~\ref{fig:image9} shows the relative 
composition of these errors and links each category to the evaluation metrics most 
directly affected by that type of error. These metric associations are intended as 
interpretive links rather than one-to-one correspondences.

Missed-vessel and extra-vessel errors were the two dominant components across all 
pipelines, together accounting for 65.60\%--75.40\% of classified structural-error 
pixels. Their relative contributions nevertheless differed across pipelines. 
For example, DCP-1024 showed a larger missed-vessel component, whereas SA-UNetv2 
showed a larger extra-vessel component. These differences indicate distinct 
residual-error profiles even when pipelines are evaluated using the same metric 
framework.

These profiles provide a secondary selection layer after the scenario-based 
shortlist. A shortlisted pipeline can therefore be further evaluated according 
to whether its dominant residual errors are compatible with the application's 
error priorities.

\section{Dataset Effects and Domain Dependence}\label{dataset-effects-and-domain-dependence}

Section~\ref{evidence-guided-model-selection} compares the five pipelines under task-conditioned priorities. This section instead examines how dataset affects segmentation performance. These differences are considered in relation to dataset heterogeneity. Annotation uncertainty and limited dataset coverage are discussed in
Section~\ref{discussion-limitations-and-conclusions}.

\subsection{Dataset Heterogeneity}\label{dataset-heterogeneity}

Section~\ref{evaluation-datasets-and-source-traceability} records dataset provenance and benchmark realization. This section further examines benchmark-derived dataset characteristics, including thin-skeleton composition, vessel occupancy, and vessel-diameter descriptors. These characteristics provide context for the dataset-dependent performance observed below. Table~\ref{tab:dataset_morphology} summarizes these benchmark-derived
morphological descriptors.

\begin{table}[H]
\centering
\caption{Benchmark-derived dataset morphology descriptors. All morphology
descriptors were recomputed from the exact reference masks used in the
benchmark.}
\label{tab:dataset_morphology}

\scriptsize
\begin{tabularx}{\textwidth}{
@{}
>{\raggedright\arraybackslash}p{0.10\textwidth}
>{\raggedright\arraybackslash}p{0.15\textwidth}
>{\raggedright\arraybackslash}p{0.20\textwidth}
>{\raggedright\arraybackslash}p{0.20\textwidth}
>{\raggedright\arraybackslash}X
@{}
}
\toprule
\textbf{Dataset} &
\textbf{Thin-skeleton share} &
\textbf{Vessel occupancy (visible FOV)} &
\textbf{Mean thin-vessel diameter} &
\textbf{Mean thick-vessel diameter} \\
\midrule
AFIO & 0.90\% & 0.15 & 2.43 & 7.80 \\
CHASEDB1 & 30.60\% & 0.10 & 2.18 & 7.77 \\
DualModal2019 & 39.00\% & 0.08 & 2.17 & 6.48 \\
HRF & 50.10\% & 0.09 & 4.16 & 13.35 \\
LES & 25.30\% & 0.10 & 4.13 & 11.17 \\
STARE & 22.00\% & 0.08 & 2.00 & 4.84 \\
\bottomrule
\end{tabularx}
\end{table}

HRF has relatively large vessel diameters, whereas STARE has relatively small diameters. AFIO stands as a special case with high visible-field vessel occupancy (0.15) but a very low thin-skeleton share (0.90\%). These descriptors summarize dataset-level differences that may help explain variations in pixel-level and structural performance. Because only six datasets were included, we did not use them to estimate generalizable correlations or transfer relationships.

\subsection{Dataset-Conditioned Performance and Variability}\label{dataset-conditioned-performance-and-variability}

\begin{figure}
    \centering
    \includegraphics[width=0.75\linewidth]{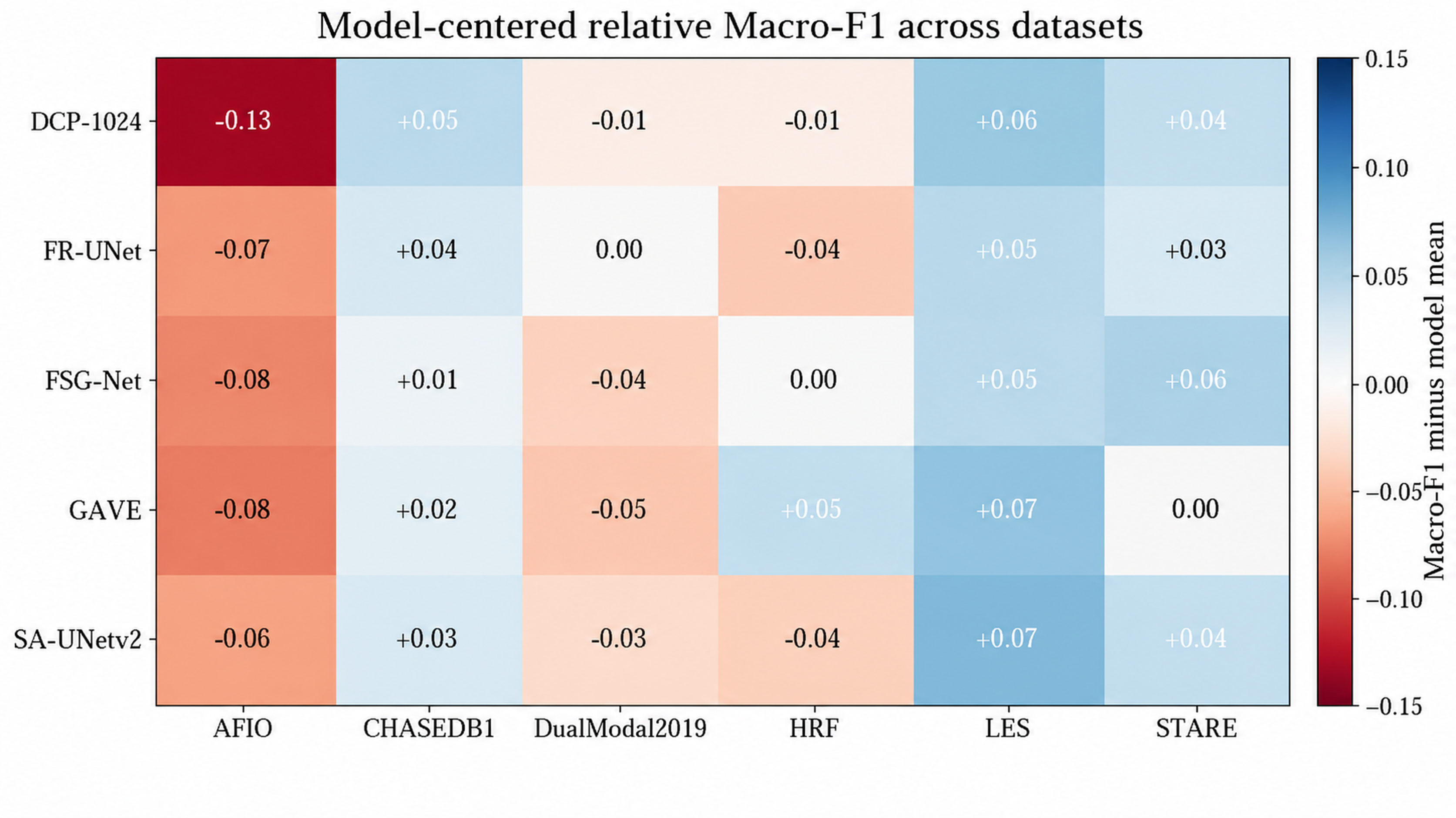}
    \caption{Model-centered relative Macro-F1 across datasets.}
    \label{fig:image10}
\end{figure}

Figure~\ref{fig:image10} uses model-centering to remove each pipeline's cross-dataset baseline and expose dataset-wide shifts. Positive and negative values therefore indicate performance above or below a pipeline's own cross-dataset mean.

Most datasets showed the shifts in the same direction across pipelines. Even though the direction was not fully consistent for HRF, its cross-pipeline spread remained below 0.1. Together, these patterns indicate that dataset identity produced a broadly shared performance shift across pipelines.

\subsection{Dataset-Conditioned Structural Performance}\label{dataset-conditioned-structural-performance}

Dataset effects extended beyond pixel overlap to thin-vessel recovery, centerline fidelity, and skeleton alignment. Tables \ref{tab:dataset_thin_vessel_recoverability} and \ref{tab:dataset_structural_fidelity} summarize these measures by dataset. The SD reported in these tables is across-model SD within a dataset.

\begin{table}
\centering
\caption{Dataset-level thin-vessel recoverability. TVS and sensitivity values are reported as the across-model mean $\pm$ standard deviation.}
\label{tab:dataset_thin_vessel_recoverability}
\footnotesize
\begin{tabular}{lll}
\toprule
\textbf{Dataset} &
\textbf{Mean TVS $\pm$ across-model SD} &
\textbf{Mean Sensitivity $\pm$ across-model SD} \\
\midrule
AFIO & 0.48 $\pm$ 0.05 & 0.57 $\pm$ 0.05 \\
CHASEDB1 & 0.50 $\pm$ 0.11 & 0.85 $\pm$ 0.03 \\
DualModal2019 & 0.80 $\pm$ 0.06 & 0.95 $\pm$ 0.01 \\
HRF & 0.45 $\pm$ 0.08 & 0.81 $\pm$ 0.04 \\
LES & 0.52 $\pm$ 0.14 & 0.85 $\pm$ 0.06 \\
STARE & 0.54 $\pm$ 0.09 & 0.83 $\pm$ 0.05 \\
\bottomrule
\end{tabular}
\end{table}

Descriptively, variation across datasets was larger than variation across models for both TVS and Sensitivity. TVS ranged from 0.45 to 0.80 across datasets, while the model-level range was approximately 0.47 to 0.63. Sensitivity showed a similar pattern, ranging from 0.57 to 0.95 across datasets and from approximately 0.77 to 0.85 across models. These results suggest stronger dataset dependence for thin-vessel recovery and overall vessel sensitivity. Table~\ref{tab:dataset_structural_fidelity} examines whether a similar pattern is also present in centerline and skeleton fidelity.

\begin{table}
\centering
\caption{Dataset-level centerline and skeleton fidelity. clDice and SF1($r=2$) are reported as across-model mean $\pm$ standard deviation.}
\label{tab:dataset_structural_fidelity}
\footnotesize
\begin{tabular}{lll}
\toprule
\textbf{Dataset} &
\textbf{Mean clDice $\pm$ across-model SD} &
\textbf{Mean SF1($r=2$) $\pm$ across-model SD} \\
\midrule
AFIO & 0.73 $\pm$ 0.04 & 0.63 $\pm$ 0.03 \\
CHASEDB1 & 0.81 $\pm$ 0.04 & 0.79 $\pm$ 0.04 \\
DualModal2019 & 0.82 $\pm$ 0.02 & 0.86 $\pm$ 0.02 \\
HRF & 0.76 $\pm$ 0.04 & 0.56 $\pm$ 0.14 \\
LES & 0.84 $\pm$ 0.03 & 0.74 $\pm$ 0.05 \\
STARE & 0.83 $\pm$ 0.04 & 0.88 $\pm$ 0.05 \\
\bottomrule
\end{tabular}
\end{table}

Structural performance also showed clear dataset dependence. HRF had the lowest mean SF1 (0.56) and relatively low TVS, indicating difficulty in both skeleton alignment and thin-vessel recovery. The high native resolution of HRF and subsequent pipeline-specific resizing may contribute to this pattern.

AFIO showed a broader reduction across evaluation dimensions. It had the lowest mean clDice (0.73), the second-lowest mean SF1 (0.63), and consistently low TVS, Sensitivity, and Macro-F1. AFIO therefore behaved as a low-performance outlier in the benchmark, possibly due to its special structural characteristics.

\subsection{Dataset-Conditioned Error Profiles}\label{dataset-conditioned-error-profiles}

\begin{figure}[tb]
    \centering
    \includegraphics[width=0.65\linewidth,trim=0.5cm 0.1cm 0.25cm 1.0cm,
    clip]{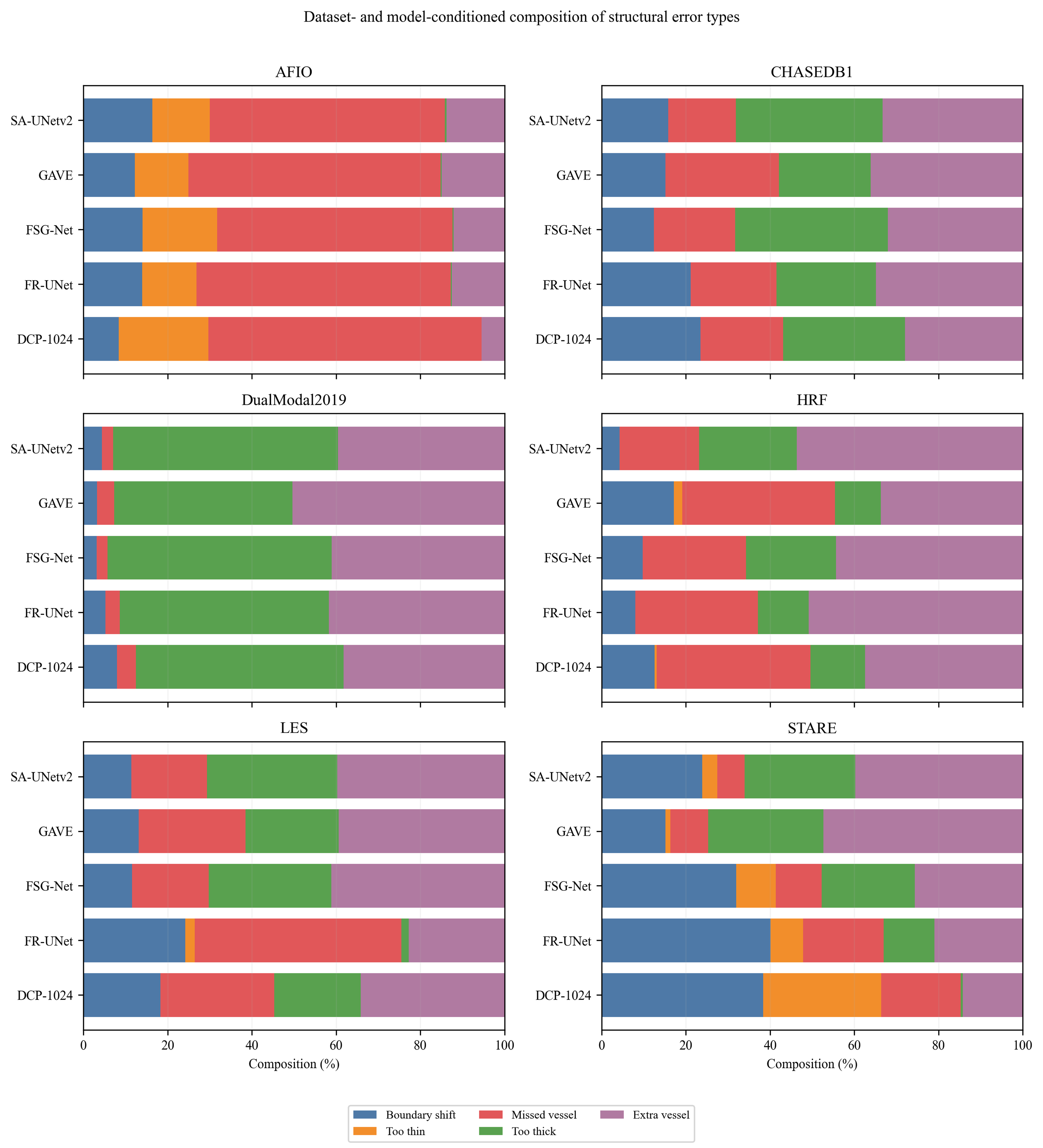}
    \caption{Dataset- and model-conditioned composition of structural error types.}
    \label{fig:image11}
\end{figure}

We then condition the same taxonomy on dataset identity to determine whether residual-error composition is driven more strongly by model or domain. Figure~\ref{fig:image11} retains the five-class structural-error taxonomy defined in Section~\ref{structural-error-taxonomy} and reorganizes the same evidence by dataset. Within each dataset, the pipelines generally showed similar residual-error compositions. The dominant error patterns varied more across datasets than across pipelines.

AFIO contained more missed-vessel errors; DualModal2019 had more too-thick and extra-vessel errors. STARE had more boundary-shift errors, while HRF and LES showed more mixed profiles. This suggests that dataset differences affect not only overall performance, but also the types of errors that remain after segmentation. These patterns may help interpret model performance, but they should not be treated as a validated rule for target-domain transfer.

The preceding benchmark established how dataset identity and target-domain conditions affect pixel-level and structural segmentation performance. Because segmentation can serve as an intermediate step for downstream retinal analyses, it is also useful to examine the evaluated methods from an application-oriented perspective. Here, classification is used as a representative downstream task, based on a recently developed method that relies extensively on vascular features \cite{li2026interpretable}. The following analysis asks a complementary downstream question: when the predicted masks are converted into vascular features and supplied to a classifier, do they remain stable and task-supporting representations?

\section{Feature-Level Evaluation for Downstream Classification}\label{feature-level-evaluation-for-downstream-classification}

\subsection{Motivation and Scope}\label{motivation-and-scope}

Segmentation masks can serve as intermediate representations for downstream analysis. We therefore use HRF disease classification as a focused task-level example to illustrate why segmentation evaluation should account for the downstream representations derived from these tasks.

\begin{figure}
    \centering
    \includegraphics[width=\linewidth]{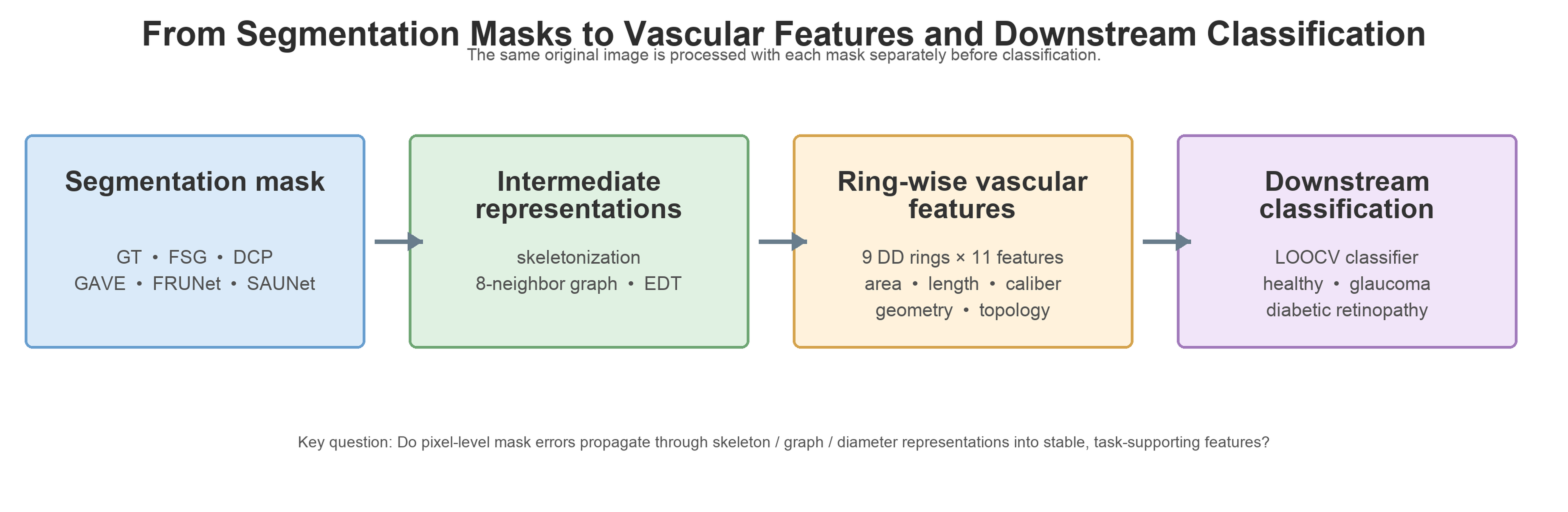}
    \caption{Complete analysis pipeline from segmentation mask to vascular features and downstream classification. The same original image is processed with each segmentation mask separately; the classifier receives only the ring-wise vascular features generated from the corresponding mask.}
    \label{fig:image12}
\end{figure}

Figure~\ref{fig:image12} summarizes how each segmentation mask is converted into the vascular-feature representation used for classification.

\subsection{Experimental Setting}\label{experimental-setting}

The experiment used 45 HRF images from three classes---healthy, glaucoma, and diabetic retinopathy---and evaluated GT, FSG, DCP, GAVE, FRUNet, and SA-UNet masks separately.

Within each of nine concentric rings defined relative to the optic-disc diameter, we computed 11 vascular feature types for each mask: vessel area fraction (VAF), vessel length density (VLD), and branch-point density (BPD), caliber diameter interquartile range (IQR), thin vessel length (L\textsubscript{thin}),~estimated thin-to-thick vessel diameter ratio (D\textsubscript{ratio}), tortuosity IQR, median signed curvature,~median absolute curvature, vascular connection count (VCC), and mean vascular connectivity (VC\textsubscript{mean}) \cite{li2026interpretable}. These descriptors include retinal vascular measurements related to vessel diameter summaries \cite{knudtson2003revised} and tortuosity- or curvature-based vascular geometry \cite{hart1999tortuosity}. For the two connectivity features, the vessel skeleton was represented as an undirected 8-connected graph. VCC represents the total number of graph connections within a ring, whereas VC\textsubscript{mean}~normalizes this quantity by graph size to characterize average local connectedness. Each method used its own feature table, and classification was evaluated with leave-one-out cross-validation using standardized features and an elastic-net-regularized logistic-regression classifier \cite{zou2005elasticnet,pedregosa2011scikit}. Detailed classifier settings are provided in Appendix~\ref{appendix-j.-supplementary-downstream-feature-level-analyses}.

An image-level classifier was retained only as contextual reference; the present analysis focuses on the explicit vascular-feature route because it permits the effect of a mask on an interpretable intermediate quantity to be examined.

\subsection{Baseline Classification and Pixel--Task Mismatch}\label{baseline-classification-and-pixeltask-mismatch}

The baseline vascular-only results are summarized in Table~\ref{tab:vascular_classification}. The ordering suggested by pixel-level segmentation scores was not preserved by downstream classification, showing that mask similarity alone is insufficient to predict task performance.

\begin{table}
\centering
\caption{Baseline vascular-only classification performance.}
\label{tab:vascular_classification}
\footnotesize
\begin{tabular}{lcccc}
\toprule
\textbf{Method} &
\textbf{Correct / total} &
\textbf{Errors} &
\textbf{Accuracy} &
\textbf{95\% Wilson CI} \\
\midrule
GAVE & 34/45 & 11 & 0.76 & [0.61, 0.86] \\
GT & 33/45 & 12 & 0.73 & [0.59, 0.84] \\
SA-UNetv2 & 29/45 & 16 & 0.64 & [0.50, 0.77] \\
FR-UNet & 28/45 & 17 & 0.62 & [0.48, 0.75] \\
FSG-Net & 27/45 & 18 & 0.60 & [0.45, 0.73] \\
DCP-1024 & 24/45 & 21 & 0.53 & [0.39, 0.67] \\
\bottomrule
\end{tabular}
\end{table}

Because classification operates on method-specific vascular features rather than mask pixels, pixel-level segmentation rankings need not be preserved downstream. The GAVE--GT reversal and the underlying classifier mechanism are examined in Appendix~\ref{j.2-detailed-explanation-of-pixelclassification-mismatch}.

\subsection{Controlled Perturbation Experiment}\label{controlled-perturbation-experiment}

We applied controlled mask dilation, mask erosion, and skeleton tip pruning to the GT-derived inputs, then re-extracted the vascular features. These perturbations represent small changes in vessel boundaries, thin branches, and terminal vessels.

\begin{table}
\caption{Summary of vascular-feature classification under controlled mask perturbations.}
\label{tab:classification_perturbations}
\scriptsize
\begin{tabularx}{\textwidth}{
@{}
>{\raggedright\arraybackslash}p{0.18\textwidth}
>{\centering\arraybackslash}p{0.18\textwidth}
>{\centering\arraybackslash}p{0.14\textwidth}
>{\raggedright\arraybackslash}X
@{}
}
\toprule
\textbf{Perturbation} &
\textbf{Baseline train $\rightarrow$ perturbed test} &
\textbf{Perturbed LOOCV} &
\textbf{Interpretation} \\
\midrule
No perturbation (GT) &
0.73 &
-- &
Original vascular-feature baseline. \\
\addlinespace
\texttt{mask\_dilate1} &
0.42 &
0.71 &
Global shifts in vessel area and caliber substantially reduce performance
when the original classifier is applied without re-adaptation. \\
\addlinespace
\texttt{mask\_erode1} &
0.44 &
0.87 &
Erosion alters thin-vessel representation and connectivity, while substantial
class-discriminative information remains after re-adaptation. \\
\addlinespace
\texttt{skeleton\_prune\_tips5} &
0.56 &
0.80 &
Removal of terminal branches changes geometric and topological features,
while much of the classification signal remains recoverable after re-adaptation. \\
\bottomrule
\end{tabularx}
\end{table}

The resulting sensitivity changes demonstrate that perturbations of similar pixel magnitude can have different downstream effects, depending on the representation used by a feature. In Table~\ref{tab:classification_perturbations}, baseline train $\rightarrow$ perturbed test measures distribution-shift performance without retraining, whereas perturbed LOOCV measures performance after re-fitting the classifier on the perturbed feature table. Full protocols and ring-wise results are provided in Appendix~\ref{j.4-full-controlled-perturbation-protocol-and-sensitivity-calculation}.

\subsection{Feature Stabilization}\label{feature-stabilization}

We next tested mechanism-specific stabilization rather than uniformly smoothing every feature. Spline centerlines, Gaussian-smoothed EDT sampling, and graph pruning were applied only where the perturbation analysis indicated potential numerical amplification.

\begin{table}
\caption{Sensitivity of selected vascular features before and after
targeted stabilization. Relative reduction is computed with respect to the
original sensitivity; larger reductions indicate greater stabilization.}
\label{tab:feature_stabilization}
\scriptsize
\begin{tabularx}{\textwidth}{
@{}
>{\raggedright\arraybackslash}p{0.12\textwidth}
>{\raggedright\arraybackslash}p{0.15\textwidth}
>{\centering\arraybackslash}p{0.10\textwidth}
>{\centering\arraybackslash}p{0.10\textwidth}
>{\centering\arraybackslash}p{0.10\textwidth}
>{\raggedright\arraybackslash}X
@{}
}
\toprule
\textbf{Feature} &
\textbf{Stabilization procedure} &
\textbf{Original sensitivity} &
\textbf{Processed sensitivity} &
\textbf{Relative reduction} &
\textbf{Interpretation} \\
\midrule
Calibre diameter IQR &
Spline centerline + Gaussian-smoothed EDT &
0.3528 & 0.2897 & 17.9\% &
Sensitivity decreased, indicating a clear improvement in stability. \\
\addlinespace
Tortuosity IQR &
Spline centerline &
2.9719 & 0.9037 & \textbf{69.6\%} &
Largest sensitivity reduction among the five evaluated features. \\
\addlinespace
Median signed curvature &
Spline centerline &
3.0228 & 1.2569 & 58.4\% &
Substantially reduced sensitivity to local directional jitter. \\
\addlinespace
Median absolute curvature &
Spline centerline &
2.6024 & 1.2569 & 51.7\% &
Reduced sensitivity to pixel-level centerline jaggedness. \\
\addlinespace
VCC &
Graph pruning &
0.3107 & 0.2911 & 6.3\% &
Modest stabilization, primarily through removal of short terminal spurs. \\
\bottomrule
\end{tabularx}
\end{table}

We kept the final stabilization scheme deliberately selective: five feature definitions were retained for processing, whereas VLD was left unchanged because its spline variant did not improve sensitivity. Detailed definitions and parameter settings are provided in Appendix~\ref{j.5-stabilization-definitions-parameter-settings-and-processing-effects}.

\subsection{Classification After Stabilization}\label{classification-after-stabilization}

The six mask inputs were then re-evaluated using the same 11-feature vascular representation, with only the selected stabilized definitions changed. The resulting classification performance is summarized in Table~\ref{tab:classification_after_stabilization}.

\begin{table}
\centering
\caption{Vascular-only classification performance after selective feature stabilization.}
\label{tab:classification_after_stabilization}
\footnotesize
\begin{tabular}{lcccc}
\toprule
\textbf{Method} &
\textbf{Correct / total} &
\textbf{Errors} &
\textbf{Accuracy} &
\textbf{95\% Wilson CI} \\
\midrule
GT & 35/45 & 10 & 0.78 & [0.64, 0.87] \\
FSG-Net & 30/45 & 15 & 0.67 & [0.52, 0.79] \\
SA-UNetv2 & 29/45 & 16 & 0.64 & [0.50, 0.77] \\
FR-UNet & 27/45 & 18 & 0.60 & [0.45, 0.73] \\
DCP-1024 & 26/45 & 19 & 0.58 & [0.43, 0.71] \\
GAVE & 25/45 & 20 & 0.56 & [0.41, 0.69] \\
\bottomrule
\end{tabular}
\end{table}

After stabilization, FSG-Net and SA-UNetv2 became the leading predicted-mask inputs for downstream vascular-feature classification on HRF. Their strong downstream performance was consistent with their strong structure-level segmentation performance, supporting the value of representation-aware evaluation. However, this analysis remains a descriptive proof of concept on HRF rather than a universal ranking of the pipelines, and further investigation on independent datasets is needed.

\section{Discussion, Limitations, and Conclusions}
\label{discussion-limitations-and-conclusions}

The present study provides a benchmark-relative framework for selecting retinal
vessel segmentation pipelines according to application-specific priorities.
Across the predefined scenarios, FSG-Net and SA-UNetv2 accounted for all
first-ranked recommendations. FSG-Net was favored when overall pixel agreement
or structural fidelity was emphasized, whereas SA-UNetv2 was favored when
thin-vessel recovery, downside risk, or resource-aware objectives received
greater weight. DCP-1024 remained competitive for mean pixel-level performance.
These results reinforce the central finding that the preferred pipeline depends
on the intended application rather than on a universal ranking.

\paragraph{Benchmark limitations and interpretation boundaries.}
Several limitations should be considered when interpreting these recommendations.
First, the evaluated pipelines were not realized under identical experimental
conditions. They differed in input resolution, preprocessing, inference strategy,
post-processing, and training provenance. FSG-Net, FR-UNet, and SA-UNetv2 were
trained locally on FIVES, whereas DCP-1024 and GAVE were evaluated using released
checkpoints because reproducible training workflows were unavailable. Similarly,
thresholds followed source-reported operating points, or a default value of 0.50
when no threshold was specified. The benchmark therefore compares complete
executable pipelines rather than architectures under strictly controlled training
conditions.

The resource comparison is subject to the same limitation. Comparable local
training measurements were available only for the three locally trained pipelines,
and several computational fields were unavailable for DCP-1024 and GAVE.
Resource-oriented recommendations should therefore be interpreted as provisional.

Second, several secondary analyses depend on benchmark-specific definitions,
including the TVS thin-vessel criterion, the 2-pixel tolerance used for SF1, and
the structural-error taxonomy defined in
Section~\ref{structural-error-taxonomy}. These definitions support consistent
comparison but are not universal. Alternative thresholds, tolerances, or
structural definitions may change numerical values, particularly when differences
between leading pipelines are small.

Third, the benchmark covers five pipelines and six public datasets comprising
244 images. It therefore does not represent the full range of retinal imaging
devices, pathologies, acquisition conditions, or image-quality variation.
The dataset-conditioned patterns described in
Section~\ref{dataset-effects-and-domain-dependence} should therefore be treated
as descriptive rather than as validated transfer rules. Final model and threshold
selection still requires validation on representative target-domain data.

\paragraph{Reference-annotation uncertainty.}
All benchmark metrics treat the supplied vessel masks as reference annotations,
although faint peripheral vessels and exact vessel boundaries remain difficult to
annotate consistently. Errors near these boundaries may therefore arise from the
prediction, the reference annotation, or both
\cite{jin2022fives,lyu2022reta}.

\begin{figure}[H]
    \centering
    \includegraphics[width=0.75\linewidth]{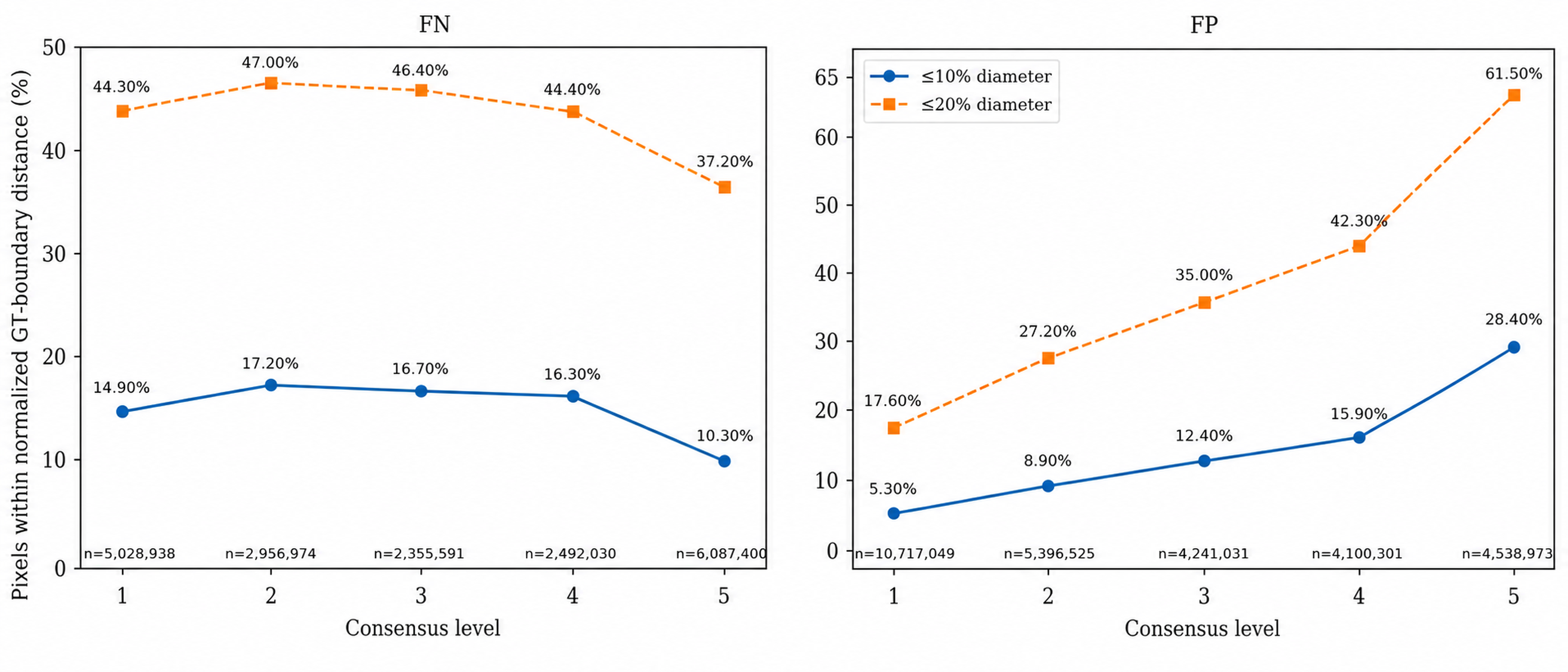}
    \caption{Proportion of shared false-negative and false-positive errors
    located within normalized ground-truth boundary bands as a function of
    cross-model consensus level.}
    \label{fig:image13}
\end{figure}

Figure~\ref{fig:image13} groups shared FN and FP pixels by the number of pipelines
making the same error and reports the fraction lying within 10\% and 20\% of the
local GT vessel diameter from the nearest annotated boundary.

High-consensus errors were frequently located near annotated vessel boundaries,
particularly for FP pixels. At consensus level 5, 61.5\% of FP pixels and 37.2\%
of FN pixels lay within 20\% of the local vessel diameter from the boundary.
This pattern may reflect reference-boundary uncertainty or shared spatial bias
across models. A supplementary boundary-exclusion analysis further showed that
Macro-F1 increased for all pipelines after excluding 10\% and 20\% boundary
bands, while the model ranking remained unchanged. Thus, boundary uncertainty may
influence absolute performance estimates without materially altering the observed
ranking under the tested settings. Full results are provided in
Appendix~\ref{appendix-f.-sensitivity-of-model-rankings-to-boundary-uncertainty}.

\paragraph{Downstream feature analysis: limitations and implications.}
The HRF downstream experiment should be interpreted as a proof-of-concept
demonstration rather than as an independent validation of the segmentation
ranking. It used only 45 images with leave-one-out cross-validation, and the
resulting accuracy estimates remain subject to substantial small-sample
uncertainty, as reflected by the relatively wide Wilson confidence intervals.
Larger cohorts and independent-dataset validation are therefore required before
the downstream ranking can be considered generalizable.

The feature-stabilization analysis also introduces methodological choices.
Spline, EDT, and graph-pruning parameters were manually selected rather than
jointly optimized. Insufficient processing may retain pixel-level noise, whereas
excessive processing may suppress genuine curvature, caliber variation, or
terminal branches. Importantly, stabilization is intended to reduce numerical
amplification of segmentation errors rather than to transform an imperfect mask
into the reference annotation. For this reason, features such as VAF and the
estimated thin-to-thick vessel diameter ratio were retained in their original
forms, and greater feature-level similarity to GT was not assumed to guarantee
better classification.

Despite these limitations, the downstream experiment highlights an important
extension of conventional segmentation evaluation. A segmentation error should
not be judged only by its pixel count, but also by how strongly it perturbs the
representation consumed by the downstream task. This motivates the broader
concept of \emph{feature-stable representations}: segmentation outputs that
preserve clinically or computationally relevant derived quantities under small
local perturbations.

\paragraph{Open problems and future research directions.}
The GT-informed oracle experiments provide diagnostic upper bounds on several
potential improvement strategies rather than deployable algorithms.

First, the oracle fusion experiment showed that errors from one pipeline can
sometimes be compensated by correct predictions from others. Under GT-informed
vote eligibility, Macro-F1 increased from 0.76 to 0.83 and IoU from 0.61 to 0.73.
This supports future development of GT-free adaptive fusion based on model
uncertainty, probability maps, image evidence, and cross-model consensus.
Detailed results are provided in
Appendix~\ref{appendix-g.-error-aware-oracle-vote-fusion-details}.

Second, the Laplacian oracle experiment showed that local image intensity contains
useful information for correcting known FN and FP errors. Across the six datasets,
mean F1 increased by 0.08--0.13 and mean IoU by 0.12--0.17. A practical extension
would therefore be to identify error-prone regions without GT and apply
image-guided correction only where needed. Full results are provided in
Appendix~\ref{appendix-h.-laplacian-oracle-repair-of-known-fnfp-pixels}.

Third, structure-aware learning and feature-stable representations should be
investigated jointly. A feasibility experiment showed that adding soft-clDice to
the SA-UNetv2 loss did not improve F1 under the tested setting
(Appendix~\ref{appendix-i.-cldice-augmented-loss-feasibility-experiment}).
This negative result does not weaken clDice as an evaluation metric, but suggests
that more targeted objectives may be needed. Endpoint-sensitive losses such as
DECL may better address small vessel discontinuities
\cite{sobisch2026declaccess}, although improvements in continuity may trade off
against pixel-level overlap \cite{juanco2023topology}.

Future work should also optimize feature-processing parameters in a controlled
manner before external validation. Let $\theta$ denote the vector of processing
parameters. A constrained objective can combine perturbation sensitivity,
feature drift, and preservation of downstream task performance:

\begin{equation*}
J(\theta)
=
\operatorname{mean}_{\mathrm{abs}}\Delta z(\theta)
+
\lambda
\max\!\left(
0,\,
\operatorname{drift}(\theta)-\tau_d
\right)
+
\mu
\max\!\left(
0,\,
\operatorname{Acc}_{\mathrm{base}}-\operatorname{Acc}(\theta)
\right).
\end{equation*}

Such a formulation makes explicit the trade-off between reducing numerical
sensitivity and preserving meaningful vascular structure and task performance.
A further extension is constrained mask refinement, in which uncertainty,
probability maps, raw image intensity, and structural priors are used to correct
isolated false positives, broken thin vessels, and irregular boundaries. These
methods should be evaluated jointly using pixel-level metrics, structural metrics,
feature sensitivity, and downstream performance.

\paragraph{Conclusion.}
Taken together, the benchmark supports a layered view of retinal vessel
segmentation evaluation. Pixel-level metrics quantify regional agreement;
structural metrics quantify centerline and connectivity preservation; cross-dataset
analysis quantifies robustness under domain variation; and feature-level analysis
tests whether segmentation outputs remain reliable when transformed into
task-specific representations. Within the present benchmark, FSG-Net and
SA-UNetv2 emerge as the strongest general candidates, but the final choice should
remain conditioned on the dominant error cost, structural requirements,
computational constraints, and target domain.

More broadly, this study suggests that segmentation pipelines should be evaluated
not only by how closely they reproduce a reference mask, but also by how reliably
they preserve the information required by downstream biomedical analyses. We hope
that the benchmark, open resources, and the open problems identified here will
motivate further work on robust, structure-aware, and application-conditioned
evaluation of biomedical image segmentation pipelines.

\section*{Funding}

This work was supported by the National Natural Science Foundation of China
Key Project (Grant No.~12231004) and the Natural Sciences and Engineering Research Council of
Canada (NSERC). 

\pdfbookmark[0]{Data and Code Availability}{bookmark-data-and-code-availability}
\section*{Data and Code Availability}\label{data-and-code-availability}

\paragraph{Open research resources.}
The benchmark comprises five executable segmentation pipelines evaluated on six datasets
containing 244 images. The complete benchmark package, including pipeline implementations,
dataset links and model provenance records, analysis outputs, and the model-selection tool, is
publicly available in the
\href{https://github.com/ritaaaa-yry/retinal-segmentation-guidance}{GitHub repository}.
The accompanying interactive ranking interface is available through the
\href{https://ritaaaa-yry.github.io/retinal-segmentation-guidance/model-ranking-page/}
{live model-ranking tool}.


\section*{Declaration of Competing Interest}

The authors declare that they have no known competing financial interests or
personal relationships that could have appeared to influence the work reported
in this paper.

\section*{Ethics Statement}

This study used only publicly available, previously released retinal image
datasets and did not involve prospective recruitment, intervention, or the
collection of new human or animal data. No new experiments involving human
participants or animals were conducted as part of this study.

\section*{Informed Consent and Confidentiality}

No patients or volunteers were recruited for this study. The analysis was based
on publicly available datasets released by their original providers. No
personally identifiable information was used in the present study.

\section*{CRediT Authorship Contribution Statement}
Ruoyu Yan: Methodology, Software, Investigation, Validation, Formal analysis, Visualization, 
Writing -- original draft.
Yuejia Liu: Methodology, Software, Validation, Formal analysis, Data curation, Visualization, Writing – original draft.
Shixin Xu: Conceptualization, Methodology, Supervision, Writing -- review and editing.
Xiaoyan Li: Methodology, Validation, Writing -- review and editing.
Xi Luo: Data curation, Validation, Writing -- review and editing.
Yuhui Deng: Data curation, Methodology, Validation, 
Writing -- review and editing.
Huaxiong Huang: Conceptualization, Methodology, Supervision, Funding acquisition,  
Writing -- review and editing.

\FloatBarrier
\pdfbookmark[0]{References}{bookmark-references}
\begingroup
\raggedright
\sloppy
\setlength{\emergencystretch}{3em}
\bibliographystyle{abbrv}
\bibliography{references}
\endgroup

\clearpage


\appendix

\pdfbookmark[0]{Appendix}{bookmark-appendix}
\part*{Appendix}
\label{appendix-supplementary-information}

The Appendices A-J record source-paper settings, metric definitions, configuration boundaries, sensitivity checks, oracle diagnostics, and provenance information. Values are included only where they are traceable to the retained manuscript evidence package or local run records.


\section{Source, Local, and Released Configuration Boundaries }
\label{appendix-a.-source-local-and-released-configuration-boundaries}

This Appendix records source-paper settings and released-configuration evidence. Table~\ref{tab:training_exposure_ledger} distinguishes local training, source-development exposure, and checkpoint-level exposure relevant to evaluation-dataset inclusion.
Table~\ref{tab:source_operating_procedures} summarizes the original design and references for the algorithms included in this study. 

\begin{table}[H]
\caption{Training-exposure ledger used to determine evaluation-dataset inclusion.
Source-paper exposure and local checkpoint provenance are distinguished so that
dataset exclusions are not overstated as a controlled common-training comparison.}
\label{tab:training_exposure_ledger}
\scriptsize
\begin{tabularx}{\textwidth}{
@{}
>{\raggedright\arraybackslash}p{0.18\textwidth}
>{\raggedright\arraybackslash}X
>{\raggedright\arraybackslash}p{0.14\textwidth}
>{\raggedright\arraybackslash}X
@{}
}
\toprule
\textbf{Pipeline(s)} &
\textbf{Relevant training / exposure evidence} &
\textbf{Dataset handling} &
\textbf{Interpretive qualification} \\
\midrule

FSG-Net; FR-UNet; SA-UNetv2 &
Locally supervised training on FIVES was performed for these benchmark realizations. &
FIVES excluded &
Direct overlap between local training and benchmark evaluation was avoided. \\

\addlinespace

DCP-1024 &
FIVES was included in the source broad-domain development setting; this benchmark used the released checkpoint rather than local retraining. &
FIVES excluded &
Source-development exposure is retained as a qualification and is not treated as absent. \\

\addlinespace

GAVE / RIP-AV vessel channel &
The released checkpoint used for GAVE inference in this benchmark was trained on DRIVE. &
DRIVE excluded &
The reported GAVE results are qualified by checkpoint exposure and should not be interpreted as an external test on DRIVE. \\

\bottomrule
\end{tabularx}
\end{table}

\clearpage
\begin{sidewaystable}[p]
\caption{Original design and source-paper operating procedures of the included pipelines.}
\label{tab:source_operating_procedures}
\scriptsize

\begin{tabularx}{\textheight}{
@{}
>{\raggedright\arraybackslash}p{0.10\textheight}
>{\raggedright\arraybackslash}X
>{\raggedright\arraybackslash}X
>{\raggedright\arraybackslash}X
>{\raggedright\arraybackslash}p{0.14\textheight}
>{\raggedright\arraybackslash}X
@{}
}
\toprule
\textbf{Pipeline} &
\textbf{Principal architectural mechanism} &
\textbf{Source-paper training and optimization} &
\textbf{Original input and preprocessing} &
\textbf{Probability-to-mask operating point} &
\textbf{Original inference and post-processing} \\
\midrule

FSG-Net \cite{seo2025fsgnet} &
Full-scale feature representation with modernized convolution blocks,
multiscale feature merging, guided residual modules, and attention-guided filtering &
Dataset-specific experiments on DRIVE, STARE, CHASE\_DB1, and HRF;
AdamW; learning rate \(1\times10^{-3}\); weight decay \(0.05\);
BCE + Dice loss with deep supervision; batch size 4 &
Dataset-specific center padding; random \(288\times288\) crops;
blur, color jitter, horizontal flipping, perspective transformation,
random resizing, and CutMix &
Fixed threshold of \(0.50\) &
Full-image prediction after padding to a multiple of 32;
output restored to the original image extent;
no additional mask post-processing reported \\

\addlinespace

FR-UNet + DTI \cite{liu2022frunet}&
Full-resolution representation, multiresolution convolution interaction,
feature aggregation modules, deep supervision, and dual-threshold iteration &
Dataset-specific experiments on DRIVE, CHASE\_DB1, STARE, DCA1, and CHUAC;
Adam; learning rate \(1\times10^{-4}\); weight decay \(1\times10^{-5}\);
BCE loss; cosine annealing &
RGB retinal images converted to grayscale and normalized;
\(48\times48\) training patches extracted with stride 6;
horizontal and vertical flipping and rotations at \(90^\circ\) intervals;
full-size images used for testing &
Dual-threshold iteration with high threshold \(0.50\) and low threshold \(0.30\)
reported in the connectivity analysis &
Deep-supervision outputs combined into a probability map;
high-confidence pixels initialize the mask, and connected
intermediate-confidence pixels are iteratively added until convergence \\

\addlinespace

SA-UNetv2 \cite{guo2026saunetv2} &
Compact U-Net with cross-scale spatial attention in skip connections,
Group Normalization, SiLU activations, reduced channel configuration,
and a BCE--MCC objective &
Separately trained on DRIVE and STARE;
Adam; learning rate \(1\times10^{-3}\);
\(0.5\,\mathrm{BCE}+0.5\,\mathrm{MCC}\);
maximum 150 epochs; batch size 8 for DRIVE and 2 for STARE &
DRIVE zero-padded to \(592\times592\) and STARE to \(704\times704\);
augmented training data used with a held-out validation subset &
Binary threshold not explicitly reported &
Sigmoid probability output;
predictions cropped back to the original image dimensions;
no additional structural post-processing reported \\

\addlinespace

DCP \cite{wei2025dcp} &
R2AU-Net backbone with localized position-wise and channel-wise domain prompts
inserted at multiple encoder--decoder skip connections &
Joint broad-domain training on ROSSA, FIVES, IOSTAR, PRIME-FP20, and VAMPIRE,
with each dataset split \(7{:}1{:}2\);
SGD; learning rate \(1\times10^{-3}\); momentum \(0.95\);
weight decay \(1\times10^{-4}\); BCE + Dice loss; cosine annealing &
All source-domain images resized to \(512\times512\);
each mini-batch contained one image from each domain &
Binary threshold not explicitly reported;
the source paper primarily evaluates probability outputs using AP and AUC &
Domain identity selects the corresponding prompt tensors during inference;
the shared backbone generates the probabilistic vessel output;
no separate binary-mask post-processing reported \\

\addlinespace

GAVE / RIP-AV vessel channel \cite{dai2024ripav} &
Representative-instance pre-training, patch-pair selection,
dual-stream patch and context encoding, patch--context fusion,
distance-aware supervision, and adversarial segmentation &
Dataset-specific experiments on AV-DRIVE, LES-AV, and HRF;
Adam with \(\beta_1=0.5\) and \(\beta_2=0.999\);
learning rate \(2\times10^{-4}\); batch size 8;
segmentation, distance-aware, and adversarial losses &
Multiscale vascular patches and surrounding contexts extracted at
dataset-specific scales; inputs resized to \(256\times256\);
random flipping, affine transformation, rotation, and shifting &
Binary threshold not explicitly reported &
Overlapping sliding-window inference with the complete image used as context;
overlapping predictions fused into artery, vein, and overall-vessel channels;
only the overall-vessel channel is retained in this benchmark \\

\bottomrule
\end{tabularx}
\end{sidewaystable}
\clearpage


\section{Threshold Sensitivity and Benchmark Operating Points}
\label{appendix-b.-threshold-sensitivity-and-benchmark-operating-points}

\begin{figure}[H]
    \centering
    \includegraphics[width=0.75\linewidth]{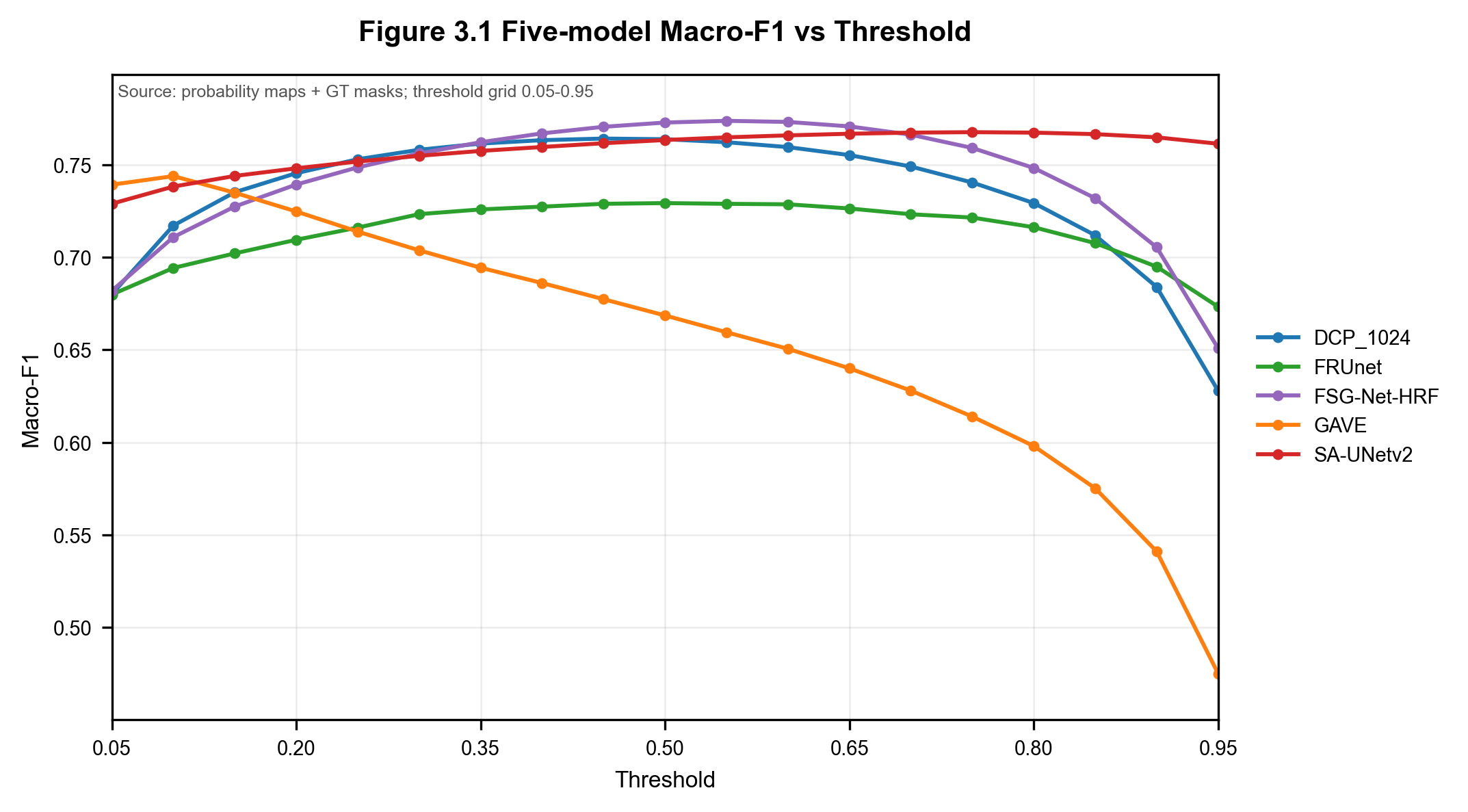}
    \caption{Sensitivity of equal-dataset Macro-F1 to a common single-threshold operating point.}
    \label{fig:image14}
\end{figure}

To assess whether the retained benchmark operating points were located in unusually sensitive regions, a single-threshold sensitivity analysis was performed using the available probability maps. This analysis is diagnostic only and does not replace the benchmark threshold policy described in Section~\ref{benchmark-threshold-policy-and-operating-points}. Figure~\ref{fig:image14} shows the resulting equal-dataset Macro-F1 response to a common threshold.

FSG-Net, SA-UNetv2, and DCP-1024 showed limited variation around 0.50, whereas GAVE was markedly more threshold-sensitive. This supports retaining the pipeline-specific thresholding procedures described in Section~\ref{benchmark-threshold-policy-and-operating-points}. The principal model-selection conclusions were not materially changed by the tested threshold variation.


\section{Configuration Sensitivity and Selection Stability}
\label{appendix-c.-configuration-sensitivity-and-selection-stability}

Configuration sensitivity was examined for the three pipelines with reproducible local training routes as a bounded robustness check rather than an exhaustive hyperparameter search. Model-specific factors were selected according to the original training mechanisms of each pipeline. For SA-UNetv2, the sensitivity analysis focused on DropBlock rate and scheduler patience; learning-rate screening did not identify a better operating point and was therefore not expanded further. For FSG-Net, weight decay and crop size were varied, whereas FR-UNet was evaluated under alternative dropout rates and batch sizes. Table~\ref{tab:configuration_factors} lists the tested factors and their rationale. Tables~\ref{tab:saunetv2_sensitivity}--\ref{tab:frunet_sensitivity} report the corresponding dataset-level Macro-F1 results.

\begin{table}[H]
\caption{Tested configuration factors and their rationale. DCP-1024 and GAVE
were not included in additional configuration sweeps because reproducible
local training routes were unavailable.}
\label{tab:configuration_factors}
\footnotesize
\begin{tabularx}{\textwidth}{
@{}
>{\raggedright\arraybackslash}p{0.18\textwidth}
>{\raggedright\arraybackslash}p{0.32\textwidth}
>{\raggedright\arraybackslash}X
@{}
}
\toprule
\textbf{Pipeline} &
\textbf{Tested factors} &
\textbf{Rationale} \\
\midrule
SA-UNetv2 &
DropBlock rate; scheduler patience &
Tests regularization strength and the timing of learning-rate reduction. \\
\addlinespace
FSG-Net &
Weight decay; crop size &
Tests regularization strength and the effect of spatial training context. \\
\addlinespace
FR-UNet &
Dropout rate; batch size &
Tests regularization strength and batch-dependent optimization behavior. \\
\bottomrule
\end{tabularx}
\end{table}

\begin{table}[H]
\caption{SA-UNetv2 configuration sensitivity across the six evaluation datasets.
Fixed settings included a $1028\times1028$ input size, batch size 2,
$0.5\,\mathrm{BCE}+0.5\,\mathrm{MCC}$ loss, DropBlock block size 7,
no data augmentation, and split seed 23.}
\label{tab:saunetv2_sensitivity}
\footnotesize
\setlength{\tabcolsep}{5pt}
\begin{tabular}{llllllll}
\toprule
\textbf{Configuration} &
\textbf{AFIO} &
\textbf{CHASEDB1} &
\textbf{DualModal2019} &
\textbf{HRF} &
\textbf{LES} &
\textbf{STARE} &
\makecell{\textbf{Equal-dataset}\\\textbf{Macro-F1}} \\
\midrule
DropBlock $=0.05$ & 0.6827 & 0.7906 & 0.7413 & 0.7272 & 0.8234 & 0.8068 & 0.7620 \\
DropBlock $=0.10$ & 0.6778 & 0.7824 & 0.7369 & 0.7224 & 0.8249 & 0.8069 & 0.7586 \\
Patience $=6$ & 0.6917 & 0.7911 & 0.7387 & 0.7257 & 0.8224 & 0.8089 & \textbf{0.7631} \\
Patience $=14$ & 0.6841 & 0.7890 & 0.7412 & 0.7273 & 0.8223 & 0.8048 & 0.7614 \\
\bottomrule
\end{tabular}
\end{table}

Relative to the SA-UNetv2 benchmark result reported in Section~\ref{evidence-guided-model-selection} (Macro-F1 approximately 0.76), the tested configurations produced only small changes, with equal-dataset Macro-F1 ranging from 0.7586 to 0.7631. The variation was therefore insufficient to materially change its benchmark-level selection role.

\begin{table}[H]
\caption{FSG-Net configuration sensitivity across the six evaluation datasets.
The source-aligned setting used weight decay 0.05 and $288\times288$ random crops.}
\label{tab:fsgnet_sensitivity}
\footnotesize
\setlength{\tabcolsep}{5pt}
\begin{tabular}{lccccccc}
\toprule
\textbf{Configuration} &
\textbf{AFIO} &
\textbf{CHASEDB1} &
\textbf{DualModal2019} &
\textbf{HRF} &
\textbf{LES} &
\textbf{STARE} &
\makecell{\textbf{Equal-dataset}\\\textbf{Macro-F1}} \\
\midrule
WD $=0.01$ & 0.6984 & 0.7787 & 0.7252 & 0.8002 & 0.8239 & 0.8012 & 0.7713 \\
WD $=0.10$ & 0.6899 & 0.7836 & 0.7317 & 0.8025 & 0.8309 & 0.8081 & 0.7744 \\
Crop $=224\times224$ & 0.6925 & 0.7694 & 0.7270 & 0.7932 & 0.8175 & 0.7867 & 0.7644 \\
Crop $=352\times352$ & 0.6967 & 0.7906 & 0.7302 & 0.8032 & 0.8295 & 0.8152 & \textbf{0.7776} \\
\bottomrule
\end{tabular}
\end{table}

Relative to the FSG-Net benchmark result in Section~\ref{evidence-guided-model-selection} (Macro-F1 approximately 0.77), the tested variants remained within a similar performance range (0.7644--0.7776). Crop size produced the largest variation, but FSG-Net remained one of the leading pipelines under the tested configurations.

\begin{table}[H]
\caption{FR-UNet configuration sensitivity across the six evaluation datasets.
The source-aligned setting used a dropout rate of 0.20 and batch size 4.}
\label{tab:frunet_sensitivity}
\footnotesize
\setlength{\tabcolsep}{5pt}
\begin{tabular}{lccccccc}
\toprule
\textbf{Configuration} &
\textbf{AFIO} &
\textbf{CHASEDB1} &
\textbf{DualModal2019} &
\textbf{HRF} &
\textbf{LES} &
\textbf{STARE} &
\makecell{\textbf{Equal-dataset}\\\textbf{Macro-F1}} \\
\midrule
Dropout $=0.10$ & 0.6944 & 0.7869 & 0.7351 & 0.7393 & 0.8197 & 0.8126 & \textbf{0.7646} \\
Dropout $=0.30$ & 0.7014 & 0.7674 & 0.7017 & 0.7066 & 0.8091 & 0.7937 & 0.7466 \\
Batch size $=2$ & 0.6818 & 0.7885 & 0.7185 & 0.7291 & 0.8190 & 0.7863 & 0.7539 \\
\bottomrule
\end{tabular}
\end{table}

FR-UNet showed greater configuration sensitivity than the two leading pipelines. Relative to its Section~\ref{evidence-guided-model-selection} benchmark result (Macro-F1 approximately 0.73), favorable settings improved equal-dataset Macro-F1 to as high as 0.7646. However, this improvement was still insufficient to exceed the best benchmark-level performance achieved by FSG-Net.


\section{Metric Definitions and Cross-Dataset Aggregation Rules}
\label{appendix-d.-metric-definitions-and-cross-dataset-aggregation-rules}

This appendix provides the formal metric definitions, benchmark settings, and aggregation rules used in Section~\ref{evaluation-metrics-and-interpretation}. Table~\ref{tab:pixel_metric_definitions} defines the pixel-level measures and their benchmark use. Table~\ref{tab:thin_structural_metrics} gives the thin-vessel and structural-fidelity definitions and settings, and Table~\ref{tab:aggregation_rules} records the image-, dataset-, and cross-dataset aggregation rules.

\begin{table}[H]
\caption{Pixel-level metric definitions and their use in the benchmark.}
\label{tab:pixel_metric_definitions}
\scriptsize
\begin{tabularx}{\textwidth}{
@{}
>{\raggedright\arraybackslash}p{0.08\textwidth}
>{\raggedright\arraybackslash}p{0.37\textwidth}
>{\raggedright\arraybackslash}X
@{}
}
\toprule
\textbf{Metric} &
\textbf{Definition} &
\textbf{Benchmark use} \\
\midrule
Sensitivity &
$\displaystyle
\frac{\mathrm{TP}}
{\mathrm{TP}+\mathrm{FN}}
$ &
Vessel recall; FN-oriented evidence. \\
\addlinespace
Precision &
$\displaystyle
\frac{\mathrm{TP}}
{\mathrm{TP}+\mathrm{FP}}
$ &
Prediction reliability; FP-oriented evidence. \\
\addlinespace
F1 &
$\displaystyle
\frac{2\,\mathrm{TP}}
{2\,\mathrm{TP}+\mathrm{FP}+\mathrm{FN}}
$ &
Primary pixel-level agreement measure. \\
\addlinespace
$F_{\beta}$ &
$\displaystyle
F_{\beta}
=
(1+\beta^2)
\frac{PR}{\beta^2 P+R}
$ &
Scenario-dependent precision--recall weighting; $\beta>1$ places greater emphasis on recall and therefore on reducing FN errors. \\
\addlinespace
IoU &
$\displaystyle
\frac{\mathrm{TP}}
{\mathrm{TP}+\mathrm{FP}+\mathrm{FN}}
$ &
Overlap-based consistency check. \\
\addlinespace
MCC &
$\displaystyle
\frac{
\mathrm{TP}\,\mathrm{TN}-\mathrm{FP}\,\mathrm{FN}
}{
\sqrt{
(\mathrm{TP}+\mathrm{FP})
(\mathrm{TP}+\mathrm{FN})
(\mathrm{TN}+\mathrm{FP})
(\mathrm{TN}+\mathrm{FN})
}
}
$ &
Class-imbalance-aware consistency check. \\
\bottomrule
\end{tabularx}

\vspace{0.4em}
\begin{minipage}{\textwidth}
\tiny
TP, true positive; TN, true negative; FP, false positive; FN, false negative.
In the $F_{\beta}$ definition, $P$ denotes precision and $R$ denotes recall
(sensitivity).
\end{minipage}
\end{table}

\begin{table}[H]
\caption{Thin-vessel and structural-fidelity metric definitions and benchmark settings.}
\label{tab:thin_structural_metrics}
\scriptsize
\renewcommand{\arraystretch}{1.25}

\begin{tabularx}{\textwidth}{
@{}
>{\raggedright\arraybackslash}p{0.17\textwidth}
>{\raggedright\arraybackslash}p{0.38\textwidth}
>{\raggedright\arraybackslash}X
@{}
}
\toprule
\textbf{Metric} &
\textbf{Definition} &
\textbf{Benchmark setting and interpretation} \\
\midrule

Thin-Vessel Sensitivity (TVS) &
\(
\begin{aligned}
D_G(x) &= 2\,DT_G(x),\\
\tau_{\mathrm{thin}}
&= \max\!\left(1\,\mathrm{px},\,0.018D_{\mathrm{OD}}\right),\\
T_G^{\mathrm{thin}}
&= \left\{x\in S_G:
D_G(x)\leq\tau_{\mathrm{thin}}\right\},\\
\mathrm{TVS}
&=
\frac{\left|T_G^{\mathrm{thin}}\cap P\right|}
     {\left|T_G^{\mathrm{thin}}\right|}.
\end{aligned}
\)
&
GT-conditioned recovery of fine vessels defined on the ground-truth skeleton.
The diameter threshold is normalized by the optic-disc diameter
\(D_{\mathrm{OD}}\), with a minimum threshold of \(1\) pixel. \\

\addlinespace

clDice &
\(
\begin{aligned}
T_{\mathrm{prec}}
&=
\frac{|S_P\cap G|}{|S_P|},\\
T_{\mathrm{sens}}
&=
\frac{|S_G\cap P|}{|S_G|},\\
\mathrm{clDice}
&=
\frac{2T_{\mathrm{prec}}T_{\mathrm{sens}}}
     {T_{\mathrm{prec}}+T_{\mathrm{sens}}}.
\end{aligned}
\)
&
Bidirectional skeleton-to-region containment, following the clDice formulation \cite{shit2021cldice}. \\

\addlinespace

Tolerant Skeleton F1 (SF1) &
\(
\begin{aligned}
P_r
&=
\frac{
\left|
\left\{p\in S_P:
d(p,S_G)\leq r\right\}
\right|
}
{|S_P|},\\
R_r
&=
\frac{
\left|
\left\{g\in S_G:
d(g,S_P)\leq r\right\}
\right|
}
{|S_G|},\\
\mathrm{SF1}_r
&=
\frac{2P_rR_r}{P_r+R_r}.
\end{aligned}
\)
&
Symmetric tolerance-based skeleton alignment. The main analysis used
\(r=2\,\mathrm{px}\). \\

\bottomrule
\end{tabularx}
\end{table}

\begin{table}[H]
\caption{Image-level, dataset-level, and cross-dataset aggregation rules.}
\label{tab:aggregation_rules}
\scriptsize
\renewcommand{\arraystretch}{1.2}

\begin{tabularx}{\textwidth}{
@{}
>{\raggedright\arraybackslash}p{0.24\textwidth}
>{\raggedright\arraybackslash}p{0.38\textwidth}
>{\raggedright\arraybackslash}X
@{}
}
\toprule
\textbf{Summary} &
\textbf{Definition} &
\textbf{Interpretation} \\
\midrule

Dataset-level mean &
\(
\displaystyle
\bar{x}_{md}
=
\frac{1}{n_d}
\sum_{i=1}^{n_d} x_{mdi}
\)
&
Average across images within a given model--dataset pair. \\

\addlinespace

Equal-dataset cross-dataset mean &
\(
\displaystyle
\bar{x}_{m}
=
\frac{1}{D}
\sum_{d=1}^{D}\bar{x}_{md}
\)
&
Assigns equal weight to each dataset regardless of its number of images. \\

\addlinespace

Within-dataset image-level SD &
\(
\displaystyle
s_{md}
=
\sqrt{
\frac{1}{n_d-1}
\sum_{i=1}^{n_d}
\left(x_{mdi}-\bar{x}_{md}\right)^2
}
\)
&
Measures image-level variability within a given model--dataset pair. \\

\addlinespace

Cross-dataset SD &
\(
\displaystyle
s_m
=
\sqrt{
\frac{1}{D-1}
\sum_{d=1}^{D}
\left(\bar{x}_{md}-\bar{x}_{m}\right)^2
}
\)
&
Measures variability across the dataset-level means. \\

\addlinespace

Worst-dataset result &
\(
\displaystyle
x_m^{\mathrm{worst}}
=
\min_{d}\bar{x}_{md}
\)
&
Represents the lowest observed dataset-level performance for model \(m\). \\

\bottomrule
\end{tabularx}
\end{table}


\section{Decision-Parameter Sensitivity Analysis}
\label{appendix-e.-decision-parameter-sensitivity-analysis}

This appendix evaluates whether the benchmark-relative recommendations depend on the prespecified decision parameters. It recomputed scores from the completed model-dataset utilities only; model outputs, checkpoints, and the source-preserving threshold policy were unchanged.

\subsection{Sensitivity to the Structural Weight \(\eta\)}
\label{e.1-sensitivity-to-the-structural-weight-ux3b7}

Structural fidelity was calculated as \(S = \eta clDice + \left( 1 - \eta \right)SF1\). This analysis held the structural fidelity preset fixed (P/T/S/R = 0/0/100/0 and   \(\rho\) = 0.30) while   \(\eta\) varied from 0.00 to 1.00 in 0.01 increments.

\begin{figure}
    \centering
    \includegraphics[width=0.75\linewidth]{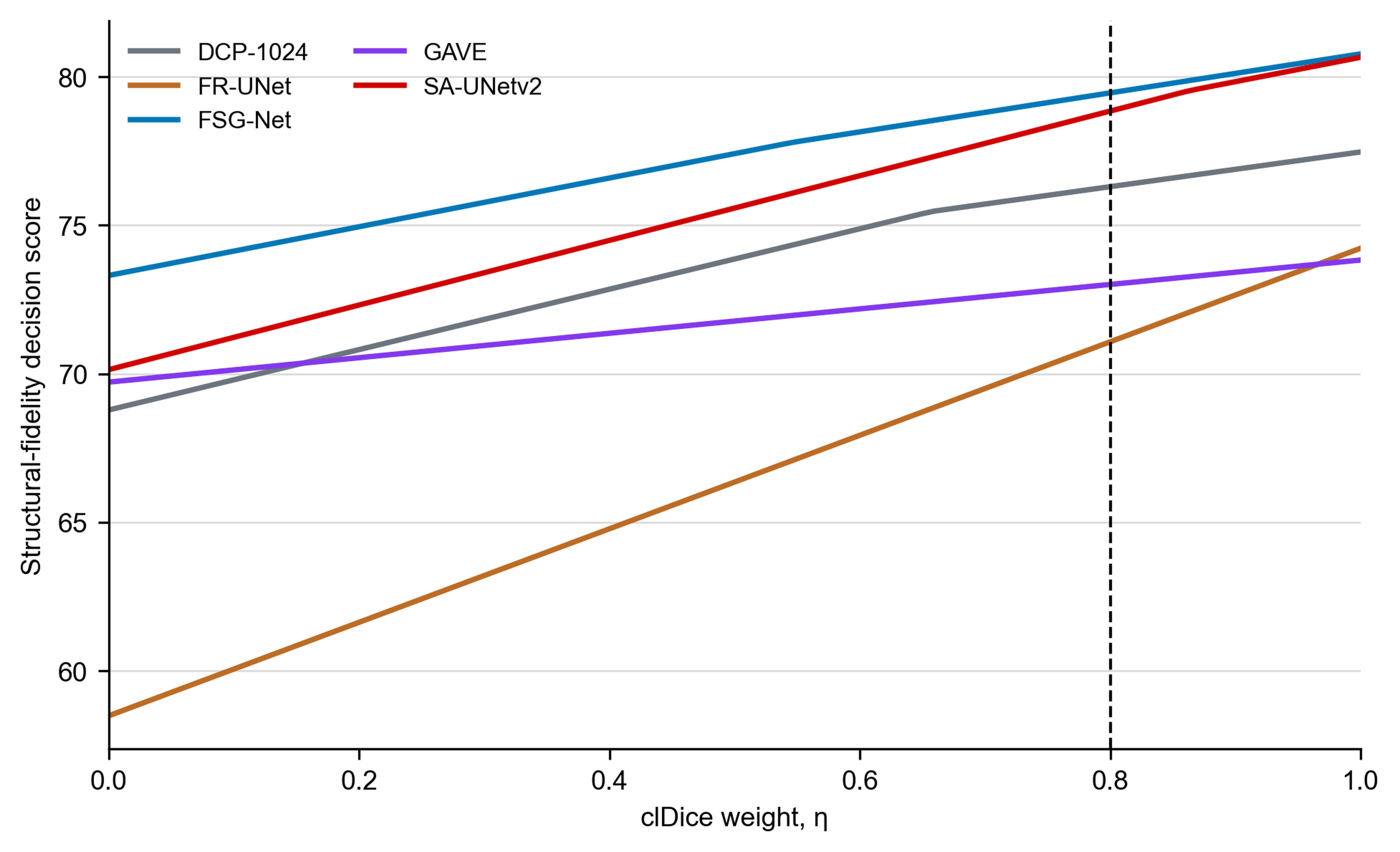}
    \caption{Sensitivity of structural-fidelity scores to the structural weight \(\eta\). All other structural-preset settings were held fixed, including \(\rho=0.30\), while \(\eta\) varied from 0 to 1. The vertical line indicates the prespecified value \(\eta=0.80\).}
    \label{fig:image15}
\end{figure}

As shown in Figure~\ref{fig:image15}, FSG-Net remained first and SA-UNetv2 remained second across the full \(\eta\) range. The FSG-Net--SA-UNetv2 score gap narrowed as greater weight was assigned to clDice. Thus, \(\eta\) affected the scores, but not the structural-fidelity recommendation under the fixed \(\rho=0.30\) aggregation.

\subsection{Sensitivity to the Cross-Dataset Risk Weight $\rho$}
\label{e.2-sensitivity-to-the-cross-dataset-risk-weight-ux3c1}

For each fixed objective, \(\rho\) varied from 0.00 to 1.00 in 0.01 increments. The calculations held \(\beta\), \(\eta\), and P/T/S/R fixed within each objective. Table~\ref{tab:rho_sensitivity} reports the top two pipelines at 0.1 intervals; bold rows mark the three operating levels exposed by the public tool (\(\rho=0.00\), 0.30, and 0.70).

\begin{table}
\caption{Sensitivity of the top-two model recommendations to the
cross-dataset risk weight $\rho$. Each cell reports the first- and
second-ranked models (Top 1 / Top 2).}
\label{tab:rho_sensitivity}
\scriptsize
\setlength{\tabcolsep}{4pt}

\begin{tabularx}{\textwidth}{
@{}
c
>{\centering\arraybackslash}X
>{\centering\arraybackslash}X
>{\centering\arraybackslash}X
>{\centering\arraybackslash}X
>{\centering\arraybackslash}X
@{}
}
\toprule
$\boldsymbol{\rho}$ &
\textbf{Mixed-objective} &
\textbf{Pixel performance} &
\textbf{Low-FN / thin vessels} &
\textbf{Structural fidelity} &
\textbf{Resource-oriented} \\
\midrule

\textbf{0.0} &
\textbf{SA-UNetv2 / FSG-Net} &
\textbf{FSG-Net / DCP-1024} &
\textbf{SA-UNetv2 / FSG-Net} &
\textbf{FSG-Net / SA-UNetv2} &
\textbf{SA-UNetv2 / GAVE} \\

0.1 & SA-UNetv2 / FSG-Net & FSG-Net / SA-UNetv2 & SA-UNetv2 / FSG-Net & FSG-Net / SA-UNetv2 & SA-UNetv2 / GAVE \\
0.2 & SA-UNetv2 / FSG-Net & FSG-Net / SA-UNetv2 & SA-UNetv2 / FSG-Net & FSG-Net / SA-UNetv2 & SA-UNetv2 / GAVE \\

\textbf{0.3} &
\textbf{SA-UNetv2 / FSG-Net} &
\textbf{FSG-Net / SA-UNetv2} &
\textbf{SA-UNetv2 / FSG-Net} &
\textbf{FSG-Net / SA-UNetv2} &
\textbf{SA-UNetv2 / GAVE} \\

0.4 & SA-UNetv2 / FSG-Net & FSG-Net / SA-UNetv2 & SA-UNetv2 / FSG-Net & FSG-Net / SA-UNetv2 & SA-UNetv2 / GAVE \\
0.5 & SA-UNetv2 / FSG-Net & FSG-Net / SA-UNetv2 & SA-UNetv2 / FSG-Net & FSG-Net / SA-UNetv2 & SA-UNetv2 / GAVE \\
0.6 & SA-UNetv2 / FSG-Net & SA-UNetv2 / FSG-Net & SA-UNetv2 / FSG-Net & FSG-Net / SA-UNetv2 & SA-UNetv2 / GAVE \\

\textbf{0.7} &
\textbf{SA-UNetv2 / FSG-Net} &
\textbf{SA-UNetv2 / FSG-Net} &
\textbf{SA-UNetv2 / FSG-Net} &
\textbf{FSG-Net / SA-UNetv2} &
\textbf{SA-UNetv2 / GAVE} \\

0.8 & SA-UNetv2 / FSG-Net & SA-UNetv2 / FSG-Net & SA-UNetv2 / FSG-Net & FSG-Net / SA-UNetv2 & SA-UNetv2 / GAVE \\
0.9 & SA-UNetv2 / FSG-Net & SA-UNetv2 / FSG-Net & SA-UNetv2 / FSG-Net & FSG-Net / SA-UNetv2 & SA-UNetv2 / GAVE \\
1.0 & SA-UNetv2 / FSG-Net & SA-UNetv2 / FSG-Net & SA-UNetv2 / FSG-Net & FSG-Net / SA-UNetv2 & SA-UNetv2 / GAVE \\

\bottomrule
\end{tabularx}

\vspace{0.4em}
\begin{minipage}{\textwidth}
\tiny
Fixed objective settings were as follows:
mixed-objective, $\beta=1$ and P/T/S/R $=50/15/20/15$;
pixel performance, $\beta=1$ and $100/0/0/0$;
low-FN/thin-vessel, $\beta=2$ and $50/50/0/0$;
structural fidelity, $\eta=0.80$ and $0/0/100/0$;
and resource-oriented, $\beta=1$ and $30/0/0/70$.
For objectives containing a structural component, $\eta$ was fixed at $0.80$.
Bold rows denote the selected reference values of $\rho$.
\end{minipage}
\end{table}

The mixed-objective, low-FN/thin-vessel, structural-fidelity, and resource-oriented recommendations retained the same top-ranked pipeline throughout the \(\rho\) sweep. Pixel performance changed from FSG-Net to SA-UNetv2 between \(\rho=0.58\) and 0.59. Accordingly, the mean-only pixel operating point (\(\rho=0.00\)) and the conservative unknown-domain operating point (\(\rho=0.70\)) represent distinct risk regimes within the same pixel-performance objective.


\section{Sensitivity of Model Rankings to Boundary Uncertainty}
\label{appendix-f.-sensitivity-of-model-rankings-to-boundary-uncertainty}

Equal-dataset Macro-F1 was recalculated after excluding pixels within 10\% and 20\% of the local GT vessel diameter from the annotated boundary. Predictions, reference masks, and aggregation were unchanged. Only the evaluated region differed. This analysis was limited to Macro-F1. Because each setting evaluates a different pixel region, the score increases should not be interpreted as performance gains.

\begin{figure}
    \centering
    \includegraphics[width=0.75\linewidth]{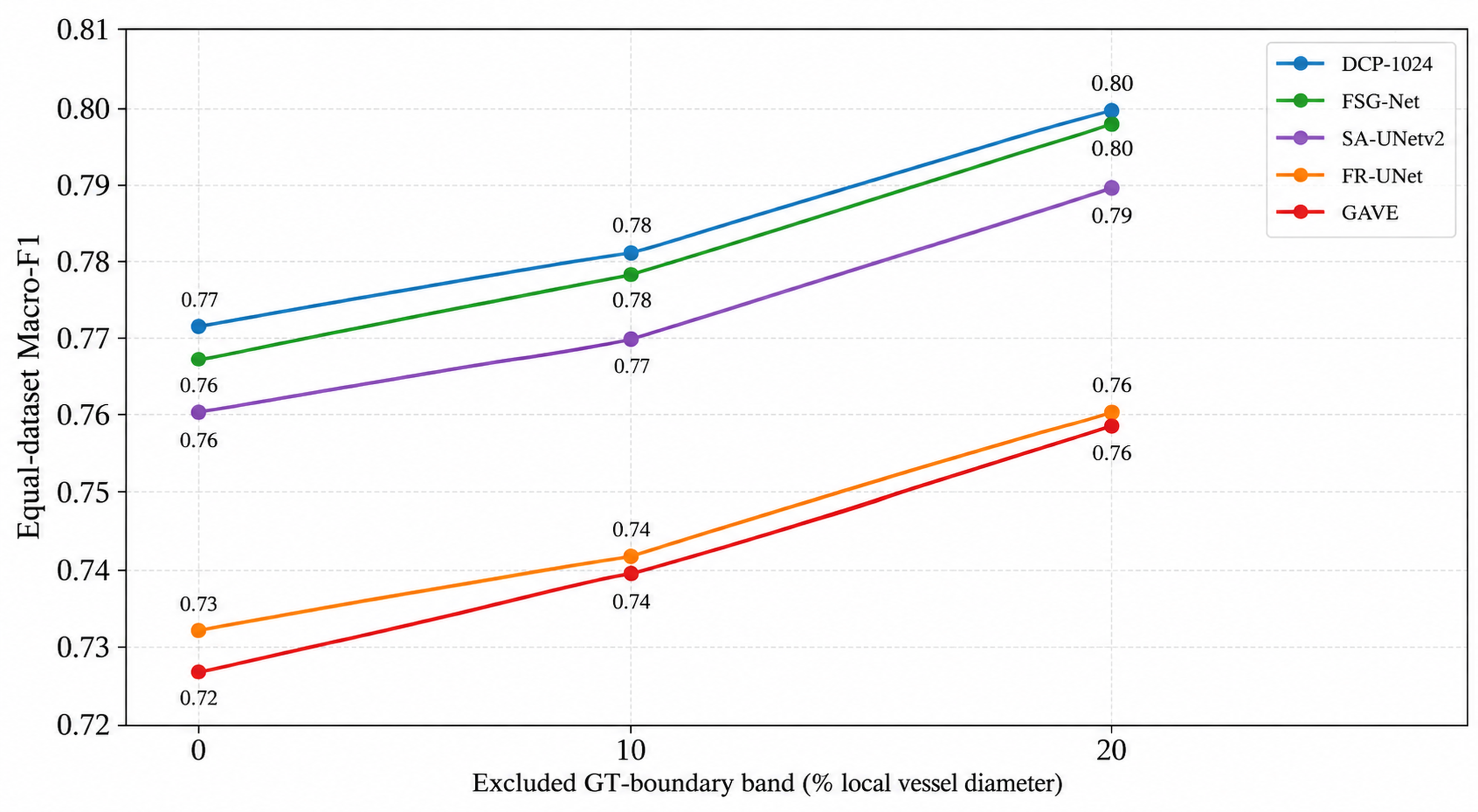}
    \caption{Sensitivity of equal-dataset Macro-F1 rankings to GT-boundary-band exclusion.}
    \label{fig:image16}
\end{figure}

As shown in Figure~\ref{fig:image16}, macro-F1 increased for all five pipelines as the excluded boundary band widened. This consistent increase indicates that boundary-adjacent pixels were a common source of measured disagreement. However, the ranking remained unchanged at 0\%, 10\%, and 20\%. The result is therefore consistent with GT-boundary uncertainty affecting absolute scores, but not the model order under the tested settings.


\section{Error-Aware Oracle Vote Fusion Details}
\label{appendix-g.-error-aware-oracle-vote-fusion-details}

\begin{figure}[H]
    \centering
    \includegraphics[width=0.75\linewidth]{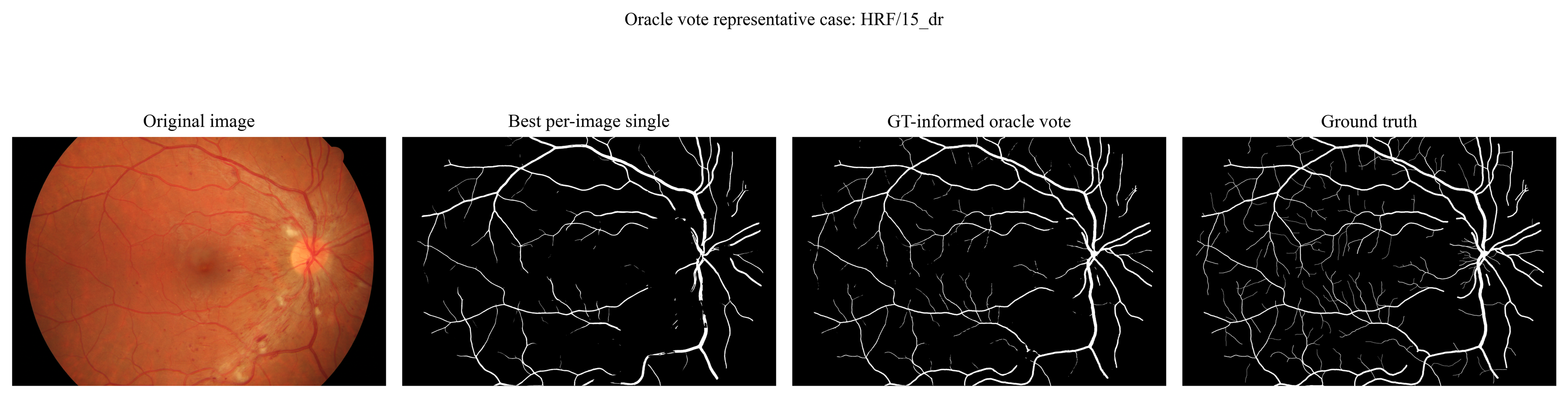}
    \caption{Oracle vote versus the best single model on HRF/15\_dr.}
    \label{fig:image17}
\end{figure}

The error-aware voting experiment estimates a GT-informed upper bound on local model complementarity. Because vote eligibility is determined using GT, the procedure is diagnostic rather than deployable. Across 244 images, oracle voting increased mean per-image F1 from 0.76 to 0.83 and IoU from 0.61 to 0.73 as shown in Table~\ref{tab:oracle_vote_fusion}. Figure~\ref{fig:image17} provides a representative example of the local complementarity captured by the oracle.

\begin{table}[H]
\caption{Aggregate performance of GT-informed oracle vote fusion across all 244 images.
The baseline uses the best-performing single pipeline selected independently for each image.}
\label{tab:oracle_vote_fusion}
\scriptsize
\setlength{\tabcolsep}{4pt}

\begin{tabularx}{\textwidth}{
@{}
>{\raggedright\arraybackslash}p{0.13\textwidth}
>{\raggedright\arraybackslash}p{0.15\textwidth}
>{\centering\arraybackslash}p{0.09\textwidth}
>{\centering\arraybackslash}p{0.12\textwidth}
>{\centering\arraybackslash}p{0.11\textwidth}
>{\centering\arraybackslash}p{0.09\textwidth}
>{\centering\arraybackslash}p{0.09\textwidth}
>{\centering\arraybackslash}X
@{}
}
\toprule

\textbf{Scope} &
\textbf{Comparison baseline} &
\textbf{Baseline F1} &
\textbf{Vote F1} &
\textbf{Mean $\Delta$F1} &
\textbf{Baseline IoU} &
\textbf{Vote IoU} &
\textbf{Error change} \\

\midrule

All datasets (244 images) &
Per-image best single pipeline &
0.76 &
0.83 &
$+0.08$ &
0.61 &
0.73 &
$-10.74$ M \\

\bottomrule
\end{tabularx}
\end{table}

For example HRF/15\_dr, F1 increased from 0.73 to 0.91 and the error count decreased from 252,810 to 88,464.


\section{Laplacian Oracle Repair of Known FN/FP Pixels}
\label{appendix-h.-laplacian-oracle-repair-of-known-fnfp-pixels}

This oracle analysis tested whether local image intensity could correct GT-identified segmentation errors. GT was used to locate and label FN and FP pixels. For each single error pixel, the four-neighbour Laplacian response was calculated from the original image as

\begin{equation}
D_{h}(x,y)
= 0.25
\left(
I(x-h,y)+I(x+h,y)+I(x,y-h)+I(x,y+h)\right)
-I(x,y).
\end{equation}

At the retained setting (\(h=1,\tau=0\)), a known FN pixel was restored as a vessel when \(D_h>\tau\), whereas a known FP pixel was removed when \(D_h\leq\tau\). All other pixels remained unchanged. This analysis measures correction potential when the error location and type are known. It is not a GT-free post-processing method.

Table~\ref{tab:laplacian_oracle_repair} reports the dataset-level improvements and correction rates. Mean F1 increased by 0.08--0.13, and mean IoU increased by 0.12--0.17. FN recovery ranged from 37.4\% to 55.6\%. FP removal ranged from 44.4\% to 52.3\%.

\begin{table}[H]
\centering
\caption{Dataset-level performance of GT-informed Laplacian oracle repair.}
\label{tab:laplacian_oracle_repair}
\scriptsize
\setlength{\tabcolsep}{8pt}
\begin{tabular}{lccccc}
\toprule
\textbf{Dataset} &
\textbf{Images} &
\textbf{$\Delta$F1} &
\textbf{$\Delta$IoU} &
\textbf{FN recovered} &
\textbf{FP removed} \\
\midrule
AFIO & 99 & $+0.13$ & $+0.17$ & 37.4\% & 51.0\% \\
CHASEDB1 & 28 & $+0.10$ & $+0.14$ & 55.6\% & 44.4\% \\
DualModal2019 & 30 & $+0.10$ & $+0.14$ & 54.4\% & 44.7\% \\
HRF & 45 & $+0.11$ & $+0.16$ & 48.7\% & 52.3\% \\
LES & 22 & $+0.09$ & $+0.13$ & 50.3\% & 51.2\% \\
STARE & 20 & $+0.08$ & $+0.12$ & 45.7\% & 45.8\% \\
\bottomrule
\end{tabular}
\end{table}


\section{clDice-Augmented Loss Feasibility Experiment}
\label{appendix-i.-cldice-augmented-loss-feasibility-experiment}

clDice remains a primary structural evaluation metric in this manuscript. The experiment results in Table~\ref{tab:cldice_loss_feasibility} ask a narrower question: whether adding a soft-clDice term to the training loss improves the HRF result under the recorded local setup. Its negative result is not evidence against using clDice as the principal structural benchmark.
The distinction is intentional: clDice as a benchmark metric measures connectivity-sensitive structure, whereas a clDice-augmented loss changes the optimization objective and can affect convergence, calibration, and class-balance trade-offs. The HRF test only addresses the latter.

\begin{table}
\caption{HRF feasibility test for incorporating soft-clDice into the training loss.}
\label{tab:cldice_loss_feasibility}
\footnotesize

\begin{tabularx}{\textwidth}{
    @{}
    >{\raggedright\arraybackslash}p{0.28\textwidth}
    >{\raggedright\arraybackslash}X
    @{}
}
\toprule
\textbf{Item} & \textbf{Recorded value} \\
\midrule
Loss under test &
$0.4\,\mathrm{BCE} + 0.4\,\mathrm{MCC} + 0.2\,\mathrm{soft\text{-}clDice}$ \\
Run/configuration &
$1028 \times 1028$ input; batch size 2; split seed 23 \\
Evaluation set &
HRF; 45 matched images \\
Pooled F1 &
0.7126115599 \\
Mean image-level F1 &
0.7083935962 \\
Interpretation &
Retained as a negative feasibility result for modifying the loss function under this setup;
clDice remains the core structural evaluation metric. \\
\bottomrule
\end{tabularx}
\end{table}


\section{Supplementary Downstream Feature-Level Analyses}
\label{appendix-j.-supplementary-downstream-feature-level-analyses}

This appendix provides supplementary downstream classification analyses, including the full classifier implementation, the feature-ranking table, intermediate-state visualizations, perturbation protocols, stabilization parameters, and complete robust-feature results.

\subsection{Detailed Classifier Implementation and Image-Level Reference}
\label{j.1-detailed-classifier-implementation-and-image-level-reference}

To position the proposed segmentation-to-classification analysis against a recognized image-level retinal classifier, we used RETFound as a reference method. RETFound is a retinal foundation model pretrained with masked autoencoding on large-scale unlabeled retinal images \cite{zhou2023retfound}.

In our implementation, each color fundus image was resized to $224\times224$ pixels and normalized before being passed through the pretrained RETFound encoder. The encoder parameters were frozen, and the mean of the classification token and patch tokens was used as the image representation. After feature standardization, a multiclass logistic-regression classifier with elastic-net regularization was evaluated using LOOCV. Unlike the vessel-based pathway, RETFound operates directly on the full fundus image and does not explicitly require vessel masks or derive vessel-area, centerline, graph, or caliber features.

The feature-based classifier used in this study has a different advantage. It represents the classification input explicitly as a set of measurable vascular quantities rather than as an implicit deep embedding. A segmentation mask is converted into a skeleton/centerline, graph, and diameter representation, after which ring-wise vascular features are calculated and supplied to the classifier. Consequently, a prediction can be traced to a particular feature, optic-disc-diameter ring, and vascular property, such as vessel-area fraction, caliber, centerline curvature, or graph connectivity. This explicit representation makes it possible to examine how a segmentation error changes an intermediate feature and how that change affects the final decision, which cannot be isolated directly from a global RETFound embedding.

On the same 45 HRF images, RETFound achieved an LOOCV accuracy of 38/45 = 84.44\%. When the complete set of numerically valid features extracted from the GT masks was used, excluding image identifiers, labels, status fields, and ring-coordinate metadata, the feature-based classifier achieved 43/45 = 95.56\%. Thus, the explicit feature route is not intrinsically less accurate than the image-level reference; with a sufficiently rich feature set, it produced a higher accuracy under the present experiment. The complete-feature result should nevertheless be interpreted as an extended GT-based reference, because it uses a substantially broader feature set than the vascular-only analysis.

The subsequent analysis therefore focuses on the 11 vascular features that are directly linked to the segmentation mask, skeleton, graph, and vessel caliber. This restriction is not intended to maximize the overall classification accuracy. Instead, it creates a controlled and interpretable setting for isolating the effect of segmentation: only the input mask is changed among GT, FSG, DCP, GAVE, FRUNet, and SAUNet, while the original image, feature definitions, ring partition, and classification protocol are held constant. The analysis can then follow the complete chain from mask error to intermediate representation, feature sensitivity, feature importance, and classification error.

Using only the 11 vascular features, the original GT classification accuracy was 33/45 = 73.33\%, and the accuracy after the selected stabilization procedures was 35/45 = 77.78\%. These results are not presented as evidence that the vascular-only subset outperforms RETFound. Rather, they show why the subset is useful for the central question of this study: RETFound provides a strong image-level reference, whereas the explicit vascular-feature pathway allows the effects of segmentation differences on intermediate vascular features and downstream classification to be examined directly.

\subsection{Detailed Explanation of Pixel--Classification Mismatch}
\label{j.2-detailed-explanation-of-pixelclassification-mismatch}

The apparent GAVE advantage over GT in the unprocessed vascular-only baseline (34/45 versus 33/45) should not be interpreted as evidence that a predicted mask is superior to the reference annotation. The classifier does not operate directly on mask pixels; instead, each mask is converted into centerline, graph, and diameter representations and summarized as \(9\times11\) ring-wise vascular features. Within each LOOCV fold, feature processing and the multiclass logistic-regression classifier are fitted separately for each segmentation method, so each method induces its own feature distribution and decision boundary. A mask may be closer to GT in pixel overlap yet shift a sensitive feature in a direction that reduces classification accuracy, whereas a less GT-like mask may accidentally preserve or even enhance class separation in feature space. The 34/45 versus 33/45 difference should therefore be interpreted as a small-sample downstream feature effect rather than evidence that GAVE produces superior segmentations.

The later reversal after feature-specific stabilization (GT 35/45 versus GAVE 25/45) reinforces this interpretation: changing the feature definitions altered the representation learned by the classifier, even though the underlying images and class labels remained unchanged. These observations separate two questions that are easy to conflate: what the classifier learns from each method-specific feature distribution, and why a local mask error can change a feature so strongly. The next subsections address these questions in that order.


To avoid basing the conclusion on a single misclassified image, we performed vascular-only LOOCV over all 45 images for each of the six segmentation inputs and recorded the mean absolute model coefficient across folds. Table~\ref{tab:top_vascular_features} summarizes each feature type across its nine rings by averaging these coefficient magnitudes and ranks feature types rather than individual ring coordinates. Values in parentheses indicate the mean coefficient magnitude for each feature type across the nine rings.
What the table reports are feature types used by the classifier in the coordinate system generated by each mask; it is not a ranking of segmentation quality. GT relies more strongly on VAF, VCmean, and Dratio, whereas FSG, DCP, FRUNet, GAVE, and SAUNet show different combinations of caliber, graph, and centerline-geometry dependence. This method-specific dependence is direct evidence that pixel-level segmentation quality cannot be translated into classification performance without considering the feature representation.

The apparent randomness of classification outcomes is therefore not pure randomness. It is produced by the combination of a small sample, method-specific feature distributions, unequal feature scales, LOOCV decision boundaries that vary with the training sample, and feature sensitivity. Pixel-level ranking is more likely to transfer to classification only when the feature-extraction step preserves reproducible class differences.

\subsection{Detailed Mask, Skeleton, Graph, and Diameter Representations}
\label{j.3-detailed-mask-skeleton-graph-and-diameter-representations}

A binary vessel mask can be transformed into three linked representations. Skeletonization reduces the vessel region to an approximately one-pixel-wide centerline, from which an 8-neighbor graph captures connectivity, branch points, and endpoints. In parallel, the Euclidean Distance Transform (EDT) estimates local vessel radius, with diameter defined as $2\times\operatorname{EDT}(x)$, enabling caliber-related measurements along the centerline.

These representations are sequentially dependent on the mask. Small boundary changes can remove thin branches, create or break graph connections, and alter local diameter estimates. As a result, minor pixel-level segmentation errors may produce disproportionately large changes in connectivity, curvature, caliber, and thin-vessel features.

Pixel errors can be selectively amplified by different intermediate representations. Errors in vessel interiors mainly affect vessel area fraction (VAF), whereas errors at thin branches or endpoints can alter skeleton connectivity, boundary shifts can change EDT-derived caliber estimates, and irregular centerlines can strongly affect curvature and directional variation. Thus, pixel-level overlap alone does not indicate how an error will propagate into downstream vascular features. This representation-dependent propagation helps explain the observed mismatch between segmentation and classification performance. The following visualizations illustrate this pathway before it is quantified in the perturbation analysis.

For the detailed analyses of cases 02\_g, 10\_dr, and 09\_dr, intermediate states of feature computation were visualized directly. The examples were not selected arbitrarily: they correspond to features that recur in the top-five rankings across the six methods while representing three distinct computational mechanisms---VAF for mask area, graph\_edge\_count for topology, and caliber\_diam\_mm\_IQR for EDT-based diameter sampling. Centerline geometry is explained in the corresponding pointwise centerline plots. This selection links model-importance evidence to intermediate-process evidence. The mechanism figures explain how a feature is generated; they are not intended to rank methods from a single image. Method ranking remains based on the full-image LOOCV results and the structure-level benchmark.

\begin{figure}
    \centering
    \includegraphics[width=0.75\linewidth]{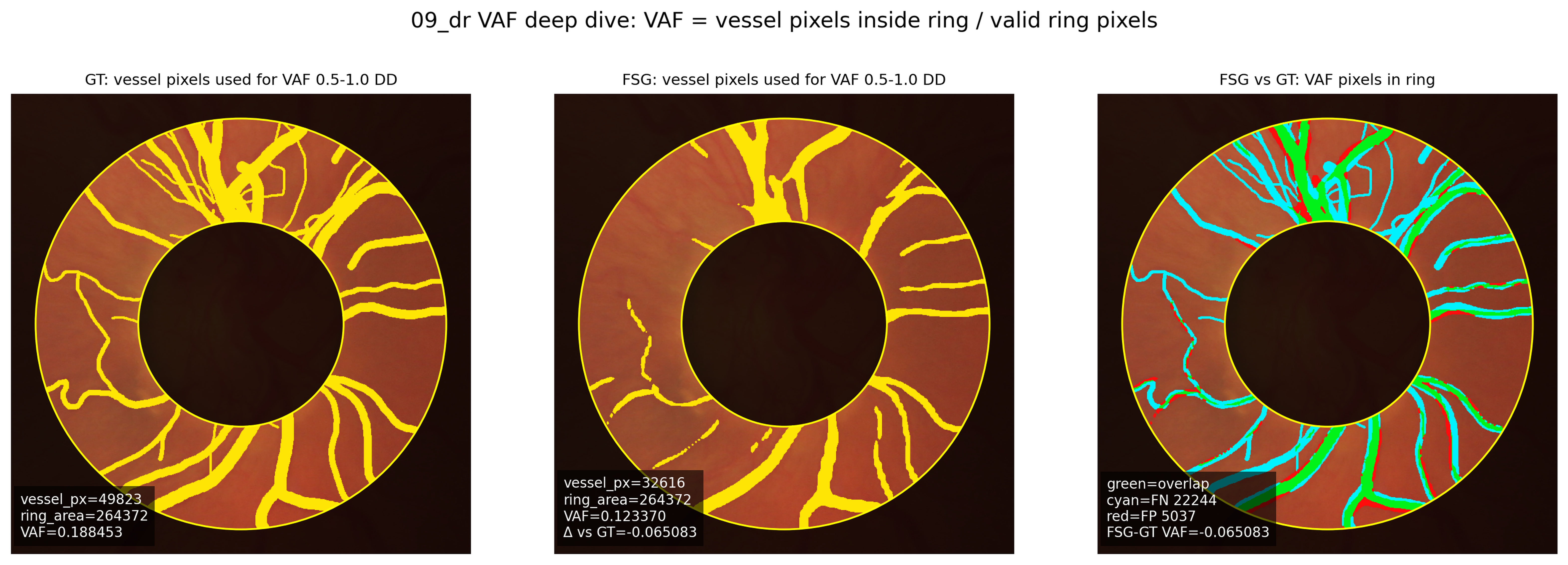}
    \caption{Ring-wise mask mechanism for VAF. Green denotes vessel pixels shared by GT and FSG, cyan denotes false negatives, and red denotes false positives. VAF is the number of vessel pixels inside the ring divided by the valid ring area.}
    \label{fig:image18}
\end{figure}

Figure~\ref{fig:image18} displays the numerator and denominator of VAF directly. False-positive pixels increase the vessel-area numerator, whereas false-negative pixels decrease it. The feature is therefore sensitive to total area, but it does not distinguish whether those pixels lie at a bifurcation or in a region of high curvature.

\begin{figure}
    \centering
    \includegraphics[width=0.75\linewidth]{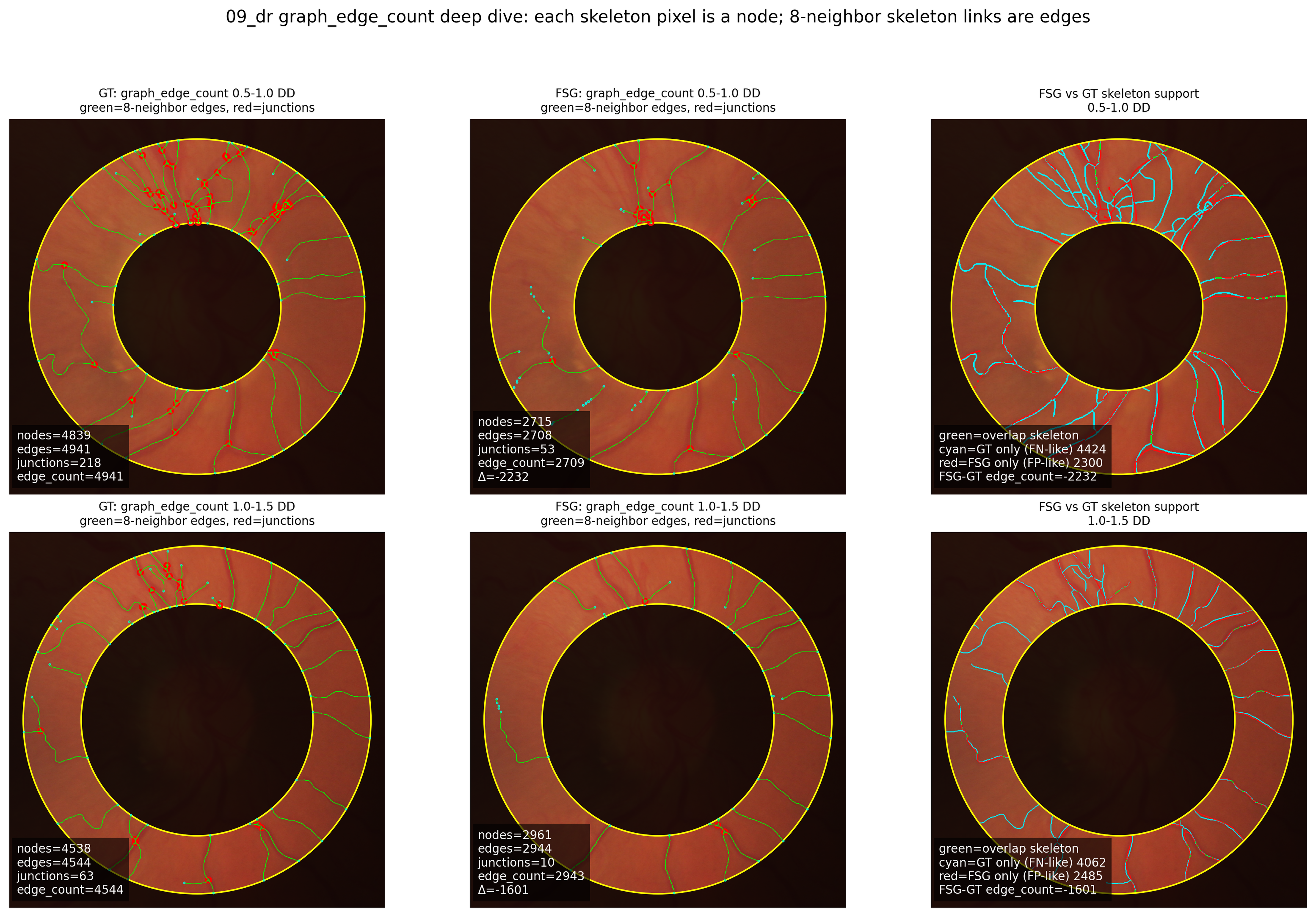}
    \caption{Pointwise interpretation of graph\_edge\_count and the skeleton graph. The red markers identify junction nodes, and graph edges link neighboring skeleton pixels in the 8-neighborhood. The edge count is the total number of such links within the selected ring.}
    \label{fig:image19}
\end{figure}

In Figure~\ref{fig:image19}, the thin green line is the skeleton and the red markers identify junction nodes. Nodes within the selected ring are connected to neighboring skeleton pixels in their 8-neighborhood. A short gap removes connections from neighboring nodes, whereas a spur adds an endpoint and a short edge. The visualization therefore decomposes the graph\_edge\_count feature into the local links that contribute to the connectivity count.

\begin{figure}
    \centering
    \includegraphics[width=0.75\linewidth]{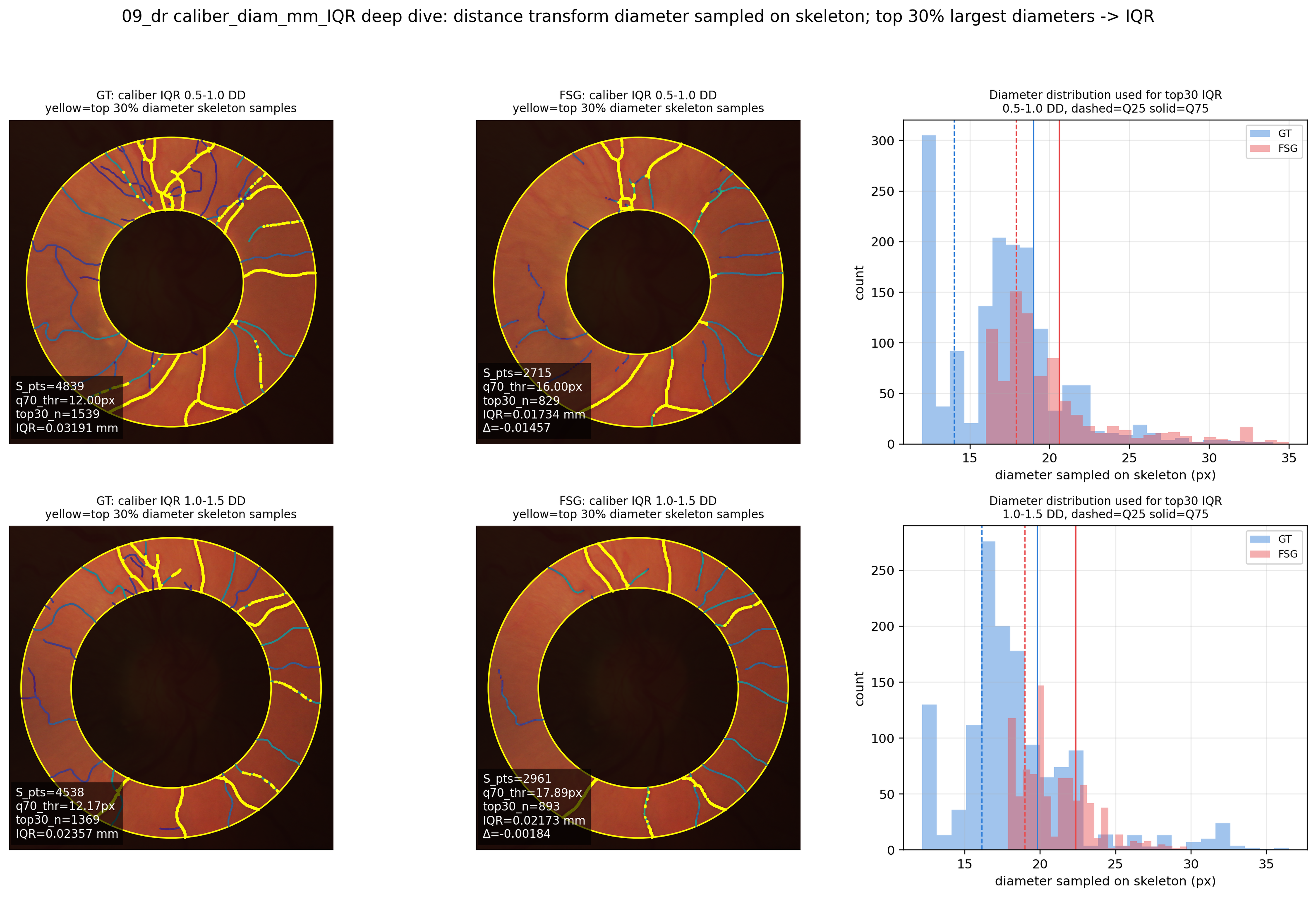}
    \caption{Diameter-sampling mechanism for caliber. EDT-derived diameters are sampled along the GT and FSG centerlines. caliber\_diam\_mm\_IQR is the range of the middle 50\% of these diameter samples.}
    \label{fig:image20}
\end{figure}

Figure~\ref{fig:image20} traces the origin of the caliber feature: the EDT converts every vessel pixel into a distance-to-boundary value; centerline sampling retrieves local radii; and the IQR summarizes the central 50\% of the diameter distribution. A small boundary displacement can therefore change many sampled diameters. This is not the same quantity as the area fraction measured by VAF.

\subsection{Full Controlled Perturbation Protocol and Sensitivity Calculation}
\label{j.4-full-controlled-perturbation-protocol-and-sensitivity-calculation}

The sensitivity analysis did not retrain the segmentation networks. Instead, controlled perturbations were applied to the GT mask or skeleton, followed by feature re-extraction. Three perturbations were considered: one-pixel mask dilation (mask\_dilate1) to simulate vessel overestimation, one-pixel mask erosion (mask\_erode1) to simulate vessel underestimation or loss of thin branches, and five-pixel skeleton-tip pruning (skeleton\_prune\_tips5) to simulate missing terminal vessels.

Each perturbation was evaluated under two classification protocols. In the baseline-trained/perturbed-test protocol (baseline\_train\_perturbed\_test), the classifier was trained on unperturbed features and tested on perturbed features, measuring performance under feature-distribution shift. In the perturbed self-LOOCV protocol (perturbed\_self\_loocv), training and testing were both performed within the perturbed feature table, assessing whether class information remained separable after perturbation. The two protocols address different questions: the first approximates deployment under distribution shift, whereas the second evaluates performance after classifier re-adaptation.

For a feature $x$ in image $i$ and ring $r$, the standardized change was calculated as
\begin{equation}
\Delta z_{i,r}=\frac{\left|x_{i,r}^{\mathrm{perturbed}}-x_{i,r}^{\mathrm{base}}\right|}{\mathrm{SD}_{\mathrm{GT},r}}.
\label{eq:J1-sensitivity}
\end{equation}
\noindent Here $x_{i,r}^{\mathrm{base}}$ is the unperturbed GT value, $x_{i,r}^{\mathrm{perturbed}}$ is the value after the perturbation, and $\mathrm{SD}_{\mathrm{GT},r}$ is the across-image standard deviation of the original GT feature in the same ring. Averaging $\Delta z_{i,r}$ over all images and rings gives the $\mathrm{mean\_abs\_delta\_z}$ reported in the main-text sensitivity table (Table~\ref{tab:feature_stabilization}). A larger value means that a perturbation of the same nominal size produces a larger feature displacement, i.e., greater feature sensitivity.

The perturbations did not make all features worse in the same way. Dilation and erosion mainly affected area, diameter, and thin-vessel threshold statistics, whereas terminal pruning more readily affected centerline geometry and graph topology. This feature--perturbation correspondence is the principal complement that feature-level evaluation adds to a single pixel-level score.
They identify which representations are most exposed to local errors, but they do not by themselves show whether the resulting features can be made more stable without changing their task meaning. That question motivates the selective stabilization step below.

\subsection{Stabilization, Parameter Settings, and Processing Effects}
\label{j.5-stabilization-definitions-parameter-settings-and-processing-effects}

The objective of stabilization is not to make a mask uniformly smoother. It is to reduce amplification of local pixel-boundary noise in the derived feature values while preserving class differences and genuine vessel structure. We therefore did not apply spline processing indiscriminately to all 11 features; operations were selected according to each feature's computation mechanism and the GT perturbation results.

For each skeleton segment, the two-dimensional pixel coordinates were ordered by cumulative arc length s. Smooth splines were then fitted separately to x(s) and y(s), rather than forcing distinct branches into a single curve. The default implementation used smooth\_sigma\_px = 1.25, resampling at step\_dd = 0.02 DD, and retaining segments with at least eight sampling points. Arc length, directional variation, curvature, and diameter sampling were computed on this smoothed centerline.
The value ``1.25 pixels'' is a smoothing budget in the spline fit, not an expansion of the vessel radius by 1.25 pixels. It controls how closely the curve follows the pixel polyline. If too small, the result remains nearly pixel-like; if too large, genuine bends may be attenuated. The value should therefore be selected using parameter scans and fidelity checks, rather than chosen solely because it minimizes sensitivity.

The EDT input remains the original binary mask. For each vessel pixel, the Euclidean distance to the nearest background pixel is computed to form a radius map; multiplying by two gives a diameter map. Gaussian smoothing is applied only to this continuous diameter map: the smoothed diameter at a point is a distance-weighted average of neighboring diameters, with larger weights assigned to closer points. The binary-mask boundary is not moved, and the operation does not create or delete vessel pixels. The smoothed diameter is then sampled at spline-centerline points before calculating caliber\_diam\_mm\_IQR.

The main-text sensitivity table summarizes only the five features retained for final post-processing. Although spline centerline length was tested for VLD, its mean sensitivity across the three retained perturbations increased from 0.3055 to 0.3220 (-5.4\%); VLD was therefore excluded from the final stabilization scheme. The reported values are averaged across one-pixel mask dilation, one-pixel mask erosion, and five-pixel skeleton-tip pruning. ``Improvement rate'' is defined as \(1-\mathrm{processed}/\mathrm{original}\); a positive value indicates lower sensitivity.
These results support selective stabilization rather than universal smoothing. The three centerline-geometry features showed the largest gains; the caliber improvement arose from spline sampling and EDT continuous averaging; and vascular connection count (VCC) showed a modest reduction targeted at short terminal spurs.

\begin{figure}[H]
    \centering
    \includegraphics[width=0.9\linewidth]{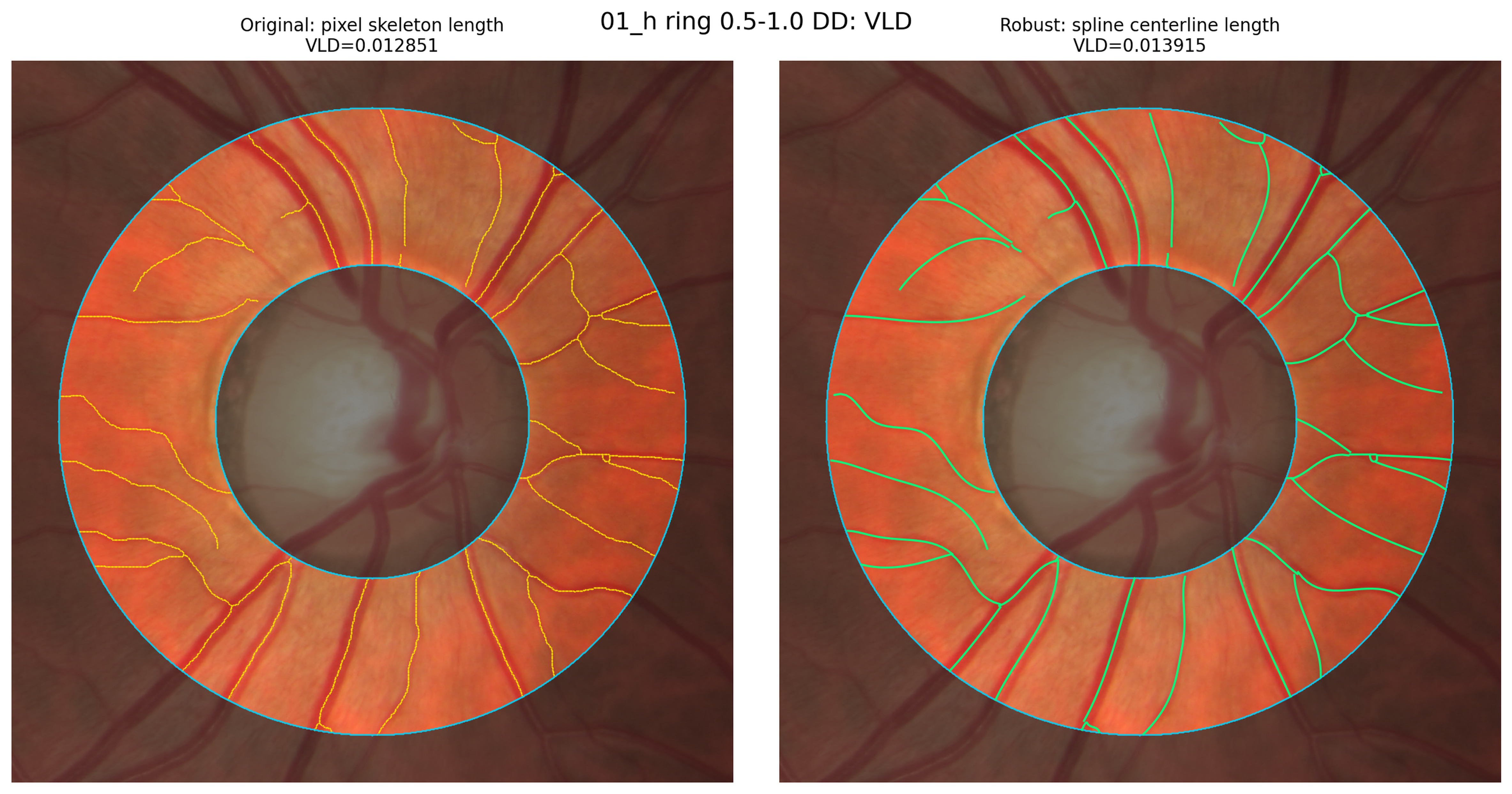}
    \caption{Pixel skeleton versus spline-smoothed centerline for VLD. The pixel polyline is replaced by a continuous arc-length-fitted spline within the 0.5--1.0 DD ring.}
    \label{fig:image21}
\end{figure}

Having selected the processing operations from the perturbation evidence, we can now ask whether the stabilized representation changes the relative classification performance of the six mask inputs, which are shown in Figure~\ref{fig:image21}. 

\begin{figure}[H]
    \centering
    \includegraphics[width=0.9\linewidth]{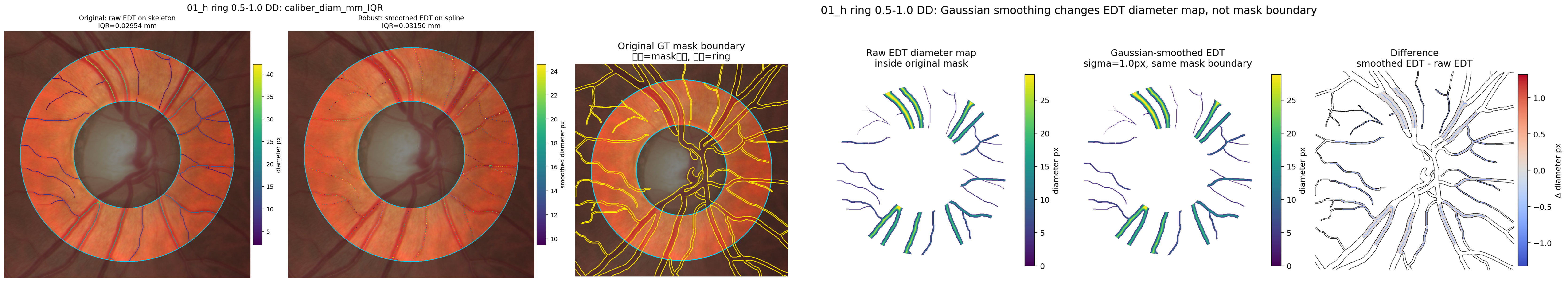}
    \caption{
    The left panel shows diameters sampled from the Gaussian-smoothed EDT along the spline centerline. The right panel shows the raw EDT, smoothed EDT, and their difference. Both panels use the original binary mask with the mask boundary.}
    \label{fig:image22}
\end{figure}

\begin{figure}[H]
    \centering
    \includegraphics[width=0.9\linewidth]{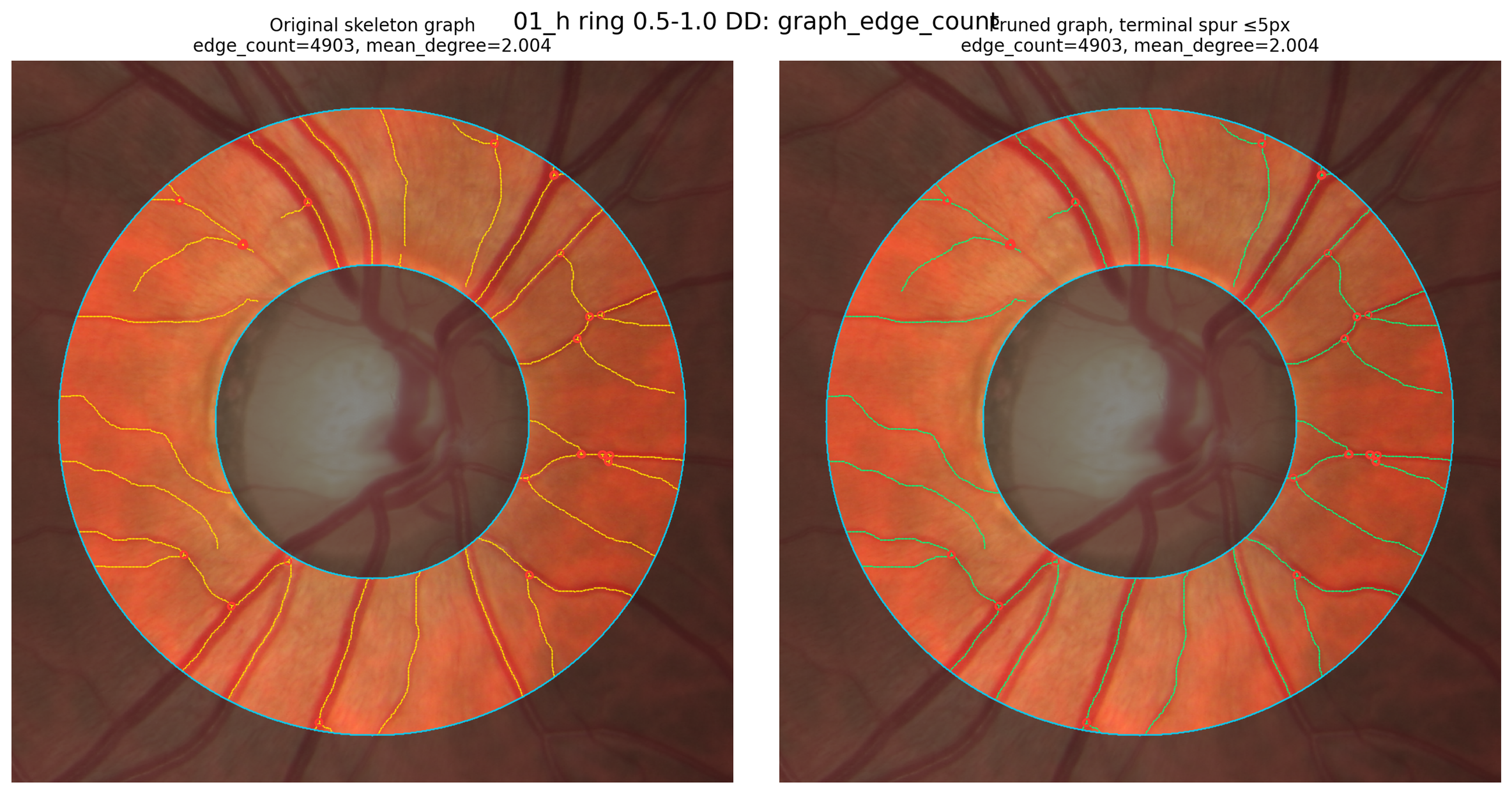}
    \caption{Graph pruning: original skeleton graph versus terminal-spur pruning. The left panel shows the original graph; the right panel removes terminal spurs no longer than 5 pixels. Red nodes indicate local graph nodes, while the main trunk connectivity is preserved as far as possible.}
    \label{fig:image23}
\end{figure}

Figure~\ref{fig:image22} shows the spline centerline with Gaussian-smoothed EDT sampling and the associated diameter-map comparison. Figure~\ref{fig:image23} shows the corresponding terminal-spur pruning of the skeleton graph.

\subsection{Complete Robust-Feature Classification and Coefficient Rankings}
\label{j.6-complete-robust-feature-classification-and-coefficient-rankings}

Within each segmentation method, all 11 vascular features were re-extracted. Only the five features described above were replaced by stabilized definitions; the remaining six retained their original definitions, including VLD. The same 45-image LOOCV was then applied. This comparison therefore measures the classifiability produced by different masks under one consistent, selectively stabilized feature-extraction protocol.

Taken together, the stabilized experiment provides the downstream counterpart to the segmentation benchmark: once the feature computation is made less sensitive to local mask noise, FSG-Net and SA-UNetv2 become the leading predicted-mask inputs on HRF. This agreement with their strong structure-level performance is encouraging.

Classification results are reported in the main-text Table~\ref{tab:classification_after_stabilization}. Accuracy is still calculated as the number of correctly classified images divided by 45; the 95\% Wilson interval describes uncertainty for this finite sample.

The original and stabilized analyses are separate experiments and should not be conflated. The original vascular-only baseline used the unprocessed feature definitions and produced GT 33/45, FSG 27/45, DCP 24/45, GAVE 34/45, FRUNet 28/45, and SAUNet 29/45. The method-specific feature-type ranking is provided in Table~\ref{tab:top_vascular_features}, whereas the robust-5 classification results are reported in Table~\ref{tab:classification_after_stabilization}. The robust-5 analysis replaced only the five selected feature definitions and retained the other six, producing GT 35/45, FSG 30/45, DCP 26/45, GAVE 25/45, FRUNet 27/45, and SAUNet 29/45. The paired robust-5-minus-original changes were +0.0444, +0.0667, +0.0444, -0.2000, -0.0222, and 0.0000, respectively. These changes are reported descriptively; no inferential test or multiple-comparison-adjusted p-value is reported for this small HRF analysis.

\begin{table}[H]
\caption{Method-specific top five vascular feature types ranked by mean absolute coefficient magnitude, averaged across rings.}
\label{tab:top_vascular_features}
\scriptsize

\begin{tabularx}{\textwidth}{
    @{}
    >{\raggedright\arraybackslash}p{0.14\textwidth}
    >{\raggedright\arraybackslash}X
    @{}
}
\toprule
\textbf{Method} &
\textbf{Top five feature types (mean absolute coefficient)} \\
\midrule

GT &
1. VAF (0.301);
2. VCC (0.234);
3. $D_{\mathrm{ratio}}$ (0.221);
4. $L_{\mathrm{thin}}$ (0.206);
5. median absolute curvature (0.149) \\
\addlinespace

FSG-Net &
1. caliber diameter IQR (0.289);
2. median absolute curvature (0.231);
3. $L_{\mathrm{thin}}$ (0.201);
4. $VC_{\mathrm{mean}}$ (0.201);
5. $D_{\mathrm{ratio}}$ (0.189) \\
\addlinespace

DCP-1024 &
1. caliber diameter IQR (0.296);
2. $D_{\mathrm{ratio}}$ (0.249);
3. $VC_{\mathrm{mean}}$ (0.229);
4. VAF (0.201);
5. median signed curvature (0.199) \\
\addlinespace

FR-UNet &
1. caliber diameter IQR (0.299);
2. $VC_{\mathrm{mean}}$ (0.276);
3. VCC (0.239);
4. VAF (0.218);
5. median absolute curvature (0.206) \\
\addlinespace

GAVE &
1. caliber diameter IQR (0.260);
2. $VC_{\mathrm{mean}}$ (0.236);
3. median absolute curvature (0.213);
4. $L_{\mathrm{thin}}$ (0.203);
5. median signed curvature (0.202) \\
\addlinespace

SA-UNetv2 &
1. caliber diameter IQR (0.277);
2. $L_{\mathrm{thin}}$ (0.209);
3. $D_{\mathrm{ratio}}$ (0.201);
4. $VC_{\mathrm{mean}}$ (0.194);
5. VAF (0.194) \\

\bottomrule
\end{tabularx}
\end{table}

\FloatBarrier

\end{document}